\documentclass[10pt,letterpaper,twocolumn]{article}

\usepackage[utf8]{inputenc} %
\usepackage[T1]{fontenc}    %
\usepackage{hyperref}       %
\usepackage{url}            %
\usepackage{booktabs}       %
\usepackage{amsfonts}       %
\usepackage{nicefrac}       %
\usepackage{microtype}      %
\usepackage{graphicx}
\usepackage{doi}
\usepackage{amsmath}
\usepackage{amssymb}
\usepackage{amsthm}
\usepackage{xcolor}
\usepackage{nicefrac}
\usepackage{enumitem}
\usepackage{bm}
\usepackage{subfig}
\usepackage{floatpag}
\usepackage{multirow}
\usepackage{cite}
\usepackage{IEEEtrantools}
\usepackage{microtype}
\usepackage{changepage}
\usepackage{tocbibind}

\floatpagestyle{plain}

\newlength{\w}
\newlength{\titlefigurewidth}
\graphicspath{{pdftex}{figures}}

\newcommand{\RRR}{{\mathbb{R} }}

\newcommand{\Qc}{{\mathcal{Q} }}
\newcommand{\Mc}{{\mathcal{M} }}
\newcommand{\Sc}{{\mathcal{S} }}
\newcommand{\Tc}{{\mathcal{T} }}
\newcommand{\bs}[1]{\boldsymbol{#1}}
\newcommand{\ttheta}{\bs{\theta}}

\newcommand{\qq}{\bs{q}}
\newcommand{\dqq}{\dot{\bs{q}}}

\newcommand{\vv}{\bs{v}}

\newcommand{\xx}{\bs{x}}
\newcommand{\dxx}{\dot{\bs{x}}}

\newcommand{\ff}{\bs{f}}
\newcommand{\gggg}{\bs{g}}
\newcommand{\hh}{\bs{h}}

\newcommand{\nn}{\bs{n}}

\newcommand{\AAA}{\bs{A}}

\newcommand{\CC}{\bs{C}}
\newcommand{\DD}{\bs{D}}

\newcommand{\II}{\bs{I}}
\newcommand{\JJ}{\bs{J}}
\newcommand{\KK}{\bs{K}}

\newcommand{\MM}{\bs{M}}

\newcommand{\PP}{\bs{P}}

\newcommand{\UU}{\bs{U}}
\newcommand{\VV}{\bs{V}}

\newcommand{\xxi}{\bs{\xi}}

\newcommand{\ttau}{\bs{\tau}}

\newcommand{\vvarphi}{\bs{\varphi}}

\newcommand{\dxxi}{\dot{\bs{\xi}}}

\newcommand{\nnull}{\bs{0}}

\newcommand{\diag}{{\mathrm{diag}}}

\renewcommand{\d}{{\mathrm{d}}}

\newcommand{\pdx}[2]{\frac{\partial #1}{\partial #2}}

\newtheorem{theorem}{Theorem}
\newtheorem{theorem*}{Theorem}

\newtheorem{definition*}{Definition}

\newtheorem{lemma*}{Lemma}

\newtheorem{corollar}{Corollary}
\newtheorem{corollary*}{Corollary}
\newtheorem{proof*}{Proof}

\theoremstyle{plain}
\newtheorem*{ps}{Problem Statement}

\theoremstyle{remark}
\newtheorem{problem}{Problem}

\makeatletter
\renewcommand*\env@matrix[1][\arraystretch]{%
	\edef\arraystretch{#1}%
	\hskip -\arraycolsep
	\let\@ifnextchar\new@ifnextchar
	\array{*\c@MaxMatrixCols c}}
\makeatother

\definecolor{TUMBlue}{HTML}{0065BD}

\newcommand{\q}{\boldsymbol{q}}
\newcommand{\qdot}{\dot{\boldsymbol{q}}}
\newcommand{\qddot}{\ddot{\boldsymbol{q}}}
\newcommand{\x}{\boldsymbol{x}}

\newcommand{\xdot}{\dot{\boldsymbol{x}}}
\newcommand{\xddot}{\ddot{\boldsymbol{x}}}

\newcommand{\JT}{\boldsymbol{J}^\top}

\newcommand{\torque}{\boldsymbol{\tau}}
\newcommand{\force}{\boldsymbol{f}}

\newcommand{\Minv}{\boldsymbol{M}^{-1}}

\newcommand{\vphi}{\vvarphi}
\newcommand{\vphidot}{\dot{\vvarphi}}
\newcommand{\vphiddot}{\ddot{\vvarphi}}

\renewcommand{\b}[1]{\boldsymbol{#1}}

\DeclareMathOperator{\pol}{pol}

\DeclareUnicodeCharacter{2212}{-}

\newcommand{\subf}[4]{\centering\subfloat[#1]{\def\svgwidth{#4}#4\input{pdftex/#4.pdf_tex}}}

\newcommand*{\tran}{^{\mathsf{T}}}
\definecolor{TUMBlue}{HTML}{0065BD}
\definecolor{TUMSecondaryBlue}{HTML}{005293}
\definecolor{TUMSecondaryBlue2}{HTML}{003359}
\definecolor{TUMBlack}{HTML}{000000}
\definecolor{TUMWhite}{HTML}{FFFFFF}
\definecolor{TUMDarkGray}{HTML}{333333}
\definecolor{TUMGray}{HTML}{808080}
\definecolor{TUMLightGray}{HTML}{CCCCC6}
\definecolor{TUMAccentGray}{HTML}{DAD7CB}
\definecolor{TUMAccentOrange}{HTML}{E37222}
\definecolor{TUMOrange}{HTML}{E37222}
\definecolor{TUMAccentGreen}{HTML}{A2AD00}
\definecolor{TUMAccentLightBlue}{HTML}{98C6EA}
\definecolor{TUMAccentBlue}{HTML}{64A0C8}

\newcommand{\thetitle}{Redundancy Resolution at Position Level}

\newcommand\blfootnote[1]{%
  \begingroup
  \renewcommand\thefootnote{}\footnote{#1}%
  \addtocounter{footnote}{-1}%
  \endgroup
}

\title{{\LARGE\bf\textsc{\thetitle}} \\\vspace{.2cm} {\textbf{\textit{A Preprint}}}}

\author{
	\begin{tabular}{cc}
		\begin{minipage}{.5\textwidth}
			\normalsize
			\begin{center}
	\href{https://orcid.org/0000-0001-5343-9074}{\includegraphics[scale=0.06]{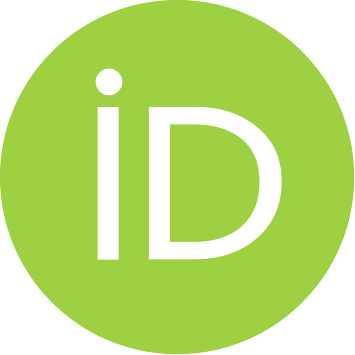}\hspace{1mm}\textbf{Alin Albu-Schäffer}} \\
	Institute of Robotics and Mechatronics\\
	German Aerospace Center (DLR)\\
	Oberpfaffenhofen, Germany\\
	\texttt{alin.albu-schaeffer@dlr.de}
			\end{center}
		\end{minipage}
	&
		\begin{minipage}{.5\textwidth}
			\normalsize
			\begin{center}
\href{https://orcid.org/0000-0003-4974-4134}{\includegraphics[scale=0.06]{orcid.pdf}\hspace{1mm}\textbf{Arne Sachtler}} \\
	School of Computation, Information and Technology\\
	Technical University of Munich (TUM)\\
	Garching, Germany\\
	\texttt{arne.sachtler@dlr.de}\\
			\end{center}
		\end{minipage}
	\end{tabular}
}

\date{\today}

\hypersetup{
pdftitle={\thetitle},
pdfauthor={Alin Albu-Schäffer, Arne Sachtler},
}

\begin{document}
\bstctlcite{IEEEexample:BSTcontrol}
\onecolumn

\maketitle
\begin{abstract}
	\normalsize Increasing the degrees of freedom of robotic systems makes them more versatile and flexible.
This usually renders the system kinematically redundant: the main manipulation or interaction task does not fully determine its joint maneuvers.
Additional constraints or objectives are required to solve the under-determined control and planning problems.
The state-of-the-art approaches arrange tasks in a hierarchy and decouple lower from higher priority tasks on velocity or torque level using projectors.
We develop an approach to redundancy resolution and decoupling on position level by determining subspaces of the configurations space independent of the primary task.
We call them \emph{orthogonal foliations} because they are, in a certain sense, orthogonal to the task self-motion manifolds.
The approach provides a better insight into the topological properties of robot kinematics and control problems, allowing a global view.
A condition for the existence of orthogonal foliations is derived.
If the condition is not satisfied, we will still find approximate solutions by numerical optimization.
Coordinates can be defined on these orthogonal foliations and can be used as additional task variables for control.
We show in simulations that we can control the system without the need for projectors using these coordinates, and we validate the approach experimentally on a 7-DoF robot.

\end{abstract}

\blfootnote{
\textbf{Copyright Note: } \copyright2023 IEEE
This is the author's version of an article that has been accepted for publication in IEEE Transactions on Robotics.
Personal use of this material is permitted. Permission from IEEE must be obtained for all other uses, in any current or future media, including reprinting/republishing this material for advertising or promotional purposes, creating new collective works, for resale or redistribution to servers or lists, or reuse of any copyrighted component of this works in other works. \textcolor{TUMBlue}{DOI: 10.1109/TRO.2023.3309097}}

\blfootnote{
\textbf{Ancillary Files: } Four ancillary files are available on the \texttt{arXiv}. 
Please see the end of this paper for details.
Some figure captions within the paper directly point to the files. The three videos are recommended as visual aids for understanding.
}

\begin{figure}[h]
    \centering
    \def\svgwidth{.75\titlefigurewidth}
    \footnotesize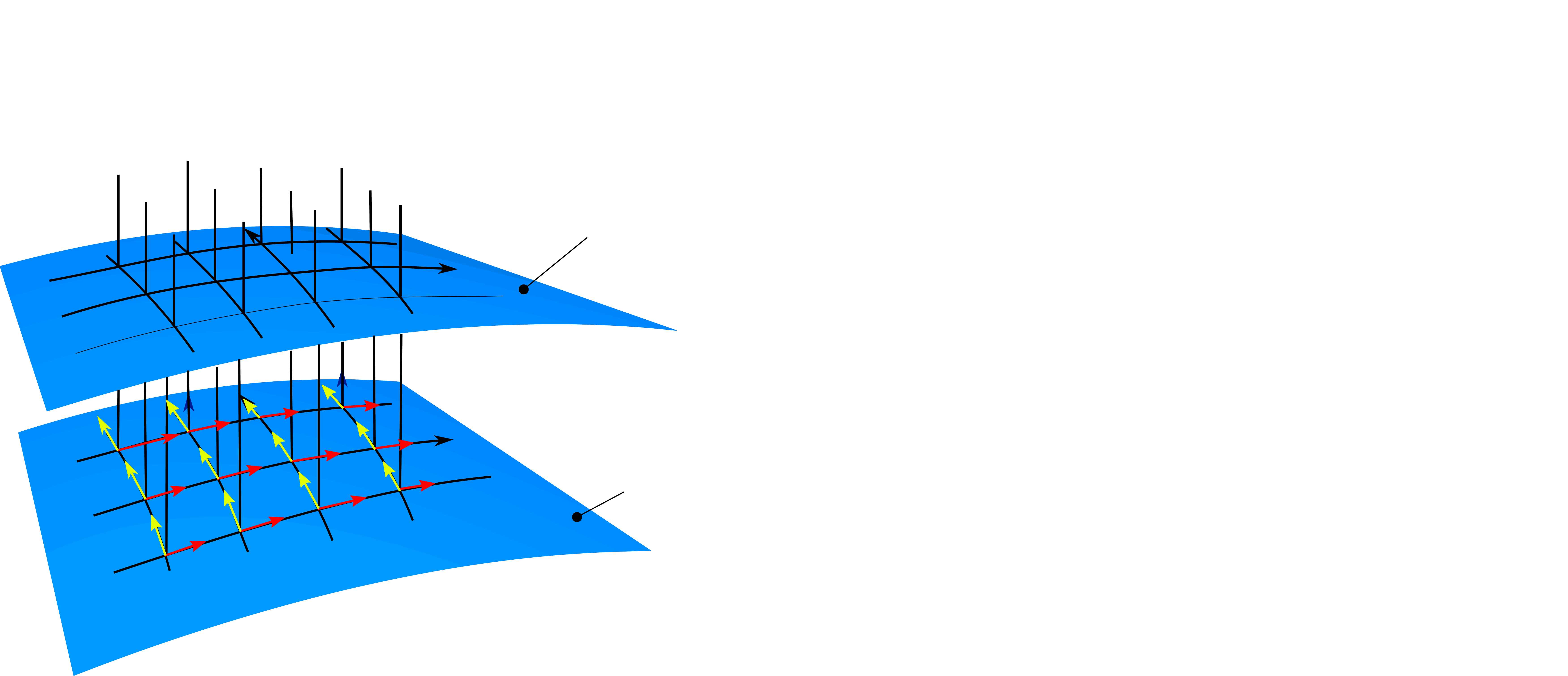
	\captionsetup{width=\titlefigurewidth}
    \caption{
    The idea of orthogonal foliations (and coordinates): given the foliation induced by the forward kinematics of a redundant manipulator, we assign coordinates $(\xi_1, \xi_2)$ on the self-motion motion manifolds (denoted in blue), such that the coordinates remain invariant when moving to another leaf of the foliation by following the task gradient (i.e the Jacobian in the kinematic case). 
    This leads to foliations that are orthogonal to the task-induced foliation in the sense of a specified metric.
    Two leaves of the orthogonal foliations are shown in red and yellow.
    }
    \label{fig:consistent_grid}
\end{figure}

\newpage
\tableofcontents

\twocolumn
\section{Introduction}\label{sec:introduction}
Robots are called \emph{redundant} if the number of degrees of freedom of the mechanical structure exceeds the degrees of freedom of the task.

Kinematics and dynamics of redundant robots belong to the classical and most well-studied theoretical foundations of robotics~\cite{Khatib1987,Khatib1995,Nakamura1987,Siciliano1991,Baerlocher1998,Liegeois1977,Nakanishi2008,Siciliano1990,Dietrich2013}.
The state of the art for redundancy resolution is based on projection operators and arranges multiple tasks in a hierarchy of priorities.
It is summarized below.

\ifx\undefined\ieee\else\begin{figure}[h]
    \centering
    \def\svgwidth{\w}
    \footnotesize\input{pdftex/manifold_consistent_grids.pdf_tex}
    \caption{
    The idea of orthogonal foliations (and coordinates): given the foliation induced by the forward kinematics of a redundant manipulator, we assign coordinates $(\xi_1, \xi_2)$ on the self-motion motion manifolds (denoted in blue), such that the coordinates remain invariant when moving to another leaf of the foliation by following the task gradient (i.e the Jacobian in the kinematic case). 
    This leads to foliations that are orthogonal to the task-induced foliation in the sense of a specified metric.
    Two leaves of the orthogonal foliations are shown in red and yellow.
    }
    \label{fig:consistent_grid}
\end{figure}\fi
\subsection{Projection-based Redundancy Resolution in a Nutshell}
Projection-based redundancy resolution is one of the basic approaches presented in the standard literature~\cite{Siciliano2008a,Murray1994,Lynch2017,Spong2005,Siciliano2008}.
The main idea is to add additional tasks that use up the remaining degrees of freedom of the mechanical structure.
Then, the projection operators ensure that lower-priority tasks are decoupled from higher-priority tasks.
We denote the forward kinematics mapping for two task priorities between the configuration (joint) space $\Qc$ and the task space $\Mc_i$ by $\hh_i : \Qc\rightarrow \Mc_i$.
Local coordinates for $\Qc$ are denoted by $\qq \in \RRR^n$ while $\xx_i \in \RRR^{m_i}, \, i=\{1,2\}$, $m_1 + m_2 \le n$ are local coordinates for two levels of task priorities, where task $1$ is higher in priority than task $2$.
The Jacobians are $\JJ_i =\pdx{\hh_i(\qq)}{\qq}$.

While the higher priority task is usually a specification of the manipulation task performed with the end-effector, the lower priority often optimizes kinematic or dynamic properties of the robot.
These objectives are formulated as functions of the configurations variables, usually being some kind of thresholded inverse distance functions for (self) collision avoidance \cite{Maciejewski1985,Dietrich2012,Hagn2009}, values required to optimize manipulability \cite{Hutzl2014,Su2019} or inertial properties~\cite{Walker1990,Mansfeld2017a}.

The classical redundancy resolution on velocity level~\cite{Liegeois1977,Nakamura1987,Siciliano1991,Nakanishi2008,Baerlocher1998} at non-singular configurations ($\det(\JJ_1(\qq)\JJ_1\tran(\qq)) \neq 0$) imposes the following relations between task and joint space velocities
\begin{subequations}
	\begin{align}
		\dqq   &= \JJ_1^\#(\qq) \dxx_1 +  \dqq_N \\
		\dqq_N &=  \PP(\qq) \JJ_2^\#(\qq) \dxx_2.   \label{eq:null-space-velocity}
	\end{align}
\end{subequations}
		
Therein,  $\dqq_N$ are null space velocities, $\cdot^{\#}$ a pseudo-inverse and $\PP(\qq)$ is a projector onto the null space of $\JJ_1$
\begin{equation}
   \PP(\qq) = \II -  \JJ_1^{\#A}(\qq)  \JJ_1(\qq),
   \label{eq:def-projector}
\end{equation}
with $\JJ_1^{\#A} = \AAA^{-1}\JJ_1\tran(\JJ_1  \AAA^{-1} \JJ_1\tran)^{-1}$ being the generalized pseudo-inverse~\cite{Doty1993}, with $\AAA$ positive-definite and symmetric, satisfying $\JJ_1(\qq) \PP(\qq) = \nnull$ and thus ensuring $\JJ_1(\qq) \dqq_N = \nnull$.

Similarly, in the framework of operational space control~\cite{Khatib1987}, redundancy resolution of the generalized joint force $\ttau$ considering primary and secondary task forces $\ff_1$ and $\ff_2$, respectively, is obtained according to \begin{subequations}
		\begin{align}
			\ttau   &= \JJ_1\tran(\qq) \ff_1 +  \ttau_N   \label{eq:torque_rr1} \\
			\ttau_N &=  \PP\tran(\qq) \JJ_2\tran(\qq) \ff_2. \label{eq:torque_rr2}
		\end{align}
\end{subequations}

Since the projector satisfies $\PP\tran(\qq) \JJ_1\tran(\qq) \ff_1 = \nnull$, it removes any generalized joint force component, which would interfere with the one generated by a primary task force $\ff_1$ over the Jacobian $\JJ_1\tran(\qq)$. 
Using the inertia tensor $\AAA = \MM(\qq)$ in the pseudo-inverse, one obtains the dynamically decoupled projection, additionally satisfying that $\ttau_N$ generates no instantaneous primary task acceleration $\ddot\xx_1$~\cite{Khatib1987}.

\subsection{Drawbacks of the Projection-based Approach}\label{sec:problems-projection}
The projection-based approaches are the state-of-the-art approach for redundancy resolution for a good reason: they provide a flexible framework by allowing almost arbitrary additional tasks.
However, there are also some problems that we would like to identify here.
Later, in Section~\ref{sec:projection} and a supplementary video, we show examples of these problems.

Generally, tasks on different priorities are incompatible:
we can not simultaneously achieve the tasks on all levels.
The control errors of lower priority tasks will not vanish but instead converge to some non-zero minimum.
The projection framework \emph{allows} to have incompatible tasks.
Incompatible tasks need projection\footnote{
	Especially for a small degree of redundancy. 
	For highly redundant systems, the performance increase of projection methods stagnates, and one might consider simply superimposing impedance controllers~\cite{Hermus2022}.}!
The state-of-the-art relies on projection to make the tasks at least \emph{locally} compatible on the level of velocities or forces.
The global incompatibility of tasks causes a couple of follow-up problems.

\begin{problem}\label{prob:no_effect}
	Lower priority tasks may have no effect.
\end{problem}
The lower priority tasks may totally conflict with the primary task in certain configurations.
In this case, the lower priority tasks do not perform redundancy resolution because not all the robot's degrees of freedom are used.

\begin{problem}\label{prob:not_converging}
	Snapping between various minima.
\end{problem}
This is probably among the largest problems in practice.
Depending on the current configuration, the robot may converge to various local minima of the secondary tasks.
Imagine a hand-guiding scenario: the robot has converged to some configuration.
Now the user interacts and moves the robot to a new configuration that is contained in the region of attraction of another local minimum of the secondary tasks.
In practice, users observe unexpected jumps in the robot's behavior when it snaps between different local minima of the secondary tasks: the robot behaves unpredictably.

The same effect occurs when controlling the robot in the primary task.
Moving in the primary tasks causes a variation of the configuration --- and thus possible movement to regions of attraction of other minima of the secondary tasks.

\begin{problem}\label{prob:not_integrable}
	No integrability, no guaranteed return.
\end{problem}
The joint velocity vector fields obtained by the pseudo-inverse of $\JJ_1$ and also the corresponding force fields are generally not integrable~\cite{Klein1983}: performing a closed loop $\xx_1(t)$ for $\xx_1(t_0)=\xx_1(t_1)$ in the main task will in general not drive the system back to the original configuration $\qq(t_0) \ne \qq(t_1)$.
Using the projection framework to add secondary tasks, helps, but does not solve this problem.
Depending on the past configuration path, the robot may converge to different local minima of the secondary tasks~(example in Sec.~\ref{sec:projection}.
\ref{prob:not_integrable} also leads to:

\begin{problem}\label{prob:not_usable_for_planning}
	Entirety of tasks are not usable for planning.
\end{problem}
Specifying the task coordinates on all levels is insufficient to determine the robot's joint configuration: one also needs past joint configurations.
Hence, planning is not possible in the task coordinates.

\begin{problem}\label{prob:no_potential}
	Projected forces are no gradients of potentials.
\end{problem}
Passivity-based multi-layer control approaches~\cite{Khatib1987,Dietrich2012,Sentis2005} like the well-known impedance controller generate the task forces in~\eqref{eq:torque_rr1},~\eqref{eq:torque_rr2} as gradient force fields $\ff_i\tran= \pdx{U_i(\xx_i)}{\xx_i}, \,i={1,2}$, derived from potential fields $U_i$ formulated in task coordinates $\xx_i$.
The secondary task force $\ff_2$ is projected (\ref{eq:torque_rr2}) before it acts on the robot.
The projection modifies the energy flows in the system and destroys the safety-critical property of passivity\footnote{Unless also feeding back the external forces~\cite{Wu2023}.}~\cite{Dietrich2017,Folkertsma2017}.
Because of the projection, we cannot use the virtual potentials in secondary tasks as potential fields (Lyapunov functions) to prove convergence of the system.
Providing effective dynamical decoupling and proving convergence under these conditions is a complex challenge, still a topic of active research today~\cite{Wu2023,Dietrich2015,Antonelli2009,Dietrich2012,Ott2015,Dietrich2020}. 
In most null space projection-based controllers, the assumption of structural feasibility of the task coordinates is made~\cite{Dietrich2020} and this problem is deferred to a planning task.

\begin{figure}
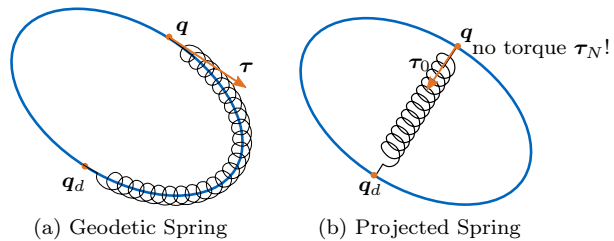

	\centering
	\subf{Geodetic Spring\label{fig:springs-geodetic}}{.4\w}{\footnotesize}{geodetic_spring}\hspace{.5cm}
	\subf{Projected Spring\label{fig:springs-projected}}{.4\w}{\footnotesize}{non_geodetic_spring}
	\caption{Comparison of a geodetic spring due to the position-level redundancy resolution approach in this paper and a projected spring by the state of the art approach (b). The spring in (b) has no effect as the joint torques vanish after the projection. The blue ellipses sketch a task self-motion manifold.}
	\label{fig:springs}
\end{figure}

\begin{problem}\label{prob:hidden_springs}
	Hidden springs.
\end{problem}
Suppose the controller in the secondary task is realized by some virtual spring $\ff_2 = -\KK_2 (\xx_2 - \xx_2^d)$ in secondary task coordinates as in impedance control.
The robot usually converges to local minima in the secondary tasks, i.e., one obtains non-zero torques $\torque = \JJ_2\tran \ff_2 \ne \nnull$ which vanish after the projection, such that $\PP\tran \JJ_2\tran \ff_2 = \nnull$. 
This corresponds to a tensioned virtual spring hidden behind the projector.
Even for compatible tasks, the spring may not relax entirely in some cases:
Fig.~\ref{fig:springs-projected} sketches this situation for a joint space spring. 

Slight variations of the configurations or the commanded motions may cause the previously hidden spring to discharge and create large forces and high velocity.
Additionally, a strongly loaded virtual spring often causes shaky behavior and unpleasant sounds at the configuration where it hides behind the projector.

\subsection{Redundancy Resolution on Position Level}
In this publication, we would like to introduce a completely new approach to redundancy resolution:
\begin{ps}
    Given the family of self-motion manifolds determined by the task space of a redundant robot, find subspaces of the configuration space that are compatible, not interfering with the task space.
\end{ps}
These compatible (or decoupled) subspaces will be locally described by constraint functions $\xxi(\qq) = c$.
Compatible means that the Jacobian of the main task coordinates $\xx = \hh(\qq)$ is orthogonal to the Jacobian of the additional task $\xxi(\qq)$, i.e., $\JJ_x \JJ_\xi\tran = \nnull$ or $\JJ_x \MM^{-1} \JJ_\xi\tran = \nnull$, depending on the desired properties of the coordinates.

Fig.~\ref{fig:consistent_grid} schematically visualizes the idea: let the blue surfaces denote two self-motion manifolds~\cite{Burdick1989} of the main task for two given task positions.
These are embedded submanifolds of the configuration space.
The desired new subspaces (or submanifolds) are chosen such that they intersect the task self-motion manifolds orthogonally\footnote{In a certain sense of \emph{orthogonal} to be derived later in Sec.~\ref{sec:orthfol}}, thereby generating coordinate systems on each self-motion manifold simultaneously.
The red and yellow surfaces sketch surfaces on which the new coordinates $\xxi_i(\qq)$ are constant.
It turns out that the entirety of self-motion manifolds can be interpreted as a foliation, which is a key concept for the following sections.
Likewise, our new compatible subspaces can be interpreted as a foliation, and the yellow and red surfaces in Fig.~\ref{fig:consistent_grid} are leaves of this foliation.
Because the new foliation is orthogonal to the task foliation, we call them \emph{orthogonal foliations}.
Such a redundancy resolution at configuration level, based on coordinates for the self-motion manifolds, would provide a more global perspective, which is not limited to the linearized vicinity of the current configuration.

Such subspaces would solve all the problems P1-P6 by using the coordinates $\xxi(\qq)$ instead of the projections $\PP(\qq)$.
This would also restore the passivity of the controlled system.
Potential fields in the coordinates $\xxi(\qq)$ will correspond to \emph{geodetic springs}, directly depending on the distance between the two end-points on the self-motion manifolds (Fig.~\ref{fig:springs-geodetic}).
The new Jacobian $\JJ_\xi(\qq)$ will span the same space as the projector matrix in the projection framework.
$\JJ_x \JJ_\xi\tran = \nnull$ corresponds to $\JJ_\xi$ spanning the same space as the Moore-Penrose-based projector $\PP(\qq)$ and $\JJ_x \MM^{-1} \JJ_\xi\tran = \nnull$ to $\JJ_\xi$ spanning the same space as the dynamically consistent projector.

In~\cite{Dyck2022}, we looked at a foliation of surfaces in task space.
This time we look at a foliation in configuration space induced by the choice of the main task.
In contrast to the previous work, however, these are not always two-dimensional, they are in configuration space, and we need another kind of measure of orthogonality.

Note that we do not aim to \emph{repair} an existing secondary task that one would use in the projection framework.
Instead, we aim to \emph{construct} a new secondary task that is compatible with the main task - and does not need projection at all.

\ifx\undefined\ieee\subsection{Roadmap}
We have identified a coupled of problems in Sec.~\ref{sec:problems-projection}.
In Sec.~\ref{sec:related-work} we discuss some existing approaches and their limitations.
Then we introduce basic concepts of robot kinematics and dynamics in terms of differential geometry Sec.~\ref{sec:diff_geometry}.
In particular, we shed a light onto the global behavior of self-motion manifolds of redundant robots and conclude that we can interpret the entirety of them as a foliation.
This is a key concept for the following sections.
In Sec.~\ref{sec:orthfol} we introduce the notion of orthogonal coordinates and discuss the conditions for their existence.
As it will turn out, they do not exist in general.
We will therefore discuss the cases in which integrability can be ensured and call the resulting coordinates $\xxi(\qq)$ \emph{orthogonal}.
If this is not the case, our proposed learning method will provide coordinates that are \emph{as orthogonal as possible} in a certain optimization sense.
Then, we call $\xxi(\qq)$ \emph{quasi-orthogonal}.

The next three sections are devoted to procedures to find such \emph{orthogonal} and \emph{quasi-orthogonal} foliations:\begin{description}
\item[Local Solution] (Sec.~\ref{sec:plane_stacks}) We discuss a very simple approximate solution in, which is only valid at a region around a predefined configuration.
However, it is a first starting point to understand the idea.
\item[Exact Solution] (Sec.~\ref{sec:growing}) We proceed to exact and global orthogonal foliations.
We choose very simple robots with only two and three degrees of freedom.
In these low dimensions we can easily visualize the results and discuss the properties of the orthogonal coordinates.
\item[Optimization Solution] (Sec.~\ref{sec:nn}) We show an optimization approach to find orthogonal and quasi-orthogonal coordinates also for higher dimensional robots.
\end{description}

We compare results of classical projection-based control to our proposed approach on an impedance-controlled simulated robot in Sec.~\ref{sec:sim}.
Additionally, we show results of the quasi-orthogonal coordinates approach on a real 7-DoF robot in Sec.~\ref{sec:hardware}.
We conclude by discussions in Sec.~\ref{sec:discussion}.
\fi
\subsection{Contributions}
Our contributions are:
\begin{itemize}
	\item[(C1)] Formulating redundancy resolution on position level for the first time;
	\item[(C2)] Introducing the differential geometric tools required to systemically develop this new theory, including the concepts of involutivity and foliations;
	\item[(C3)] A condition for the existence of orthogonal foliations corresponding to exact position-level redundancy resolution;
	\item[(C4)] In-depth analysis of orthogonal foliations for two and three degrees of freedom robots;
	\item[(C5)] Numerical schemes for approximate solutions to orthogonal coordinates for higher dimensional robots; and
	\item[(C6)] Simulation and hardware experiments for position-level redundancy resolution using passive impedance control in task and orthogonal coordinates and comparison to the state-of-the-art projection-based control.
\end{itemize}

\section{Related Work}\label{sec:related-work}
Some research groups have previously dealt with the non-integrability problem (\ref{prob:not_integrable}).
Already in 1991, Mussa-Ivaldi and Hogan \cite{Mussa-Ivaldi1991} proposed using a specially weighted pseudo-inverse of the Jacobian to obtain an integrable system.
Combining this with kinematic control, they achieved that one can start at some configuration, perform a closed path in task space, and return to the initial configuration.
This is a subtask to what we want to achieve but does not provide coordinates $\xxi(\qq)$.
One year later De Luca et al.~\cite{DeLuca1992} found special start configurations for which the system is cyclic.

More recently, Hauser and Emmons~\cite{Hauser2018} have proposed an approach to \emph{global redundancy resolution} constructing a probabilistic roadmap in configuration space that allows continuous pseudo-inversion of the forward kinematics; and Xie et al.~\cite{Xie2022} use a recurrent neural network for repetitive motions of redundant robots.
These publications provide solutions to global redundancy resolution by ensuring that each main task configuration is achieved by exactly one joint configuration.
This way, one achieves integrability, but loses flexibility as the remaining degrees of freedom are eliminated and cannot be used for secondary tasks.
In this paper, we want to retain all the degrees of freedom.

Haug~\cite{Haug2023} recently proposed a cyclic differentiable manifold representation of redundant manipulator kinematics.
He achieves cyclic behavior and still retains the freedom to use redundancy.
The approach is based on the insight that in an enrivonment around some \emph{base point} $\bar{\qq}^j$ \cite{Haug2022}:
\begin{equation}\label{eq:haug_subspaces}
    \qq(\xx, \vv) = \bar{\qq}^j + \VV^j \vv - \UU^j \boldsymbol{\psi}^j(\xx, \vv)
\end{equation}
where $\UU^j = \JJ\tran_x(\bar{\qq}^j)$ and $\VV^j \in \mathbb{R}^{n\times(n-m)}$ spans the null space of $\JJ_x(\bar{\qq}^j)$ such that $\JJ_x(\bar{\qq}^j)\VV^j = \boldsymbol{0}$ and $\VV^{j\mathsf{T}} \VV^j = \II$. 
The function $\boldsymbol{\psi}^j(\xx, \vv)$ is smooth and is determined iteratively by a Newton-Raphson method during operation~\cite{Haug2023}.
The values $\vv$ can be interpreted as coupled local coordinates, describing the arm configuration within the subspace around $\bar{\qq}^j$.
Because they are coupled to the main task $\xx$, the function $\boldsymbol{\psi}^j(\xx, \vv)$ depends on $\xx$ \emph{and} $\vv$.

Constructing a null space basis $\JJ_{\xi^\prime}(\qq)$ independently of an integral manifold $\xxi(\qq)$, i.e., purely based on linear algebra, is, of course, always possible and has been frequently addressed in the literature~\cite{Park1999,Ott2008b,Dietrich2015,Siciliano1990,Monari2023}.
Performing a null space projection based on~\eqref{eq:null-space-velocity} or~\eqref{eq:torque_rr2} can be equivalently formulated using the null space basis, see Sec.~\ref{sec:Jacobians-and-Projections}.
However, it is well-known~\cite{Klein1983} that the velocity vector fields obtained by the projector $\PP(\qq)$ in~\eqref{eq:null-space-velocity}, and equivalently, obtained by $\dot \xxi^\prime = \JJ_\xi^\prime(\qq) \dqq$, are in general not integrable, not fulfilling the conditions of the Frobenius theorem~\cite{Frobenius1877}.
\ifx\undefined\ieee%
For example, performing a loop and returning to the same robot joint configuration $\qq(t_0) = \qq(t_1)$ does not generally lead to a zero time integral of the null space velocities. 
So, when defining  $\xxi^\prime(t_1) = \xxi^\prime(t_0) + \int_{t_0}^{t_1} \dot\xxi^\prime(\tau) \, d\tau$ one would, in general, have $\xxi^\prime(t_1) \neq \xxi^\prime(t_0)$: $\xxi^\prime$ is not a function of $\qq$ only, but also depends on the path taken to reach $\qq$.
\fi

\section{A Geometric Light on Redundant Robots}\label{sec:diff_geometry}
We address the question of generalized configuration coordinates, which are orthogonal to given high-priority task coordinates.
This requires moving from vector spaces and linear algebra to smooth manifolds and the tools of differential geometry and topology. 
The first differential geometry-based analysis of redundant robots goes back to~\cite{Burdick1989}, who introduced the notion of self-motion manifolds to robotics. 
Differential-geometric robot kinematics and dynamics was given in~\cite{Selig1996,Bullo2004,Haug2022} and has since then been constantly used in robotics~\cite{Park1998,Bullo1999,Lachner2020}, although not yet belonging to the standard, widely used repertoire. 

Let $\Qc$ denote the configuration space manifold of a robot and $\Mc$ the task space manifold. 
We call $\mathcal{H}: \Qc\rightarrow \Mc$ the forward kinematics map.
Further, let $n = \dim \Qc$ and $m = \dim \Mc$.
A robot will be called redundant for a specific task if $m < n$ \cite{Khatib1995}.
Being redundant depends not only on the robot's degrees of freedom but also on the choice of task space.
For instance, when $\Mc = SE(2)$ or $\Mc = SE(3)$, this is precisely the case if $n > 3$ or $n > 6$.

Of course, various local coordinates can describe the task manifold, as highlighted in \cite{Murray1994} for the representation of orientations. 
The same applies to the configuration space manifold $\mathcal{Q}$.
Most statements will be coordinate-invariant when following a differential geometric perspective.

\begin{figure}
    \centering
    \def\svgwidth{\w}
    \scriptsize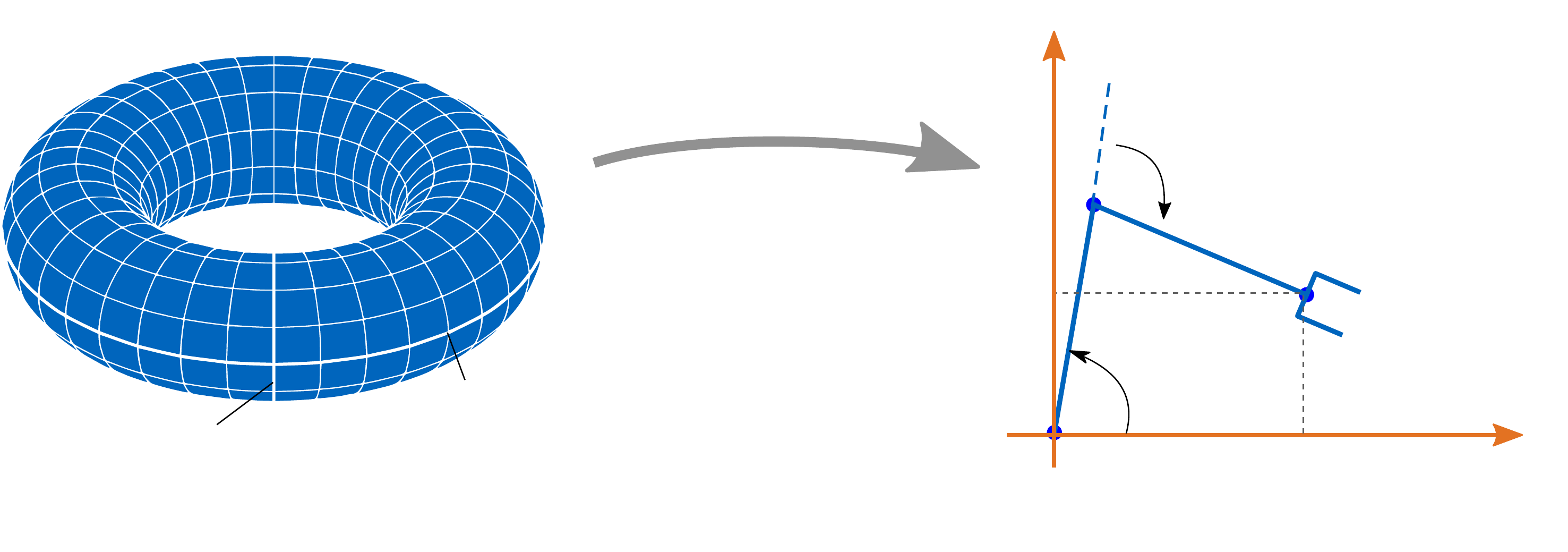
    \caption{
        The forward kinematics is a mapping between the joint manifold $\mathcal{Q}$ and the task manifold $\mathcal{M}$. 
        The task space manifold in this example is $\mathcal{M}=SE(2)$, which we sketch in a coordinate representation on the right.
    }
    \label{fig:forwkin}
\end{figure}

We denote local coordinates of a point in $\mathcal{Q}$ by $\q \in \mathbb{R}^n$ and point in $\mathcal{M}$ by $\x \in \mathbb{R}^m$.
In coordinates the mapping $\mathcal{H}$ will be denoted $\hh: \mathbb{R}^n \rightarrow \mathbb{R}^m$ such that $\x = \hh(\q)$.
Fig.~\ref{fig:forwkin} shows an example of such a mapping.
We will use a fair amount of abuse of notation in this paper: by $\q / \x$ we refer to both a point $\qq \in \Qc / \xx \in \Mc$ and to its coordinates $\qq \in \mathbb{R}^n / \xx \in \mathbb{R}^m$.

We assume there are no points in the configuration space $\Qc$ where the map $\mathcal{H}$ becomes singular, i.e. its push-forward has constant rank.
A singularity-free manifold $\bar{\Qc}$ can be constructed by removing the singular points from $\Qc$.
Note that $\bar{\Qc}$ is still a (now open) manifold when removing singular configurations.
We will use the symbol $\Qc$ to denote the singularity-free manifold throughout the paper.

\subsection{Push-Forwards and Pullbacks}
\begin{figure}
    \centering
    \def\svgwidth{\w}
    \footnotesize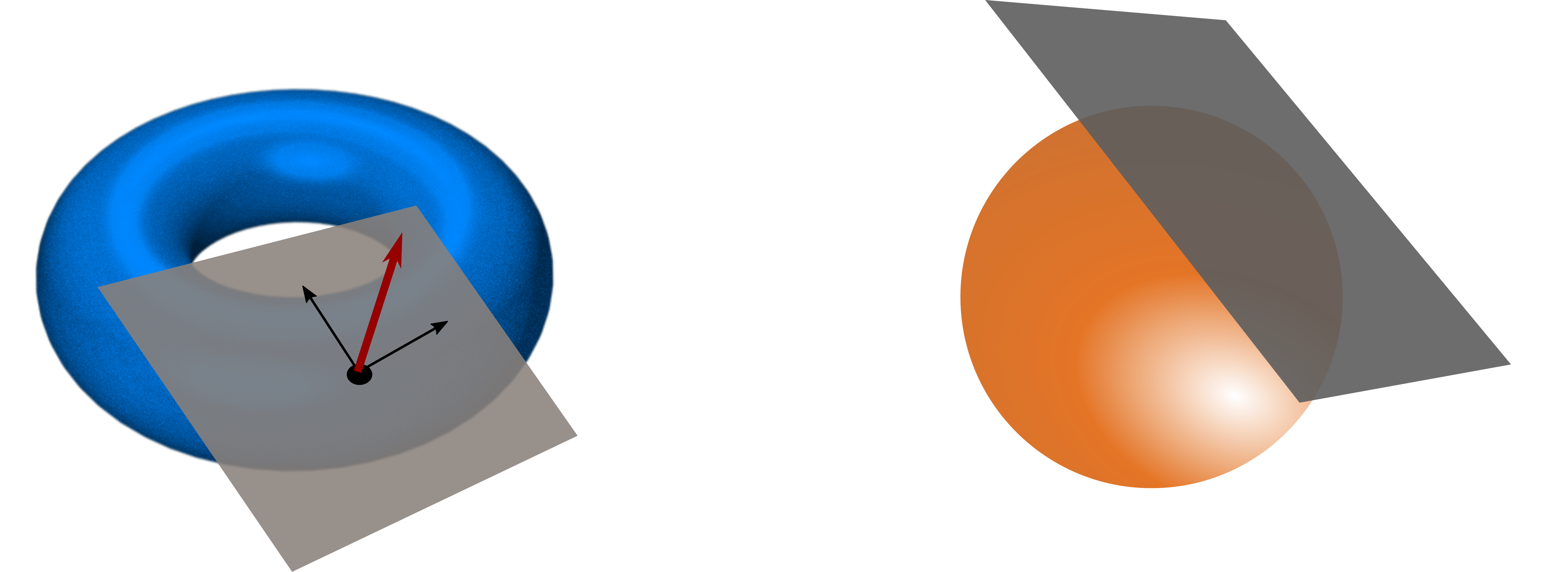
    \caption{
        The push-forward transforms vectors in the tangent space of the configuration space to vectors in the tangent space of the task space manifold.
    }
    \label{fig:pushforward}
\end{figure}

Velocities live in the tangent spaces $\Tc_q\Qc$ and $\Tc_x\Mc$ of the manifolds.
Generalized forces being covectors lie within the dual cotangent spaces $\Tc_q^*\Qc$ and $\Tc_x^*\Mc$.
The push-forward $\mathcal{H}_*$ of $\mathcal{H}$ maps vectors in $\Tc_q\Qc$ to vectors in $\Tc_{\mathcal{H}(q)}\Mc$ providing a local linear approximation of $\mathcal{H}$ (Fig.~\ref{fig:pushforward}).
Similarly, the pullback $\mathcal{H}^*$ maps covectors in $\Tc_{\mathcal{H}(q)}^*\Mc$ to covectors in $\Tc_q^*\Qc$.
Written in coordinates, the push-forward $\mathcal{H}_*$ can be thought of the Jacobian $\JJ =\pdx{\hh(\qq)}{\qq}$ and the pullback the transposed Jacobian.
This vocabulary is beneficial to apply statements and theorems from the differential geometry world to robotics.

\subsection{Forward Kinematics, Submersions \& Foliations}
From a differential geometric perspective, the forward kinematics map of a redundant robot is a \emph{submersion}: a map whose push-forward is surjective everywhere \cite{Lee2012}.
Note that this requires the absence of singularities, as assumed in this paper.
From the submersion level set theorem \cite[Cor. 5.13]{Lee2012} tells us that for each $\xx_0 \in \Mc$ the level set $\mathcal{H}(\q) = \xx_0$ of $\mathcal{H}$ is a closed embedded submanifold $\mathcal{S}_{0} \subset \mathcal{Q}$, i.e., $\forall \q \in \mathcal{S}_0: \mathcal{H}(\qq) = \xx_0$.
Further, the dimension of $\mathcal{S}$ is $n-m$.
These level set submanifolds are nothing but the well-known self-motion manifolds \cite{Burdick1989, Murray1994}.

Taking the entirety of the level-set submanifolds of $\Qc$ induced by $\mathcal{H}$, a foliation of codimension $m = \dim \Mc$ is obtained.
The foliation has as many dimensions as the degree of redundancy $r = n - m $.
Each leaf of the foliation is a self-motion manifold for one task configuration.
This foliation view on forward kinematics provides insight into the global behavior of the self-motion manifolds: some structure is present.
\emph{The self-motion manifolds fit together locally like slices in a flat chart} \cite[p. 501]{Lee2012}.

\begin{figure}
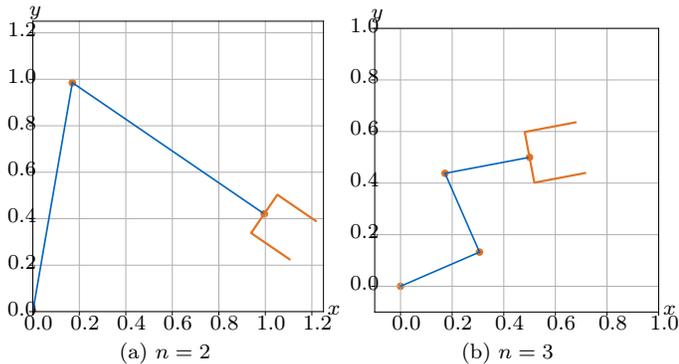

    \centering
    \subf{$n=2$\label{fig:exrobo2}}{.52\w}{\footnotesize}{exrobo}
    \hfill
    \subf{$n=3$\label{fig:exrobo3}}{.52\w}{\footnotesize}{robot_n3}
    \caption{Planar robots with $n=2$ and $n=3$ degrees of freedom used as example in this paper.}
    \label{fig:exrobo}
\end{figure}

Consider the two degrees of freedom (DoF) example robot in Fig.~\ref{fig:exrobo2}, whose configuration manifold is the 2-torus $\mathbb{T}^2$ (cf. Fig.~\ref{fig:forwkin}).
A one-dimensional task manifold and, thus, a scalar forward kinematics function is required to obtain a redundant system.
Without loss of generality, let us choose $h(\qq) = x$, i.e., we only care about the $x$-component of the end-effector position (horizontal).
The blue lines in Fig.~\ref{fig:submersion} show some level sets of the forward kinematics function $h$.
Note that the manifolds are closed on the torus $\mathbb{T}^2$ (Fig.~\ref{fig:fkin_on_torus_torus}).
Except at the singularities $\qq_1 = [0, 0]$, $\qq_2 = [\pm \pi, 0]$, $\qq_3 = [0, \pm \pi]$ and $\qq_4 = [\pm \pi, \pm \pi]$ the self-motion manifolds look locally like slices of a flat chart. 
Also, note that those nine singular configurations in coordinate representation correspond to only four points on the torus.
\ifx\ieee\undefined%
Fig.~\ref{fig:n2_fkin_and_quiver} additionally shows the gradient vector field of the forward kinematics function $h$, corresponding to the Jacobian row.
The orange arrows show the gradient vector field pre-multiplied by the mass-inertia matrix.
\fi

\begin{figure}[h!]
    \centering
    \subf{In the $q_1$ and $q_2$ plane\label{fig:submersion}}{.62\w}{\tiny}{submersion}
    \hspace{.5em}
    \subf{On the 2-torus\label{fig:fkin_on_torus_torus}}{.4\w}{\normalsize}{foliation_2_torus}
    \caption{
        Foliation on the 2-torus induced by a forward kinematics function. 
        The leaves of the foliation correspond to the well-known self-motion manifolds. 
        In the left image, the opposing sides of the rectangle are identified according to the torus topology. 
        The orange crosses mark the singular configurations.
    }
    \label{fig:fkin_on_torus}
\end{figure}

\ifx\ieee\undefined
\begin{figure}[h!]
	\centering
	\def\svgwidth{1.1\w}
	\footnotesize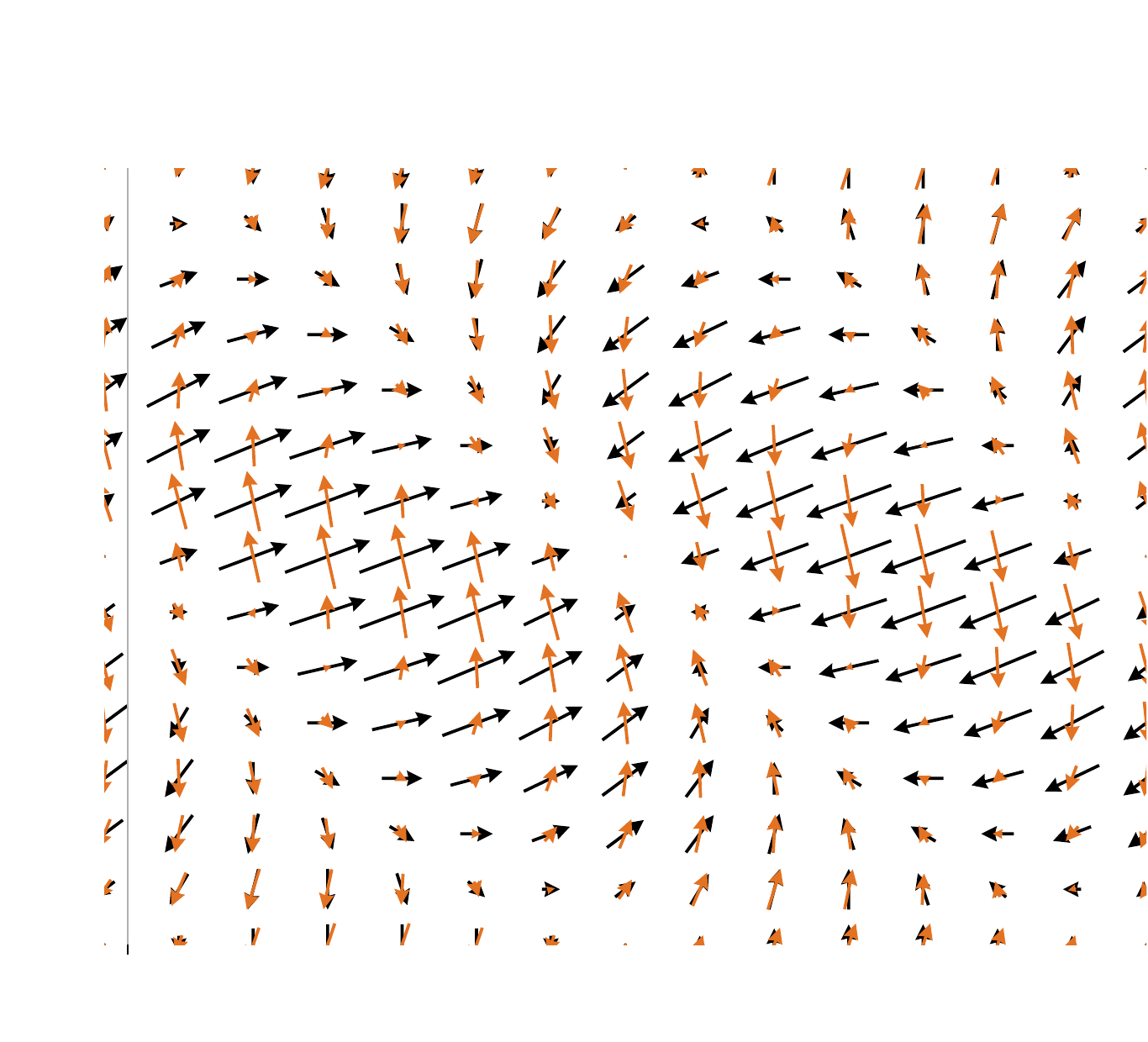
	\caption{Isolines of the forward kinematics function (blue) for the example robot with two degrees of freedom. Additionally, the arrows show the gradient of the forward kinematics function of the Riemannian manifold for two different choices for the metric. Black arrows: Euclidean metric. Orange Arrows: inertia tensor as metric.}
	\label{fig:n2_fkin_and_quiver}
\end{figure}
\fi

\begin{figure}[h!]
    \centering
    \subfloat[One leaf\label{fig:one_leaf_with_y}]{
        \includegraphics[height=4.5cm]{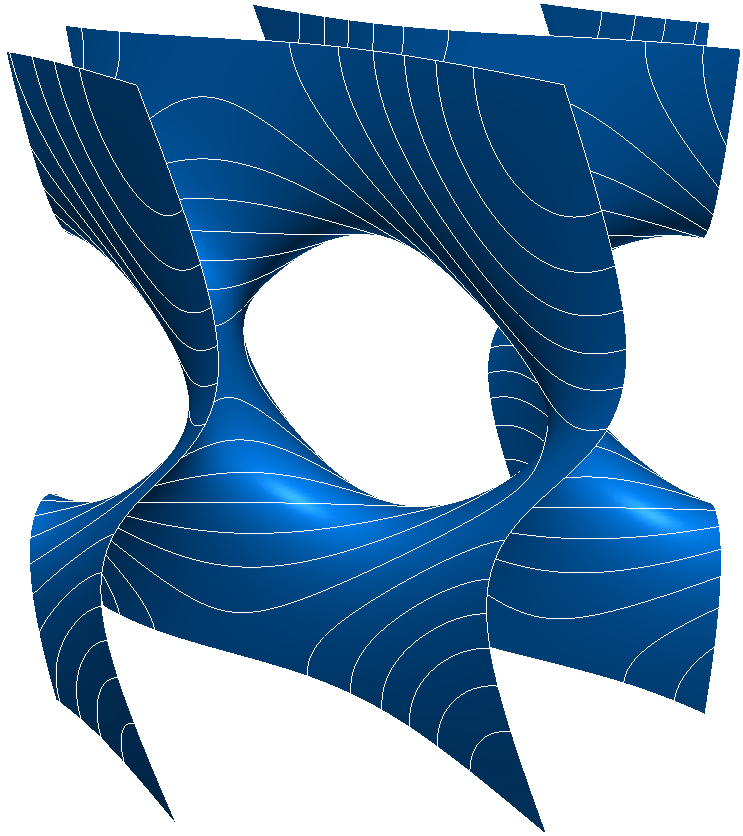}
    }
    \subfloat[Multiple leaves\label{fig:three_dof_foliation}]{
        \includegraphics[height=4.5cm]{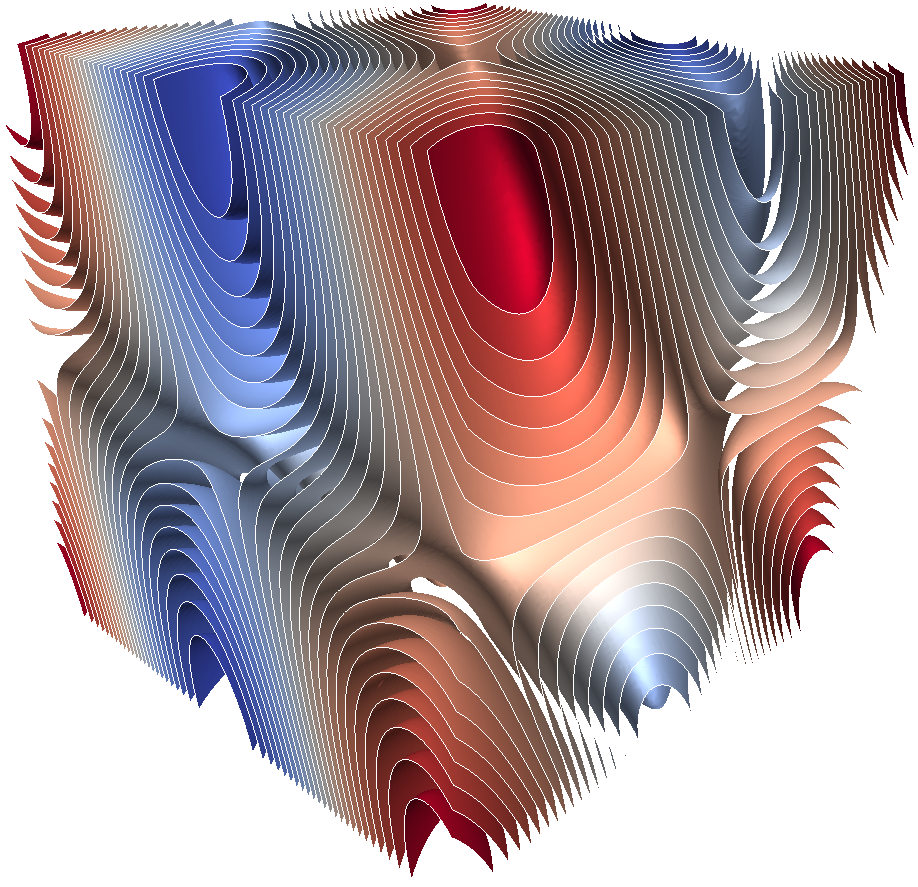}
    }
    \caption{
        Self-motion manifold(s) for a planar manipulator with three degrees of freedom and the $x$-component of the end-effector position as single task space. 
        The surfaces are generated using marching cubes \cite{Lewiner2003} on the forward kinematics function. 
        On the left: a single leaf of the foliation. 
        You may want to 3D-print the model from the supplementary material for a more intuitive visualization.
        The white lines additionally show level sets of the $y$-coordinate. 
        On the right: multiple leaves of the foliation.
    }
    \label{fig:manifold_base}
\end{figure}

Similarly, consider the second example robot with 3 DoF in Fig.~\ref{fig:exrobo3}.
We again consider only the $x$-component of the end-effector position and, consequently, obtain submanifolds of dimension two.
Consider the submanifold of $\Qc$ corresponding to $h(\qq) = 0$ shown in Fig.~\ref{fig:one_leaf_with_y}.
Now, we obtain a curved surface as self-motion manifold.
The surface is still closed on the 3-torus $\mathbb{T}^3$.
We show only one leaf of the foliation.
An additional rendering of multiple leaves for different level sets of $h(\qq)$ is shown in Fig.~\ref{fig:three_dof_foliation}.
Imagine taking all of them.
Then the manifolds look locally like a stack of paper.

\begin{figure}
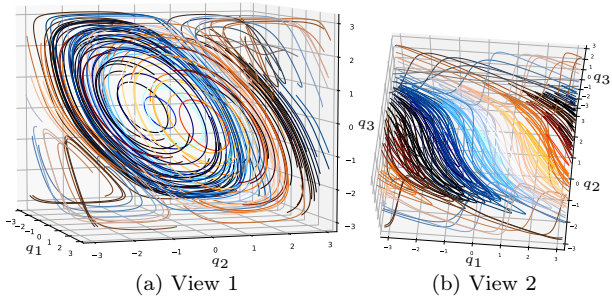

    \centering
    \subf{View 1}{.6\w}{\tiny}{spaghetti1}
    \subf{View 2}{.38\w}{\tiny}{spaghetti2}
    \caption{
        Foliation of the 3-torus. 
        Opposing faces of the cube are identified. 
        Considering the planar Cartesian position of a 3 DoF planar robot as task induces a one-dimensional foliation. 
        The video \texttt{smm.mp4} in the ancillary files helps to understand this figure.}
    \label{fig:3dof-2tasks}
\end{figure}

Finally, let us consider another choice of task coordinates for the 3 DoF robot by taking the end-effector position's $x$ and $y$ components.
We then obtain line-shaped self-motion manifolds as shown in Fig.~\ref{fig:3dof-2tasks}.
Note that one can classify the manifolds into two categories.
One class of lines encircles the $q_2 = q_3 = 0$ line, while the others do not.
Also, some positions in task space generate one self-motion manifold, while others generate two disconnected ones.
The latter ones correspond to the redundant version of \emph{elbow-up and elbow-down} configurations.
The video in the supplementary material helps to understand Fig. 8.
Consider~\cite{Burdick1989} for a more rigorous analysis.
The main observation required for this paper is that the lines form a foliation of $\Qc$.
Each point $\qq \in \Qc$ is part of exactly one of the curves.

\section{Orthogonal Foliations}\label{sec:orthfol}
This paper aims to find new foliations orthogonal to the task-induced foliation $\hh(\qq)$ to the best extent possible. 
Thinking of the leaves of the new foliations as submanifolds of $\mathcal{Q}$, we denote the constraints inducing these foliations by $\xxi(\qq)$.
Orthogonality is defined here as follows:
Given the Jacobians $\JJ_x$ and $\JJ_\xi$ defined in terms of
\begin{subequations}
    \begin{align}
        \dxx & = \JJ_x(\qq) \dqq     \label{eq:Jacobians_x} \\
        \dxxi & = \JJ_\xi(\qq) \dqq,     \label{eq:Jacobians_xi}
    \end{align}
\end{subequations}
we would like to find $\xxi(\qq)$ such that the two Jacobians are orthogonal with respect to the metric tensor $\AAA$:
\begin{equation}\label{eq:unified}
    \JJ_x(\qq) \AAA^{-1}  \JJ\tran_\xi(\qq) = \b{0},
\end{equation}
with $\AAA=\II$ or $\AAA=\MM(\qq)$ to achieve kinematic or dynamic decoupling, respectively\footnote{See Sec.~\ref{sec:Jacobians-and-Projections} for clarifying why the inverse of $\AAA$ appears in~\eqref{eq:unified}.}.

The property (\ref{eq:unified}) corresponds to the condition on the Jacobians and projectors for the classical velocity- and torque-based redundancy resolution.
However, instead of first defining a null space base matrix $\JJ_\xi$ and then checking for integrability, we follow another procedure:
we start with a parametrizable foliation and then ask for orthogonality of the Jacobians.
If orthogonality is achievable everywhere, we will call the resulting foliations \emph{orthogonal foliations}.

On the other hand, if the task-induced foliation does not admit orthogonal foliations satisfying~\eqref{eq:unified} everywhere, we will minimize the residuum over the entire considered configuration space.
In the latter case, we will call the foliations \mbox{\emph{quasi-orthogonal foliations}}.

\subsection{Geometric Intuition of Orthogonal Foliations}\label{sec:geometric_intuition}
Before introducing the proposed approach's rigorous differential geometric formulation, we explain the main ideas on simple 2 and 3 DoF examples.

Let us consider again the 3 DoF robot with 1-dimensional task space (cf. Fig.~\ref{fig:manifold_base}). 
Fig.~\ref{fig:consistent_grid} sketches two leaves of the foliation induced by the forward kinematics.
On a base leaf, say $\mathcal{S}_0$ defined by $h(\qq)=x_0$, we choose a chart $\xxi_{x_0}$ with coordinates $\{\xi_1, \xi_2\}$.
The Jacobian\footnote{The superscript in $\JJ_\xi^{x_0}$ shall denote that this Jacobian is only valid at configurations $\qq$ where $h(\qq) = x_0$.} $\JJ_\xi^{x_0}$ associated to $\xxi_{x_0}$ forms a basis for the tangent space at each point, as denoted by the red and yellow arrows.
The vectors corresponding to the Jacobian $\JJ_x$ of the task coordinate are dark blue.
Regarding the manifolds as embedded in Euclidean space, $\JJ_x$ is at each point orthogonal to the manifold and thus normal to the tangent space, satisfying $\JJ_x\, \JJ_\xi\tran=\b{0}$.
Now, integrating along the Jacobian $\JJ_x$ starting from a point $\xxi_{x_0}=\{0,0\}$ from $ \mathcal{S}_0$, one obtains a curve which intersects $\mathcal{S}_1$. 
We assign the coordinates $\xxi_{x_1}=\{0,0\}$ to this point.
Using this procedure, we can define an equivalence relation between all points in the chart $\xxi_{x_0}$ and points locally from $\mathcal{S}_1$, thus inducing a chart $\xxi_{x_1}$ on $\mathcal{S}_1$.
This way, we constructed two new foliations $\xxi_1$ and $\xxi_2$ orthogonal to $\xx$ in the Euclidean sense.
The leaves $\xi_1=0$ (red) and $\xi_2=0$ (yellow) are exemplarily displayed in Fig.~\ref{fig:consistent_grid}.
Note that the procedure described above works for any joint space dimension $n$, as long as the task space is one-dimensional and thus the dimension of the task-space induced foliation is $n-1$.

\begin{figure}
    \centering
    \def\svgwidth{1.1\w}\footnotesize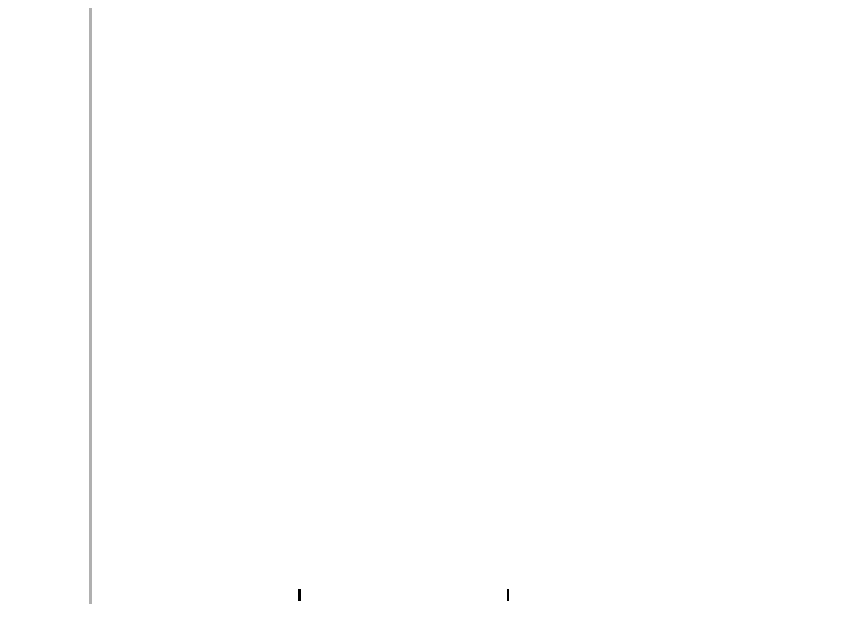
    \caption{
        Orthogonal foliations for the example of a bi-dimensional configuration space and a single task coordinate for dynamic consistency. 
    }
    \label{fig:2dof-foliations-Jacobians}
\end{figure}

The above example leads to foliations $\xxi$ satisfying the generalized orthogonality condition \eqref{eq:unified} for $\AAA = \II$.
For the case $\AAA = \MM(\qq)$, the geometric interpretation and visualization become more complex, and we discuss it for the simpler 2 DoF case with 1-dimensional task space.
The task-induced foliation is then one-dimensional. 
The blue curves in Fig.~\ref{fig:2dof-foliations-Jacobians} show some of its leaves $\mathcal{S}_i$, and dark blue arrows show the Jacobian $\JJ_x$.
As usual, it satisfies $\JJ_x(\qq) \vv = 0$ for any tangent vector $\vv \in \Tc_q\mathcal{S}_i$\footnote{
	By $\Tc_q\mathcal{S}$ we refer to the subspace $\Tc_q\mathcal{S} \subset \Tc_q\Qc$ of the tangent space $\Tc_q\Qc$ spanned by the tangent vectors of the embedded submanifold $\Sc \subset \Qc$, i.e. the tangent vectors of $\Tc_{\xi(\qq)}\Sc$ given as embedded vectors in $\Tc_{\qq}\Qc$.
}.
The condition $\JJ_x \MM^{-1} \JJ\tran_\xi =0$ thus implies that now $\MM^{-1} \JJ\tran_\xi$ is tangent to the foliation and forms a basis for the null space of $\JJ_x$.
Red arrows display this vector field in Fig.~\ref{fig:2dof-foliations-Jacobians}.
In this case, $\JJ_\xi$ being orthogonal to the $\xxi$ foliation, does not belong to the tangent bundle of the task foliation but is rotated accordingly (orange arrows).
Finally, integrating the $\MM^{-1} \JJ\tran_x $ vector field (black arrows) yields leaves of the $\xxi$ foliation, displayed as orange lines.

Let's conclude: the orthogonal foliation for dynamic consistency has $\MM^{-1} \JJ\tran_x $ as tangent vectors and $\JJ_\xi$ as normal vectors, while the task foliation has $\MM^{-1} \JJ\tran_\xi$ as tangent vector and $\JJ_x$ as normal vector.
Again, these results can be easily extended to configuration spaces of any dimension as long as the task space has a single dimension.

\begin{figure}
    \def\svgwidth{.9\w}\normalsize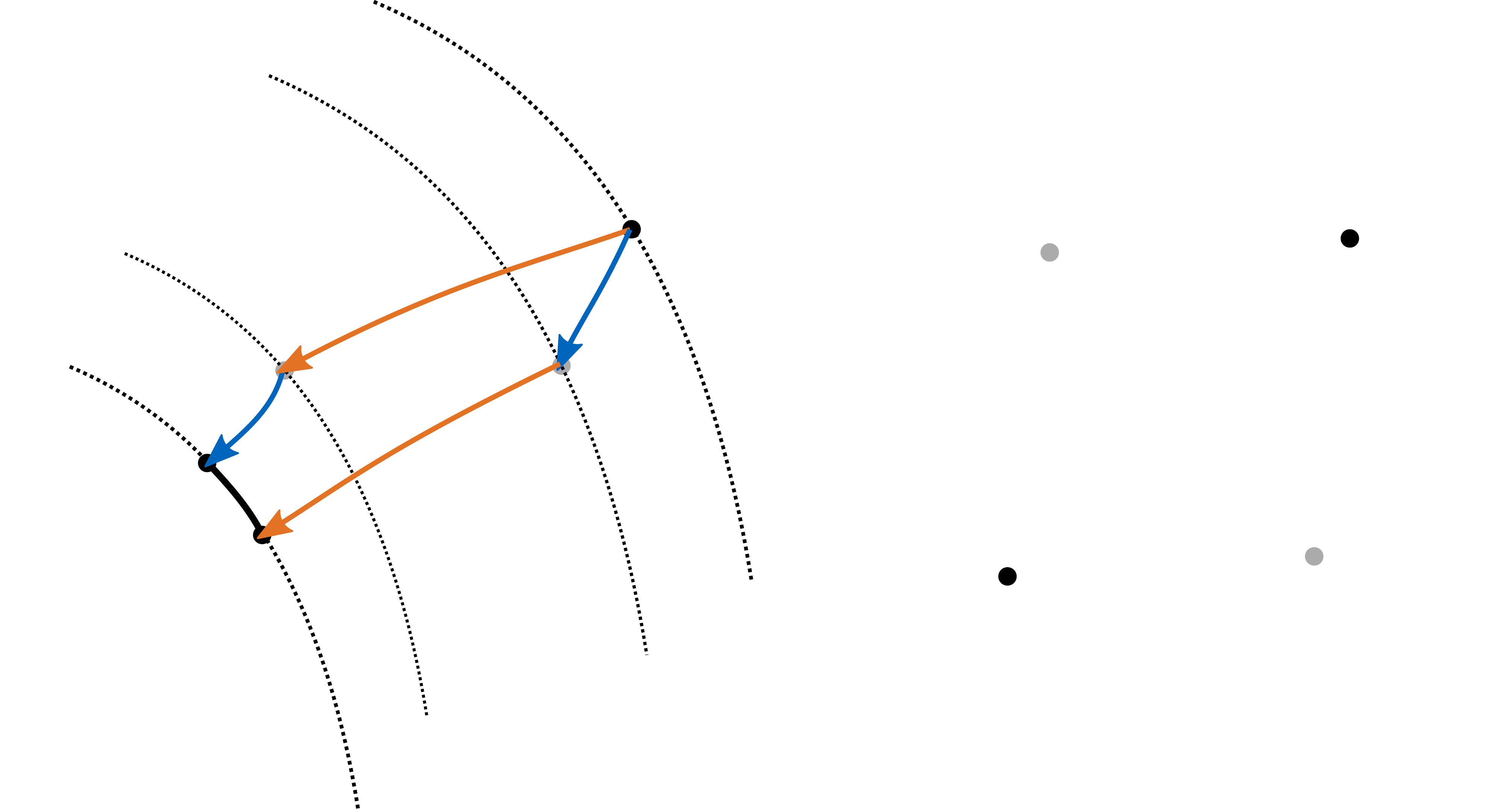
    \caption{The distribution of vector fields spanned by the task space Jacobian is generally non-involutive. Our solution provides the best integral manifold approximation (gray surface) in a specified optimalily sense.}
    \label{fig:noninv}
\end{figure}

Finally, let us consider the 3 DoF case, with 2-dimensional task space. 
We again choose the Euclidean orthogonality condition for simplicity: $\AAA = \II$.
The task foliation consists of curves, as displayed in Fig.~\ref{fig:noninv}.
At each point, the rows of $\JJ_x$ span a plane orthogonal to the corresponding task foliation leaf (or self-motion manifold).
They define two vector fields in the tangent bundle.
For these vector fields to be integrable, they must satisfy the Frobenius theorem \cite{Frobenius1877}.
In general, this condition will, however, not be satisfied.

Take a configuration $\qq_0$.
This will be part of precisely one leaf $\Sc_0$ of the task foliation.
Let this leaf of the foliation be defined by $\forall \qq \in \Sc_0: \hh(\qq) = \xx_0$.
Now, we follow the vector fields given by the rows of the Jacobians as shown in Fig.~\ref{fig:noninv}.
We take two paths leading to the same task configuration $\hh(\qq_a) = \hh(\qq_b) = \xx_1$.
Due to the non-involutivity we will generally have $\qq_a \ne \qq_b$, where $\qq_a, \qq_b \in \Sc_1$.
Therefore, the two final configurations $\qq_a$ and $\qq_b$ cannot form, together with $\qq_0$, a single leaf of an orthogonal foliation.
When navigating from $\xx_0$ to $\xx_1$ based on the gradient vector fields, the final configuration at $\xx_1$ will depend on the path between $\xx_0$ and $\xx_1$ in task space.
This is the essence of non-integrability.
In the infinitesimal case, the difference $\qq_b - \qq_a$ corresponds to the Lie bracket $[\JJ_{x_1}\tran, \JJ_{x_2}\tran]$ of the two vector fields.
In general, there is no integral surface that, even locally, is orthogonal to every leaf of the task foliation and would intersect $\Sc_1$ in a single point independent of the path.

However, in this case, one can ask to find the surface, which is \emph{as orthogonal as possible} to the task foliation, in terms of a cost function.
This means that $\JJ_x$ shall be \emph{as tangent as possible} to the desired foliation $\xxi$.
A leaf of this foliation, passing through $\qq_0$ is visualized in gray in Fig.~\ref{fig:noninv}.

\subsection{Null Space Jacobians and Projectors}
\label{sec:Jacobians-and-Projections}
Let us return to the relation between a null space basis $\JJ_\xi(\qq)$ and projection matrix $\PP(\qq)$.
Independent of the existence question about an integral manifold, a null-space basis on velocity level, as defined in \eqref{eq:Jacobians_x}-\eqref{eq:unified}, can always be constructed. 
In compact form,~\eqref{eq:Jacobians_x}-\eqref{eq:Jacobians_xi} can be written as
\begin{equation}\label{eq:definition_jacobian_Xi}
    \vphidot = \JJ(\qq) \dqq,
\end{equation}
with $\vphidot = \begin{bmatrix}\dot \xx\tran & \dot \xxi\tran\end{bmatrix}\tran$ and
\begin{equation}\label{eq:definition_jacobian_Xi_stacked}
    \JJ(\qq) = 
    \begin{bmatrix}
    \JJ_x(\qq)  \\ \JJ_\xi(\qq)
    \end{bmatrix}.
\end{equation}
Since $\dot \xxi$ is constructed such that $\JJ(\qq)$ is full rank, the stacked Jacobian $\JJ(\qq)$ can be inverted. 
It can be verified using \eqref{eq:unified} that the inverse is
\begin{equation}\label{eq:inverse_of_pseudoinverses}
    \JJ^{-1}(\qq) = \begin{bmatrix} \JJ_x^{\#A}(\qq) & \JJ_\xi^{\#A}(\qq) \end{bmatrix},
\end{equation}
since $\JJ(\qq) \JJ^{-1}(\qq) = \II$. 
This implies as well $ \JJ^{-1}(\qq) \JJ(\qq) = \II$ leading to
\begin{equation}
    \JJ_x^{\#A}(\qq) \JJ_x(\qq) + \JJ_\xi^{\#A}(\qq)\JJ_\xi(\qq) = \II.
\end{equation}
Therefore, there is a simple relation between the projector $\PP(\qq)$ defined as in \eqref{eq:def-projector}, the task Jacobian, and the null space basis $\JJ_\xi(\qq)$:
\begin{equation}
    \PP(\qq) = \II - \JJ_x^{\#A}(\qq) \JJ_x(\qq) = \JJ_\xi^{\#A}(\qq) \JJ_\xi(\qq).
\end{equation}
This highlights that in cases in which exact orthogonality as demanded in \eqref{eq:unified} can be achieved for the foliation $\xxi(\qq)$, projection-based control approaches are equivalent on velocity and torque level to the controllers based on the orthogonal manifold. 
At the same time, the latter has the advantage of additionally providing a coordinate (position) level. 

\subsection{On Coordinate Independence}
By looking at a concrete example, let us clarify why the integrability conditions on $\JJ_{\x}$ are indeed coordinate-independent.
Take the 3 DoF manipulator with two-dimensional task space and consider two different choices of task coordinates \begin{equation}
    \begin{bmatrix}
        x \\ y   
     \end{bmatrix} = \hh_c (\qq) \text{ and }
    \begin{bmatrix}
        r \\ \varphi
     \end{bmatrix} = \hh_p (\qq),
\end{equation}
where both specify the position of the end-effector, once in Cartesian and once in polar coordinates.
In the latter case, $r$ represents the radius and $\varphi$ the angle.
The associated ($\mathbb{R}^{2\times 3}$) Jacobians are \begin{equation}
    \begin{bmatrix}
        \dot{x} \\ \dot{y} 
    \end{bmatrix} = \JJ_c(\qq)\qdot \text{ and }
    \begin{bmatrix}
        \dot{r} \\ \dot{\varphi} 
    \end{bmatrix} = \JJ_p(\qq)\qdot,
\end{equation}
where $\JJ_c = \nicefrac{\partial \hh_c}{\partial \qq}$ and $\JJ_p = \nicefrac{\partial \hh_p}{\partial \qq}$.

\begin{figure}
    \def\svgwidth{.9\w}\footnotesize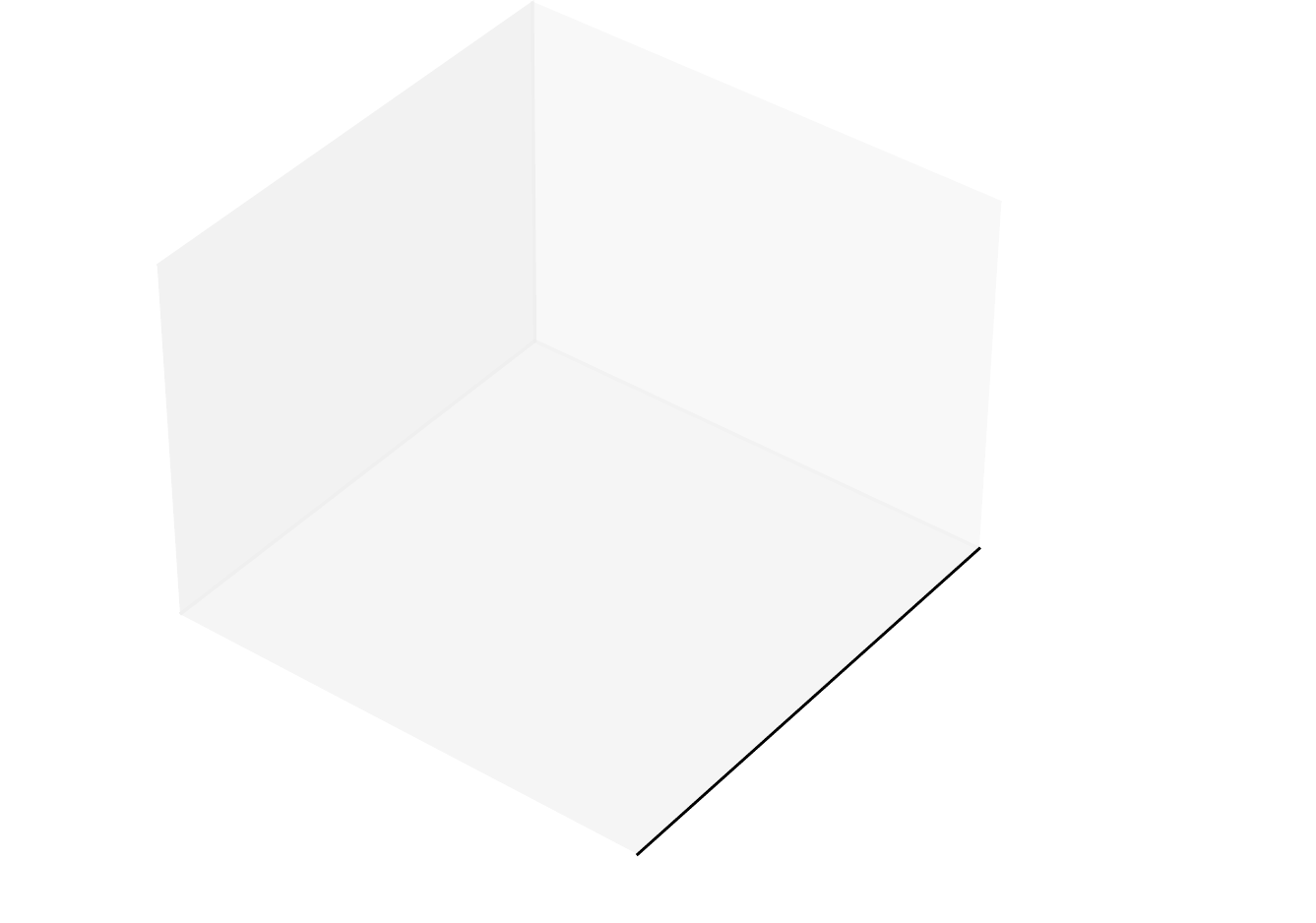
    \caption{
        The subspace spanned by the rows of the task Jacobian does not depend on the choice of task coordinates. 
        The task is the position of the end-effector. 
        We evaluate the Jacobian and plot the rows for Cartesian coordinates (red and orange) and for polar coordinates (black and light-blue). 
        The gray plane visualizes the 2d subspace spanned by all four vectors.
    }
    \label{fig:coordinate_invariance}
\end{figure}

Next, consider the function $\pol(x, y) = (r, \varphi)$ mapping positions in the plane from Cartesian to polar coordinates.
One can write the function $\hh_p$ as a composition of $\hh_c$ and $\pol$, i.e., \begin{equation}\label{eq:composition_polar}
    \hh_p(\qq) = \pol(\hh_c(\qq)).
\end{equation}
Let $\JJ_{\pol} \in \mathbb{R}^{2\times 2}$ denote the Jacobian of $\pol$ and let us differentiate (\ref{eq:composition_polar}) \begin{equation}
    \begin{bmatrix}
        \dot{r} \\ \dot{\varphi}
    \end{bmatrix} = \JJ_{\pol} \JJ_c \qdot = \JJ_{p} \qdot.
\end{equation}
It follows that $\JJ_p = \JJ_{\pol} \JJ_c$.
If $\JJ_{\pol}$ is full-rank, then $\JJ_p$ and $\JJ_c$ will have the same null space.
Moreover, the rows of $\JJ_c$ and $\JJ_p$ all span the same subspace of $\mathbb{R}^3$, which is orthogonal to the self-motion manifold at each point.
Fig.~\ref{fig:coordinate_invariance} shows this for one point on one of the task foliations.

This reasoning is valid for any dimension of configuration and task space. 
In differential geometric terms, the orthogonal linear subspaces at all points of the task foliation form a distribution\footnote{Note that this is not a distribution in the probability sense, but a distribution in the differential geometric sense: an assignment of tangent subspaces to every point on the manifold. 
This may help for an introduction: \cite[Ch. 19]{Lee2012}.\label{foootnote:distribution}} whose integrability we are interested in. 
This distribution only depends on the mapping to the task manifold but not on its charts.
Different choices of coordinates on the task manifold will create different vector fields.
These, however, will all span the same distribution.

\subsection{Differential Geometric Formulation}
The dynamics of a multi-body system is naturally associated with a Riemannian manifold $(\mathcal{Q}, g)$ \cite{Eisenhart1928}, i.e., we equip the configuration manifold with a Riemannian metric $g$.
For mechanical systems, a natural choice for the metric is the inertia tensor of the system, which is given in a chart by $\MM(\qq)$.
We use the symbols $g_I$ and $g_M$ to denote the metrics, which are in coordinates given by $\II$ and $\MM(\qq)$, respectively.
Likewise, the metric $g_A$ denotes a metric that can be any Riemannian metric, including $g_I$ and $g_M$.\footnote{These choices of using the Euclidean and the inertia tensor as metric are not complete. Arbitrary custom metrics can be used here, e.g., the stiffness matrix for stiffness consistency \cite{Dietrich2015}.}

In the previous sections, we considered Euclidean configuration space, identifying the covectors with their vector counterparts (w.r.t. the Euclidean metric) and visualizing them accordingly.
While this is not very elegant from a differential geometric perspective, we think that the section provides valuable and comprehensive insights for roboticists.

From the differential geometric perspective, the rows of the Jacobian, being derivatives of scalar functions, are coordinate versions of differential one-forms (or covectors).
Consider a task space parametrized by task coordinates $\xx \in \mathcal{M}$ and a forward kinematics map $\xx = \hh(\qq)$. 
The differential one-form associated with each coordinate $x_i$ can be written in a chart using the Jacobian (e.g. \cite{Park1998}) \begin{equation}\label{eq:one-form-def}
    \d h_i = \sum\limits_{j=1}^{n}\pdx{h_i(\qq)}{q_j} \d q_j = \JJ_{x_i} \d\q.
\end{equation}
Taking a self-motion manifold $\Sc_0$ defined by $\hh(\qq)=\xx_{0}$ for an arbitrary constant $\xx_0$, and considering a velocity $\vv \in \Tc_q\Sc_0$, the reciprocal relation $\d h_i (\vv) = 0$ holds.
Application of a one-form to a vector is independent of any metric.

Moreover, having defined a metric $g_A$ on $\Qc$ naturally associates a vector to any covector. 
In particular, the gradient $\nabla_A h_i$ is the vector associated with the one-form $\d h_i$.
The gradient $\nabla_A h_i$ does not only depend on the function itself but also on the metric.
In coordinates, the gradient $\nabla_A h_i$ is obtained by multiplying the partial derivatives by the inverse metric tensor, an operation known as \emph{raising an index} in tensor calculus\footnote{In (\ref{eq:one-form-def}) and (\ref{eq:gradient-def}) we stick to the convention of representing contravariant vectors as columns and covectors as rows.}
\begin{equation}\label{eq:gradient-def}
    \nabla_A h_i = \AAA^{-1} \begin{bmatrix}
        \pdx{h_i(\qq)}{q_1} \\ \vdots \\
        \pdx{h_i(\qq)}{q_m}
    \end{bmatrix} = \AAA^{-1} \JJ_{x_i}\tran.
\end{equation}
An inverse operation associates a covector to a vector by \emph{lowering an index}. 
In our case this is $\JJ_{x_i}\tran= \AAA \nabla_A h_i$ in a chart.

Therefore, (\ref{eq:unified}) can be interpreted as applying one-forms to gradients.
We take the one-forms $\d h_i$ for $i\in\{1..m\}$, and apply them to the gradients $\nabla_{A}\xi_j$ for $j\in\{1..r\}$ and require them to vanish \begin{equation}\label{eq:apply-one-forms}
    \d h_{i}(\nabla_{A}\xi_{j}) = 0. 
\end{equation}
Due to symmetry of the metric, we can swap the roles in (\ref{eq:apply-one-forms})\begin{equation}\label{eq:main-condition-diffgeo}
    \d\xi_j(\nabla_{A}h_i) = 0.
\end{equation}

Again, the metric does not appear explicitly but is, of course, contained in the definition of the gradient. 
However, the same condition can be expressed in the form of vector orthogonality as $\nabla_{A}\tran h_i \AAA \nabla_{A}\xi_i= 0$ or in the form of covector orthogonality as $ \d h_i \AAA^{-1} \d \xi_i\tran = 0$.
This should clarify why we talk about orthogonality w.r.t. the metric $\AAA$ in \eqref{eq:unified}, although its inverse appears in the formula. 
From this perspective, it becomes clear that in Sec.~\ref{sec:geometric_intuition} and Fig.~\ref{fig:consistent_grid}, Fig.~\ref{fig:2dof-foliations-Jacobians}, and Fig.~\ref{fig:noninv} we actually visualize gradients w.r.t. either the metric $\II$ or $\MM(\qq)$. 

Gradients and one forms can be defined in a coordinate-independent way.
Therefore, (\ref{eq:main-condition-diffgeo}) can be thought of as a coordinate-free version of our main condition (\ref{eq:unified}).

To conclude under which conditions functions $\xi_j$ satisfying (\ref{eq:main-condition-diffgeo}) exist, we need to introduce the notion of an exterior differential system (EDS) \cite{McKay2020,Bryant2003}.
An EDS is a tuple $(\mathcal{Q}, \mathcal{I})$ of a manifold and an ideal of k-forms.
Integral elements of the EDS are linear subspaces of the tangent space of $\mathcal{Q}$ on which forms in $\mathcal{I}$ vanish \cite{McKay2020}.

Condition (\ref{eq:main-condition-diffgeo}) can be reinterpreted using EDS-language:
find an EDS $(\mathcal{Q}, \mathcal{I})$ such that the one-forms $\d\xi_j$ in $\mathcal{I}$ are the differentials of functions $\xi_j$ and such that the integral elements are spanned by $\nabla_{\AAA}h_i$.
This is opposite to classical EDS problems, where the ideal of k-forms $\mathcal{I}$ is given and the integral elements are unknowns.

Let $\Delta_h$ denote a distribution\footref{foootnote:distribution} of vector fields, which is locally spanned by \begin{equation}
    \Delta_h = \text{span} \{\nabla_{g_A}h_1, \ldots, \nabla_{g_A}h_m \}.
\end{equation} 
The distribution $\Delta_h$ will generally not be involutive~\cite{Klein1983}.

From the Frobenius theorem for EDSs \cite[Thm.~1.3]{McKay2020} we know that the integral manifolds of $\mathcal{I}$ form the leaves of a foliation of $\mathcal{Q}$ if and only if the vector fields on which all forms in $\mathcal{I}$ vanish are closed under the Lie bracket and thus involutive.
This is generally not the case for our redundancy resolution problem, as the distribution $\Delta_h$ is generally not involutive.
However, it will be trivially involutive, if one has a one-dimensional task space and a single task coordinate, as a one-dimensional distribution is always involutive. 
We now take the Frobenius theorem and replace the geometrical vocabulary with their respective robotics terms:

\begin{theorem}[Existence of Orthogonal Foliations]\label{thm:existence}
    A task-induced foliation admits orthogonal foliations if and only if the distribution formed by the gradients of the task coordinates is involutive.
\end{theorem}
\begin{corollar}
A one-dimensional task space always admits orthogonal foliations.
\end{corollar}

Note that the above statements are coordinate-invariant and depend only on the forward kinematics map. 
Indeed, taking different coordinate charts on the task manifolds leads to different gradients, but they all span the same space.

Take, for instance, a three-dimensional configuration space and a single task coordinate. 
Two orthogonal foliations $\xi_1$ and $\xi_2$ must be determined in this case. 
This kind of problem has been studied in some publications, where it is known as \emph{triply orthogonal system} \cite{Bobenko2003} or \emph{triply orthogonal web} \cite{McKay2020}.
However, these publications ask that the foliations $\xi_1$ and $\xi_2$ are also orthogonal. 
This is a condition we relax in this work, as one usually cannot expect it to be satisfied.

Theorem~\ref{thm:existence} provides a necessary and sufficient condition for the task-induced foliation to admit orthogonal foliations.
Throughout the paper, we will deal with cases where this condition is satisfied and cases where it is not.
In the latter case, we find an approximate solution by relaxing the conditions and optimizing a cost function.

\section{Local Solution: Plane Stacks}\label{sec:plane_stacks}
Before starting a more rigorous study on generating orthogonal foliations $\xxi(\qq)$, we start with a simple baseline version: we use linear, locally orthogonal foliations.
In particular, we will approximate each orthogonal coordinate $\xi_j(\qq)$ by a function, which is linear in $\qq$.

We construct a foliation where the leaves are planes.
Consider a configuration $\q_0 \in \mathcal{Q}$, which we call the \emph{base configuration for the locally orthogonal foliations} and also consider a hyperplane $E_0 = E(\q_0, \b{n}_0)$ with support vector $\q_0$ and unit normal vector $\b{n}_0$.
The signed distance of a query point $\q$ to the hyperplane can be computed using the Hesse normal form of the plane \begin{equation}\label{eq:hesse_normal_form}
	d(E_0, \q) = \b{n}_0\tran (\q - \q_0)
\end{equation}
This signed distance function can also be interpreted as a coordinate function with a trivial Jacobian being the covector version of $\b{n}_0$.

Equation (\ref{eq:hesse_normal_form}) also creates a foliation for which each leaf $E_n = E(\qq_n, \b{n}_0)$ is a hyperplane of constant distance and thus parallel to the base hyperplane $E_0$. 
Locally, around $\q_0$, the function $\xi_j(\qq) = d(E_0, \qq)$ provides approximate orthogonal coordinates if the normal vector $\b{n}_0$ is chosen appropriately.
For each locally orthogonal foliation $\xi_j(\qq)$, we require a unit normal vector $\nn_j$ such that they are mutually linearly independent.
Additionally, we require the normal vectors $\nn_j$ to span the orthogonal complement of the rows of $\JJ_x(\qq_0)$ for Euclidean metric or of the columns of $\MM^{-1}(\qq_0)\JJ_x\tran(\qq_0)$ for inertia metric.
In other words, we can say that the unit vectors associated with the functions $\xi_j(\qq)$ span the null space of $\AAA^{-1}(\qq_0)\JJ_x(\qq_0)$.

The entirety of locally orthogonal coordinates can be written in compact form \begin{equation}\label{eq:locally-orthogonal-coordinates}
	\xxi(\qq) = \underbrace{\begin{bmatrix}
		\nn_1&
		\ldots&
		\nn_r
	\end{bmatrix}\tran}_{\JJ_\xi} (\qq - \qq_0).
\end{equation}
The Jacobian $\JJ_\xi$ of $\xxi(\qq)$ is constant. 
Based on this construction, we ensure that (\ref{eq:unified}) is satisfied at $\qq_0$: at $\qq_0$, the orthogonal coordinates $\xxi(\qq)$ are dynamically decoupled from the task.
The performance on the regions around $\qq_0$ will depend on the curvature of the self-motion manifolds and the metric.
For configurations far from the base configuration, the simple local solution by plane stacks (\ref{eq:locally-orthogonal-coordinates}) cannot be expected to perform well --- let's move to more elaborate and global solutions.

\section{Exact Solution: Coordinate Growing}\label{sec:growing}
In this section, we show how an orthogonal foliation can be numerically generated for the case of one-dimensional task space.
To perform the numerical propagation of the orthogonal foliation coordinates, we start by selecting one value for the task-coordinate function and call it $x_0$.
Then, we compute the corresponding self-motion manifold $\Sc_0$ such that
\begin{equation}
	\forall \q \in \Sc_0: \hh(\qq) = \xx_0.
\end{equation}
We call $\Sc_0$ the \emph{base manifold}.
Note that $\dim \Sc_0 = n-1$.
Then, we define a chart $\xxi_0: \Sc_0 \rightarrow \mathbb{R}^{n-1}$ on the base manifold.

We determine the coordinates $\xxi(\q_1)$ of a point $\q_1 \in \Sc_1$ on another leaf $\Sc_1$ by taking the differential equation \begin{equation}\label{eq:gradient_flow}
	\dot{\q} = (\xx_0 - \hh(\qq))\nabla_{\AAA}\hh(\q)
\end{equation}
with initial condition $\q(0) = \q_1$.
The flow (\ref{eq:gradient_flow}) will converge onto the base manifold at a point $\hat{\q}\in\Sc_0$.
We then set $\xxi(\q_1) = \xxi_0(\hat{\q})$ to obtain coordinates of $\q_1 \in \Sc_1$ in the propagated chart.

In the following sections, we investigate the resulting coordinates for the two example robots.
We show results for Euclidean (kinematically decoupled) and inertia tensor (dynamically decoupled) as metric.

The point of convergence of the gradient flow (\ref{eq:gradient_flow}) is numerically evaluated using a Runge-Kutta integration scheme \cite{Hairer2008}.
We stop the integration when the change of $\q$ between two integrator steps falls below a threshold of $10^{-6}$.

\subsection{In Two-Dimensional Configuration Space}\label{sec:growing_2d}
Take the example manipulator with two degrees of freedom (Fig.\ref{fig:exrobo2}) and select the $x$-component of the end-effector position as task coordinate.

\subsubsection{Euclidean Metric (\textbf{A} = \textbf{I})}
\begin{figure}
	\centering
    \subf{In $q_1q_2$-plane\label{fig:n2_integrated_plane}}{\w}{\footnotesize}{n2_m1_integrated_isolines}\\
    \subfloat[On the 2-torus\label{fig:n2_integrated_torus}]{
    	\includegraphics[width=.8\w]{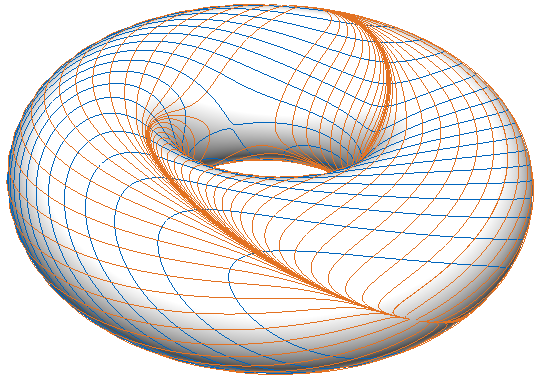}
    }
    \caption{Following the gradient flow of the $x$-coordinates for Euclidean metric. 
    }
    \label{fig:n2_integrated_isolines}
\end{figure}

We begin with the choice $\AAA = \II$, i.e., we choose the standard Euclidean metric on $\mathcal{Q}$.
Fig.~\ref{fig:n2_integrated_plane} shows isolines of the forward kinematics in blue.

Take a grid of points in the configuration space denoted by the gray crosses.
We take the gradient flow $\pm\nabla_{\b{I}}\hh(\qq)$ and integrate it into both directions for each point of the grid.
The orange lines in Fig.~\ref{fig:n2_integrated_plane} show the resulting integral curves.
Dependent on the sign, the flow converges to one of the singularities.
Note the toroidal topology of the configuration space.
Especially, the points of convergence on both sides of the plot are the same point on the torus (Fig.~\ref{fig:n2_integrated_torus})!
The entirety of integral curves \emph{foliates} the configuration space $\mathcal{Q}$.

Pick a base manifold $\Sc_0$ such that $x_0 = 1$ (black curve in Fig.~\ref{fig:n2_integrated_plane}).
The base manifold is homeomorphic to the 1-sphere $\mathbb{S}^1$: we cannot find a global diffeomorphism onto the real numbers $\xi_0: \mathcal{S}_0 \rightarrow \mathbb{R}$.
Instead, we can find a smooth function $\hat{\xi}_0: \mathcal{S}_0 \rightarrow \mathcal{SO}(2)$.
The diffeomorphism $\hat{\xi}_0$ must be used if the entire base manifold $\Sc_0$ shall be covered.
We apply a simplification here and map the entire manifold $\Sc_0$ but one point to $(-\pi, \pi]$ to stay within a single coordinate chart.
The drawback is that $\xi_0$ fails at $\pm\pi$, and we cannot pass through that point.

Finally, we evaluate the resulting coordinate function $\xi: \mathcal{Q} \rightarrow (-\pi, \pi]$ by pulling a point $\q \in \Qc$ onto the base manifold (\ref{eq:gradient_flow}) and evaluating $\xi_0$ at that point.
This procedure is visualized for two example configurations in Fig.~\ref{fig:n2_integrated_isolines}. 
When starting from arbitrary joint configurations (black crosses), the flow will trace out a path (red lines) and converge at a configuration on the base manifold (violet crosses).
Fig.~\ref{fig:n2_integrated_color} shows the function color-coded.

The function is smooth in large regions of Fig.~\ref{fig:n2_integrated_color}.
However, there are also edges of jumps in the color rendering of $\xi$.
Consider the edge at the origin $\q = [0, 0]\tran$ of the plot.
The function $\xi$ becomes increasingly steep when moving on this edge towards the center and discontinuous directly at the origin.
Considering the toroidal structure of the space and solving the puzzle (top-right corner of Fig.~\ref{fig:n2_integrated_color}) shows that the discontinuous spot coincides with a task-space singularity.
As before, we assume we never reach singular configurations and forget about the singularity at the center for now.
Within the white region of Fig.~\ref{fig:n2_integrated_color} is another discontinuous spot, an artifact of the mapping of $\mathbb{S}^1$ onto $\mathbb{R}$.

\subsubsection{Mass Metric (\textbf{A} = \textbf{M})}
We choose the same chart $\xi_0$ on $\Sc_0$: the value of $\xi$ on $\Sc_0$ does not change at all!
What does change are the values of points in $\Qc\setminus\Sc_0$.
Compared to the Euclidean case, the coordinates on $\Sc_0$ grow in different directions as the gradient in (\ref{eq:gradient_flow}) depends on the metric.

\begin{figure}[h!]
	\centering
    \subf{In $q_1q_2$-plane}{\w}{\footnotesize}{n2_m1_integrated_isolines_mass}
    \hfill
    \subfloat[On the 2-torus]{
    	\includegraphics[width=.8\w]{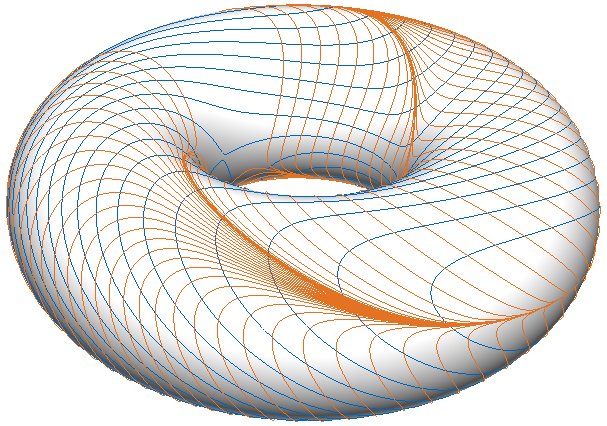}
    }
    \caption{Following the gradient flow of the $x$-coordinate for inertia tensor as metric ($\AAA = \MM)$.}
    \label{fig:n2_integrated_isolines_mass}
\end{figure}

\begin{figure*}
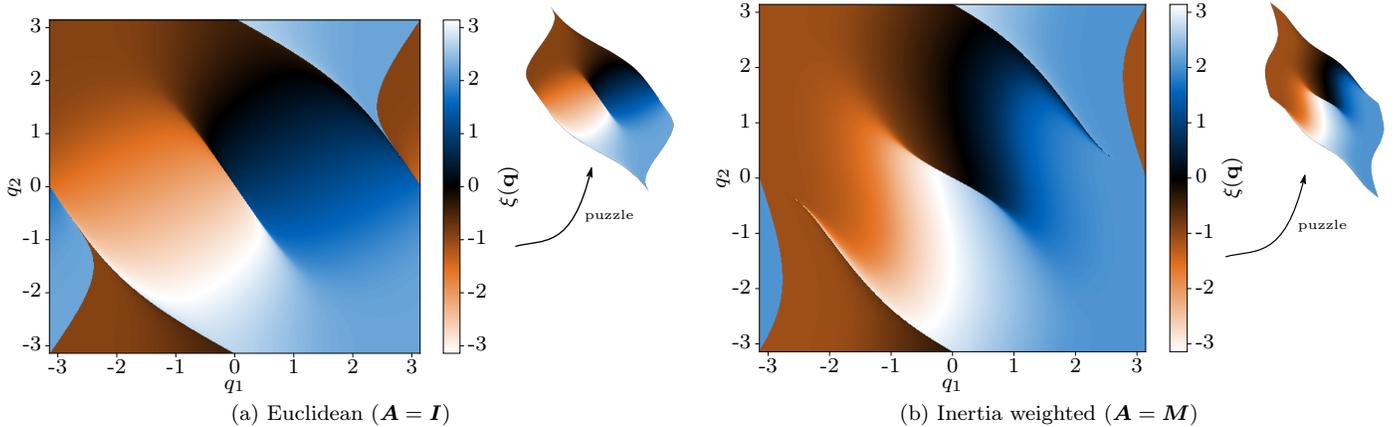

	\centering
	\subf{Euclidean $(\AAA = \II)$\label{fig:n2_integrated_color}}{1.1\w}{\footnotesize}{n2_m1_colorcode}\hfill
	\subf{Inertia weighted $(\AAA = \MM)$\label{fig:n2_integrated_mass_color}}{1.1\w}{\footnotesize}{n2_m1_integrated_isolines_mass_color}
    \caption{Orthogonal coordinate function $\xi(\q)$ for Euclidean metric (left) and non-Euclidean metric (right). The color encodes the value of the function $\xi(\qq)$. Sketches on the top-right: most discontinuities can be pushed to the boundary when considering the toroidal structure.}
\end{figure*}

Fig.~\ref{fig:n2_integrated_isolines_mass} shows integral curves of $\pm\nabla_{\MM}\hh(\qq)$ (orange), as well as the contour lines of the task coordinate (blue) and the grid of start points $\q_0$ (gray crosses) for the integration and Fig.~\ref{fig:n2_integrated_mass_color} shows the coordinate function color-coded.

\subsection{In Three-Dimensional Configuration Space}
Next, we consider $n=3$ degrees of freedom and choose the $x$-component of the end-effector pose as task coordinate.
Hence, we obtain a $r=2$-dimensional task-induced foliation of the configuration space.

We again select a base manifold $\mathcal{S}_0$ and define a coordinate chart on it (cmp. Fig.~\ref{fig:consistent_grid}).
Parametrization of a surface embedded in three-dimensional space is a well-studied problem within the field of computer graphics~\cite{Floater2005}, where this is required for, among others, texture mapping.
We consider the simplest case of parametrization: a surface of disk-topology.
This does not involve the need for cuts of the surface for parametrization.

\ifx\ieee\undefined
Within the computer graphics field the desired object is a parametrization $\Phi: \mathcal{X} \rightarrow \mathcal{S}_0$, where $\mathcal{X} \subset \mathbb{R}^2$ is a region in the parameter space.
A mapping $\Phi$ is one-to-one if \begin{enumerate}
	\item $\Phi$ maps the boundary $\partial\mathcal{S}_0$ of the surface to the boundary $\partial\mathcal{X}$ of a convex region $\mathcal{X} \subseteq \mathbb{R}^2$ and
	\item $\Phi$ is harmonic. \cite{Rado1926,Kneser1926,Choquet1945}.
\end{enumerate}
Within this context \emph{harmonic} refers to the property $\Delta \Phi = 0$, i.e., the application of the Laplace-Beltrami operator to $\Phi$ vanishes everywhere \cite{Axler2001}.
The parametrization function $\Phi$ is the inverse of our desired chart of the base manifold $\xi_0 = \Phi^{-1}$.
\fi

\begin{figure}[h!]
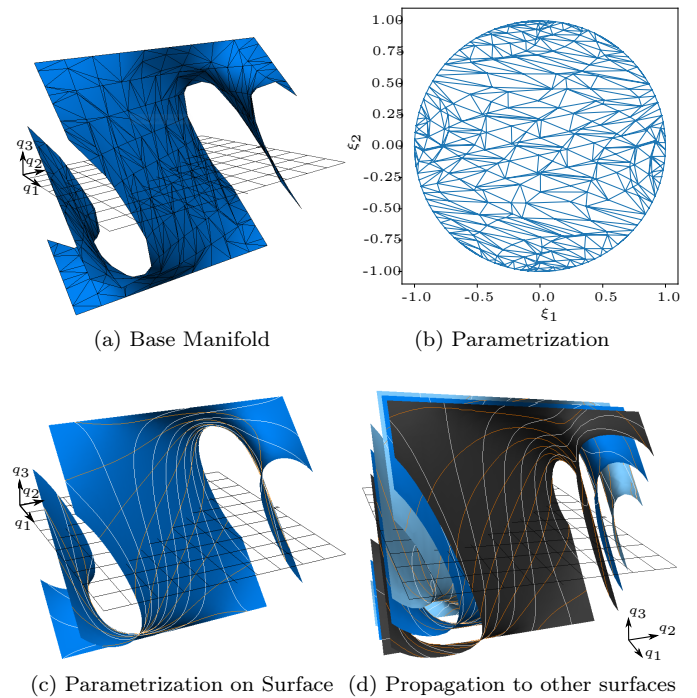

	\centering
	\subf{Base Manifold\label{fig:s0_mesh}}{.24\textwidth}{\tiny}{base_manifold}
	\subf{Parametrization\label{fig:param_domain}}{.24\textwidth}{\tiny}{parametrization_s0}\\
	\subf{Parametrization on Surface\label{fig:surface_parametrized}}{.24\textwidth}{\tiny}{parametrized_s0}
	\subf{Propagation to other surfaces\label{fig:ordered_spaghetti}}{.24\textwidth}{\tiny}{growing_multiple_shells}
	\caption{
		Coordinate growing from a base manifold $\mathcal{S}_0$. 
		\textbf{(a)} Triangulation of the base manifold shown for a coarse resolution. 
		\textbf{(b)} Parametrization of the base manifold onto the unit circle. 
		The edges do not intersect, and we obtain a bijection between the unit circle and the base manifold. 
		\textbf{(c)} Parametrization shown on the base manifold using isolines of the parametrization. 
		\textbf{(d)} Propagation to other leaves of the foliation.
	}
	\label{fig:parametrization_s0}
\end{figure}

We select $x_0 = 0$ and constrain $q_1 \geqslant 0$, i.e., we consider a subregion of the configuration space $\Qc$, and the simplified model generated here will only be valid in this subregion.
By this constraint, we achieve a disk topology of $\mathcal{S}_0$.
Further, we apply marching cubes \cite{Lewiner2003} and obtain a triangle mesh approximating the base manifold $\mathcal{S}_0$ (Fig.~\ref{fig:s0_mesh}).
Because of the mesh structure, we can use discrete differential geometry to rewrite the continuous constraints for $\Phi$ in a discrete fashion, and the condition boils down to solving a linear system of equations as summarized in \ifx\ieee\undefined Sec.~\ref{app:harmonic}. \else \cite{Floater2005}. \fi
Boundary conditions are required for the linear system, which we generate by mapping the boundary of the base manifold to the unit circle.
Afterward, we solve for the interior vertices and obtain a discrete version of a bijective mapping between the base manifold and the unit disk in the plane (Fig.~\ref{fig:param_domain}).
Fig.~\ref{fig:surface_parametrized} additionally shows isolines of the parameter space on the base manifold.
The visualizations are for a coarse triangle mesh resolution to visualize the mesh structure in the parametrization domain.

Following the above procedure, we obtain a discrete version of $\xxi_0(\q)$ by having a unique value $\xxi_0(\q)$ for every mesh vertex.
For every other point $\q \in \Sc_0$, not coinciding with a vertex of the mesh, we query a KDTree~\cite{Maneewongvatana1999} generated from the mesh to find the face containing $\q$ (approximately) and interpolate between the values for $\xxi_0$ of the three vertices of the face.
Then, we follow the gradient flow (\ref{eq:gradient_flow}) to pull any configuration $\qq\in\mathcal{Q}$ onto the base manifold $\hat{\q} \in \Sc_0$ and interpolate $\xxi_0({\hat{\q}})$ as stated above.
We have achieved a coordinate function $\xxi: \bar{\Qc} \rightarrow \mathbb{R}^2$.

Fig.~\ref{fig:ordered_spaghetti} shows the result for $\AAA = \II$.
We show three self-motion manifolds and the isolines of the coordinate functions $\xi_1$ and $\xi_2$.
The dark blue manifold in the middle of the stack is the base manifold $\Sc$ at $x_0 = 0$, while the other manifolds (gray and light blue) correspond to $x = \pm 0.3$.

\section{Optimization Solution: Neural Network}\label{sec:nn}
The previous section showed how a coordinate function defined on one of the self-motion manifolds could be propagated to other points in the configuration space in the case of only one task coordinate.
This approach has three drawbacks: (1) the algorithm is rather expensive in execution time, (2) there is no easy way to obtain a Jacobian $\JJ_\xi$ besides the obvious numerical Jacobian, and (3) it only works for one-dimensional tasks.
For a higher dimensional task space, there are multiple gradient flows (\ref{eq:gradient_flow}), one for each row of the Jacobian.
Therefore, there are multiple directions in which the coordinates of the base manifold would need to be propagated. For the same approach to work for higher dimensional task spaces, all the flows (\ref{eq:gradient_flow}) starting from a point  $\q_1 \in \Sc_1$ would need to end at the same point in $\q_0 \in \Sc_0$.
This is, in general, not possible due to non-involutivity.

We relax the orthogonality condition to \emph{as orthogonal as possible}, use a smooth parametric function $\xxi_\theta(\qq)$, and optimize the parameters $\ttheta$ such that the resulting foliation is \emph{quasi-orthogonal}.
Another benefit of this approach is that it allows for analytical Jacobians.

Neural networks provide a flexible and universal function approximator.
By formulating partial or ordinary differential equations as loss functions, machine learning techniques can be used to find solutions~\cite{Graepel2003,Sachtler2022}.
We write the parametric function $\xxi_\theta(\qq)$ as neural network with $n$ input and $r$ output neurons.
In particular, we use a network with two hidden layers and $\tanh$ activation functions
\begin{equation}\nonumber
	\xxi_\theta(\qq) = \b{W}_{\mathrm{out}} \tanh \left[\b{W}_2 \tanh\left(\b{W}_{1}\begin{smallmatrix}\sin\\\cos\end{smallmatrix}(\qq) + \b{b}_{1} \right) + \b{b}_2 \right] + \b{b}_{\mathrm{out}},
\end{equation}
where $\tanh$ acts componentwisely and $\ttheta$ is the collection of all elements of $\b{W}_i$ and $\b{b}_i$.
The function $\begin{smallmatrix}\sin\\\cos\end{smallmatrix}: \mathbb{R}^n \rightarrow \mathbb{R}^{2n}$ stacks the sines and cosines of the elements of the input vector.
Using this transformation, we implement the toroidal structure of the input space.

Within this paper, we show results for robots with revolute joints only. However, the framework and the concepts are not restricted to purely revolute robots. The methods can also be used for robots having prismatic joints. In that case, the $\sin$ and $\cos$ terms will be omitted for prismatic joints.
\ifx\ieee\undefined
Fig.~\ref{fig:neural_net} visualizes the neural network model.

\begin{figure}
	\centering
	\def\svgwidth{.5\textwidth}\footnotesize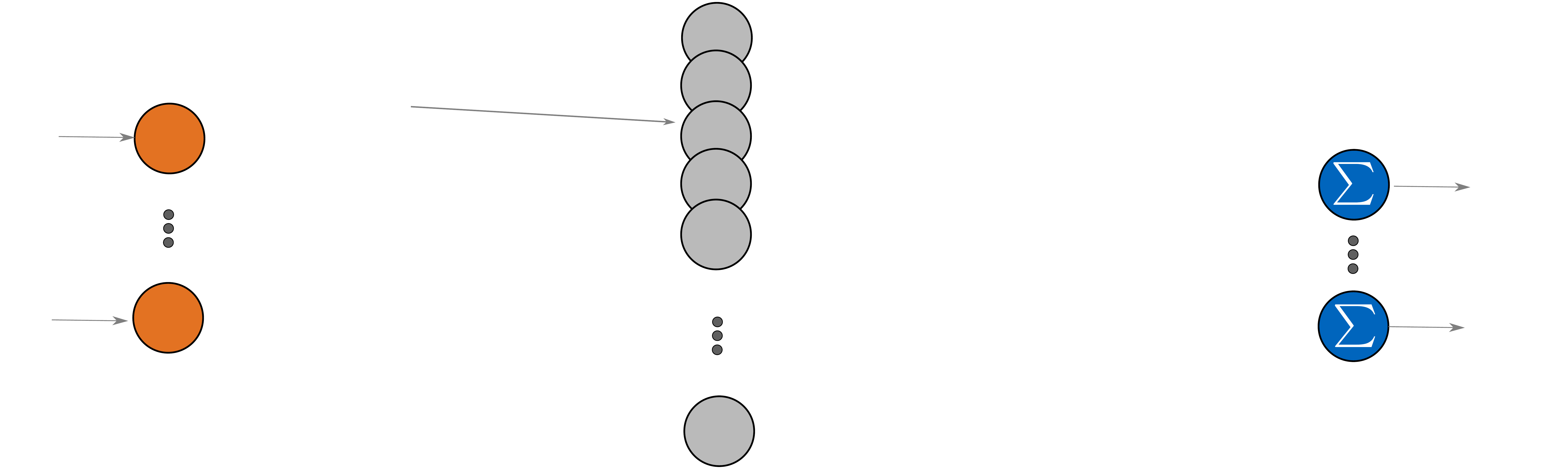
	\caption{
		Neural network with two hidden layers. 
		The outputs are the coordinate functions $\xi_i$. 
		Before feeding the inputs $q_i$ to the actual network we apply the sine and cosine of each input value in order to implement the toroidal structure of the input space.
	}
	\label{fig:neural_net}
\end{figure}
\fi

Differentiating the network output with respect to its input yields the Jacobian $\JJ_{\xi,\theta}(\qq) = \frac{\partial \xxi_\theta}{\partial \qq}$.
The expression $\frac{\partial \JJ_{\xi,\theta}(\qq)}{\partial \ttheta}$ describes the effect of the parameters $\ttheta$ on the input-output Jacobian of the network.
All the network parameters $\ttheta$ except the output layer's biases affect the Jacobian $\JJ_{\xi,\theta}(\qq)$.

Condition (\ref{eq:unified}) is used to derive a scalar cost function that penalizes the cosine of the pairs $\nabla_{\AAA}h_i$ and $\d\xi_{\theta,j}$.
We write this in matrix-vector notation and introduce $\vv_{x,i}\tran(\qq)$ being the $i$-th row in $\JJ_x(\qq)$ and $\vv_{\xi,j}\tran(\qq)$ being the $j$-th row of $\JJ_{\xi,\theta}(\qq)$: 
\begin{align}\label{eq:loss_one_sample}
    L(\q, \ttheta) &= \frac{\lambda_1}{2mr}\sum\limits_{i=1}^{m} \sum\limits_{j=1}^{r} \left(\frac{\vv_{x,i}\tran \b{A}^{-1} \b{v}_{\xi,j}}{|\b{A}^{-1} \b{v}_{x,i}| \cdot |\b{v}_{\xi,j}|}\right)^2 \\&+ \frac{\lambda_2}{2\binom{r}{2}}\sum\limits_{i=1}^{r} \sum\limits_{j=1}^{r} \left[\left(\frac{\b{v}_{\xi,i}\tran \b{A}^{-1} \b{v}_{\xi,j}}{|\b{A}^{-1}\b{v}_{\xi,i}| \cdot |\b{v}_{\xi,j}|}\right)^2[i > j]\right]\nonumber,
\end{align}
where $[b]$ is the Iverson bracket, which is $1$ when the $b$ is true and $0$ otherwise, and $\binom{n}{k}$ the binomial coefficient.

The first part of the cost function penalizes the square of the cosine between mutual rows of $\JJ_x\AAA^{-1}$ and $\JJ_{\xi,\theta}$, while the second part penalizes the cosine between two mutual rows of $\JJ_{\xi,\theta}\AAA^{-1}$ and $\JJ_{\xi,\theta}$.
Those two terms can be weighted using the constants $\lambda_1$ and $\lambda_2$.
The second part of the cost function resembles the full-rank condition.
By using the cosines of the angles between the gradients, we ensure that the magnitude of the gradients is not penalized.
This is required because the globally optimal solution would be $\xxi(\qq) = \text{const}$ otherwise.

We chose $\lambda_1 = 1000$ and $\lambda_2 = 1$, i.e., we assign much higher priority to the decoupling between task- and orthogonal coordinates than mutual decoupling between the orthogonal coordinates.
Generally, when not allowing constant functions for $\xi_i(\qq)$, (\ref{eq:loss_one_sample}) cannot be optimized to zero.
Consider again the 3 DoF manipulator with 1 DoF task space:
optimizing (\ref{eq:loss_one_sample}) to zero would imply that we have found a triply orthogonal system.
This, is only possible in exceptional cases for specific task foliations~\cite{Bobenko2003,McKay2020}.
Therefore, optimizing (\ref{eq:loss_one_sample}) will find a trade-off between the two terms.

Suppose we have $K$ samples $\qq_k$ of the selected subset of $\Qc$. The total cost will be \begin{equation}\label{eq:cost}
	L(\ttheta) = \frac{1}{K}\sum\limits_{k=1}^{K} L(\q_k, \ttheta).
\end{equation}
We sample the training data on the selected regions of the joint space $\bar{\Qc}$.

Finally, we perform gradient descent on $L(\ttheta)$.
By construction the function $\xxi_\theta$ is always an integral of $\JJ_{\xi,\theta}$, so we get the coordinate function $\xxi_\theta(\qq)$ for free when fitting the Jacobian $\JJ_{\xi,\theta}(\qq)$ to the condition (\ref{eq:unified}).
The optimization simultaneously finds a set of vector fields as well as an integral of them.

The solution $\xxi_\theta(\qq)$ to (\ref{eq:unified}) is not unique and will depend on the initialization of the parameters.
After initialization, we optimize on the cost function (\ref{eq:cost}).
This can be interpreted as bending and warping the randomly initialized foliation.

The training procedure is split into multiple epochs.
We sample new training data in each epoch using a uniform distribution on $\Qc$ to prevent overfitting.
Some further details on the training procedure can be found in~\cite{Sachtler2020}.
For the implementation, we use TensorFlow~\cite{Abadi2015}, which allows for evaluating analytical gradients and Jacobians and generating fast code running on the GPU.
We denote the optimized parameters $\hat \ttheta$.

\subsection{Two Degrees of Freedom Example Revisited}\label{sec:nn_two_dof}
Let us revisit the two degrees of freedom example from Sec.~\ref{sec:growing_2d}.
The planar configuration space allows visualizing the results straightforwardly.
Globally, no smooth function $\xxi(\qq)$ creates a non-singular parametrization of the task-induced foliation everywhere because it involves parametrizing a 1-sphere with a single coordinate.
Still, we train the model globally, i.e. we sample training data from the entire $\Qc$ and check how the network handles this issue.

\begin{figure}[h!]
	\centering
	\subfloat{\def\svgwidth{1.1\w}\footnotesize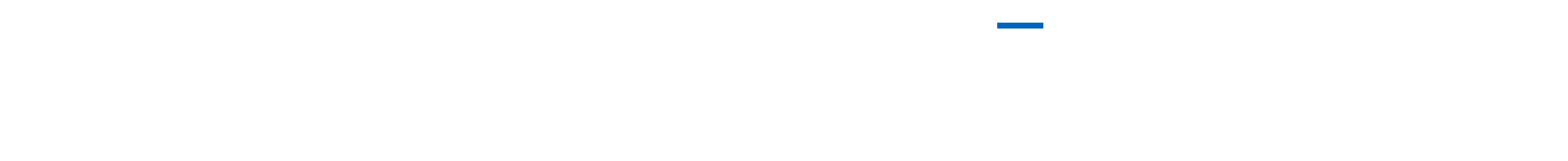}\\\vspace{-.2cm}
	\setcounter{subfigure}{0}
	\subf{Euclidean ($\AAA = \II$)\label{fig:nn_n2_eucl}}{.55\w}{\tiny}{res_n2_m1}
	\subf{inertia tensor ($\AAA = \MM$)\label{fig:nn_n2_noneucl}}{.55\w}{\tiny}{res_n2_m1_dyn}
    \caption{Resulting contour lines of the trained neural network for different metrics. The lines show isolines of the task coordinate $x$ (blue) and orthogonal coordinate (orange). Solid lines refer to positive values and dashed lines show negative values. The background shows the local training accuracy.}\label{fig:nn_n2}
\end{figure}

\begin{figure}
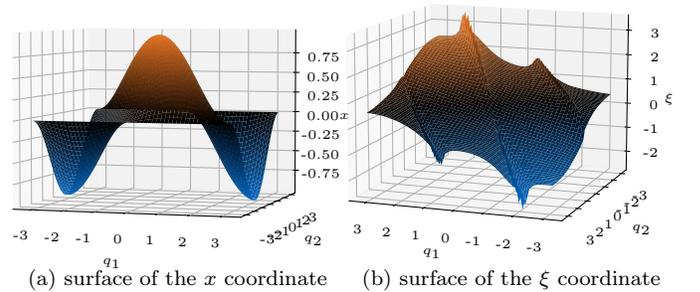

	\centering
	\subf{surface of the $x$ coordinate\label{fig:surface_n2_x}}{.55\w}{\tiny}{two_dof_x}
	\subf{surface of the $\xi$ coordinate\label{fig:surface_n2_xi}}{.55\w}{\tiny}{two_dof_xi}
	\caption{
		Surface plots of the the task coordinate function $x$ (left) and the neural network approximation of a orthogonal coordinate $\xi$ (right) for a two degrees of freedom manipulator for the Euclidean case ($\AAA = \II$).}
	\label{fig:res_n2_m1_mesh}
\end{figure}

We first fit the neural network using $\AAA = \II$ and evaluate isolines (orange) of the resulting coordinate function $\xxi_{\hat{\theta}}(\q)$ in Fig.~\ref{fig:nn_n2_eucl}. 
Isolines of the forward kinematics function $h(\q)$ are shown in blue for reference.
The background color of the plot encodes the angle between $\nabla_{\AAA}h$ and $\d\xi$.
If the background was white everywhere, the training would be optimal.
However, as expected, there are certain regions where the model does not perform well visualized by the colored regions in Fig.~\ref{fig:nn_n2}.
We observe that each value exists twice on each self-motion manifold, as each isoline of $\xi(\qq)$ intersects each self-motion manifold twice.
As expected we did not obtain a global coordinate function on the self-motion manifolds.
Fig.~\ref{fig:surface_n2_xi} additionally shows the functions $h(\qq)$ and $\xi(\qq)$ as surface plot.

Fig.~\ref{fig:nn_n2_noneucl} shows the results for $\AAA = \MM$.
Similarly to the Euclidean case, the model performs well in large regions (white background color), and the function only uniquely parametrizes half of the self-motion manifolds. 

Compare the results obtained from the trained coordinate function and the numerical results from Sec.~\ref{sec:growing_2d}. 
Most importantly, we observe that the isolines of the trained function match the lines generated by the flow $\nabla_{\AAA}h$ to a large extent.
However, the similarity worsens when reaching the attracting lines growing out of the singular configurations.
The trained coordinate function $\xi_{\hat{\theta}}$ is, in contrast to the numerical results, smooth.
Therefore, $\xi_{\hat{\theta}}$ can never be a global chart of the self-motion manifolds, which we also observe in the results.

\subsection{Non-Involutive Case}\label{sec:nn_noninv_two_dof}
In the last section, we considered an example having an exact solution; now, a model with no exact solution, i.e., the distribution of gradients of the task coordinates, is non-involutive.
The numerical propagation of coordinates according to Sec.~\ref{sec:growing} fails, as the gradients of the task coordinates cannot be integrated to submanifolds of $\mathcal{Q}$.

Still, we can fit coordinate functions using the cost function (\ref{eq:cost}) and aim at finding a $\xi_\theta$ which is \emph{as orthogonal as possible}.

We consider again the manipulator with three degrees of freedom (Fig.~\ref{fig:exrobo3}) and select both, the $x$- and $y$-component of the end effector position as task coordinates. 
Therefore, we obtain a non-involutive system and train a neural network model using the cost function (\ref{eq:cost}).

\begin{figure}
    \includegraphics[width=.23\textwidth]{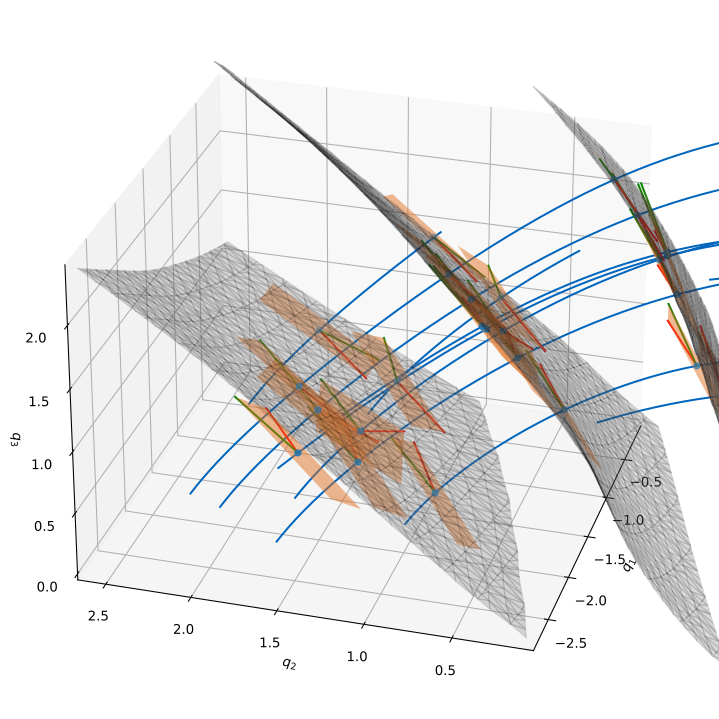}
    \includegraphics[width=.23\textwidth]{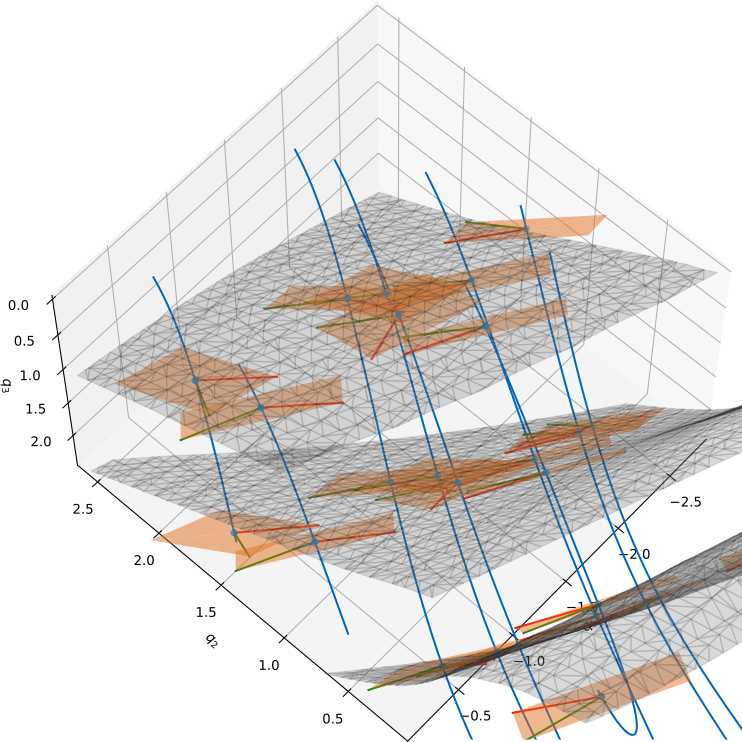}
    \caption{
	    Results for a subset of the joint space of a three DoF robot with two task coordinates. 
	    The blue lines show leaves of the task-induced foliation (self-motion manifolds) and the gray surfaces correspond to isosurfaces of the coordinate function the neural net found.}
    \label{fig:plaettchen}
\end{figure}

The results are shown in Fig.~\ref{fig:plaettchen} for a section of the configuration space.
The blue lines show some leaves of the task-induced foliations, i.e., self-motion manifolds.
Further, the red lines show $\nabla_{\II}h_1$, and the green lines show $\nabla_{\II}h_2$ scaled to unit length at that point.
Each pair of $\nabla_{\II}h_1$ and $\nabla_{\II}h_2$ spans a plane visualized by the red rectangles. The planes belong to the distribution which we aim to integrate - an integral foliation for the distribution would have those planes as tangent planes.
Finally, the gray surfaces correspond to isosurfaces of the foliation obtained by the learning process.
These are surfaces where $\xxi(\qq)$ is constant.
We observe that, despite the distribution not being involutive, the red planes are to a good approximation tangent to the isosurfaces. 

\begin{figure}[h!]
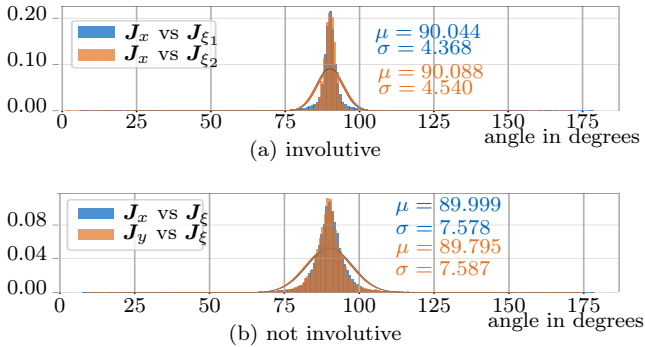

	\centering
	\subf{involutive\label{fig:hist_inv}}{\w}{\footnotesize}{hist_inv_ls}\\
	\subf{not involutive\label{fig:hist_noninv}}{\w}{\footnotesize}{hist_noninv_ls}
	\caption{
		Histogram of angles between gradients of the task- and orthogonal coordinates for randomly sampled configurations for a trained neural network.
		Top: involutive example, bottom: non-involutive example.
		The symbol $\mu$ denotes the mean and $\sigma$ the standard deviation.
	}
	\label{fig:angle_histogram}
\end{figure}

To quantify the effect of the non-involutivity, we also train a neural model for an involutive case.
This is achieved by removing the $y$-component of the end-effector position from the task coordinates.
We uniformly sample $N=10^5$ joint configurations and compute the angle between the gradients.
Fig.~\ref{fig:hist_inv} shows the histogram of the residual angles between $\d\xi_j$ and $\nabla_{\MM}h_i$.
The results can be compared to the histogram in Fig.~\ref{fig:hist_noninv}, which shows the results for the non-involutive case.
The histogram is, as expected, much more needle-shaped in Fig.~\ref{fig:hist_inv} than in Fig.~\ref{fig:hist_noninv}, and the standard deviations of the angles are lower for the involutive case.

\section{Simulation}\label{sec:sim}
We use the 3 DoF example manipulator (Fig.~\ref{fig:exrobo3}) for experimental validation.
For the dynamics, we assume a uniformly distributed mass of $m=3$ for each link.

To employ the controller (\ref{eq:total_control}) a desired trajectory in terms of $(\vphi_d, \vphidot_d, \vphiddot_d)$ is required.
We use a signal of pre-defined jumps $\hat{\vphi}(t)$ on which we apply a second-order low pass with natural frequency $\omega_0 = 2\pi$ and damping ratio $\zeta = 0.7$.
This way, we obtain a twice-differentiable signal that can be used as the desired trajectory.
\ifx\ieee\undefined%
We simulate the dynamics of the system (\ref{eq:multi-body-dyn}) in closed loop with the controllers using a Runge-Kutta integration scheme.
Fig.~\ref{fig:ff_and_fb} shows a block diagram of the simulated system.

\begin{figure*}
 	\centering
 	\def\svgwidth{2\w}
 	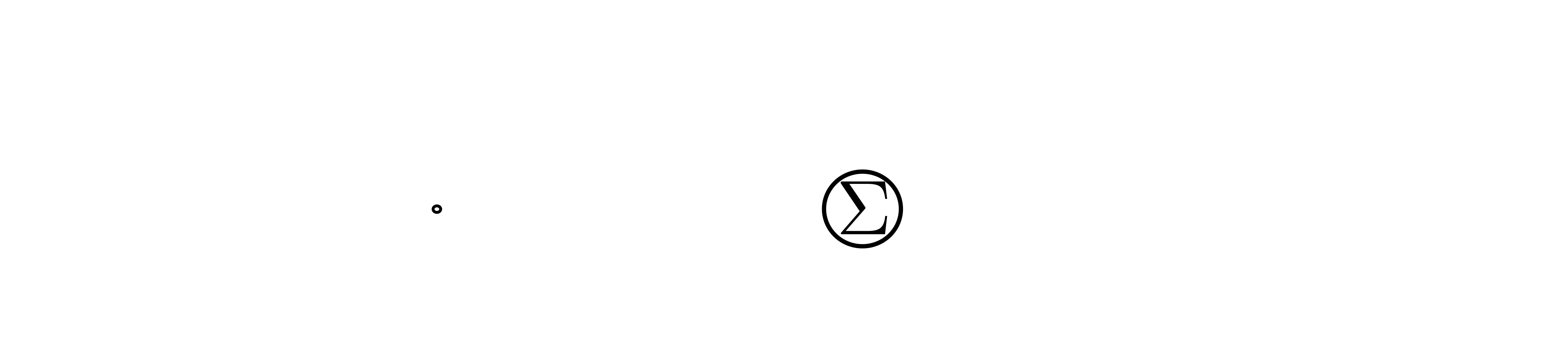
 	\caption{
 		Setting used for the experiments. 
 		We feed a discontinuous jump signal to a second-order low pass in order to generate a twice-differentiable signal. 
 		Then, the signal is fed to a feed-forward controller and a feedback controller. 
 		The controllers operate in augmented coordinates, which is equivalent to summing up controllers of the individual sets of coordinates for our choice of coordinates.}
 	\label{fig:ff_and_fb}
\end{figure*}
\fi
In the following, we show simulation results for two different settings.
First, we consider an involutive case by taking, as before, only the $x$-component of the end-effector as task coordinate.
Afterward, we take a non-involutive system using the $x$- and $y$-component as task coordinates.

In both cases, we show results using the locally orthogonal foliations (Sec.~\ref{sec:plane_stacks}) and the neural network approximation (Sec.~\ref{sec:nn}).
We simulate the manipulator for every setting using the trajectory tracking controller (\ref{eq:total_control}).
For the neural network models, we additionally apply pure impedance control (\ref{eq:impedance_controller}) in dedicated experiments.

\subsection{State-of-the-Art: Projection-based Reference Controller}\label{sec:projection}
As a state-of-the-art reference controller for redundancy resolution in combination with impedance control, we use the operational space framework~\cite{Khatib1987}.
In this framework, control torques stemming from secondary tasks are projected into the null space of the main task.

We choose the primary task coordinates $x, y$ describing the position of the end-effector.
Therefore, the forward kinematics function is $\hh(\qq) = [x, y]\tran$ and the Jacobian is $\JJ_x \in \mathbb{R}^{2\times n}$.
As secondary task we choose joint targets $\qq_d$ and use a joint impedance controller. 
\begin{align}\label{eq:projection_with_joint_task}
    \torque_P &= \JJ_x\tran  \left(- \KK_x \left[\hh(\qq) - \xx_d\right] - \DD_x \JJ_x \qdot \right) \\ \nonumber
    &+ \PP_x\tran \left(- \KK_q \left[\qq - \qq_d\right] - \DD_q \qdot\right)\\ \nonumber
    &+ \gggg(\qq),
\end{align}
where $\PP_x = [\II - \JJ_x^{\# M} \JJ_x\tran]$ is a dynamically consistent~\cite{Dietrich2015} torque projection matrix.

\begin{figure}[h!]
    \def\svgwidth{1.1\w}
    \tiny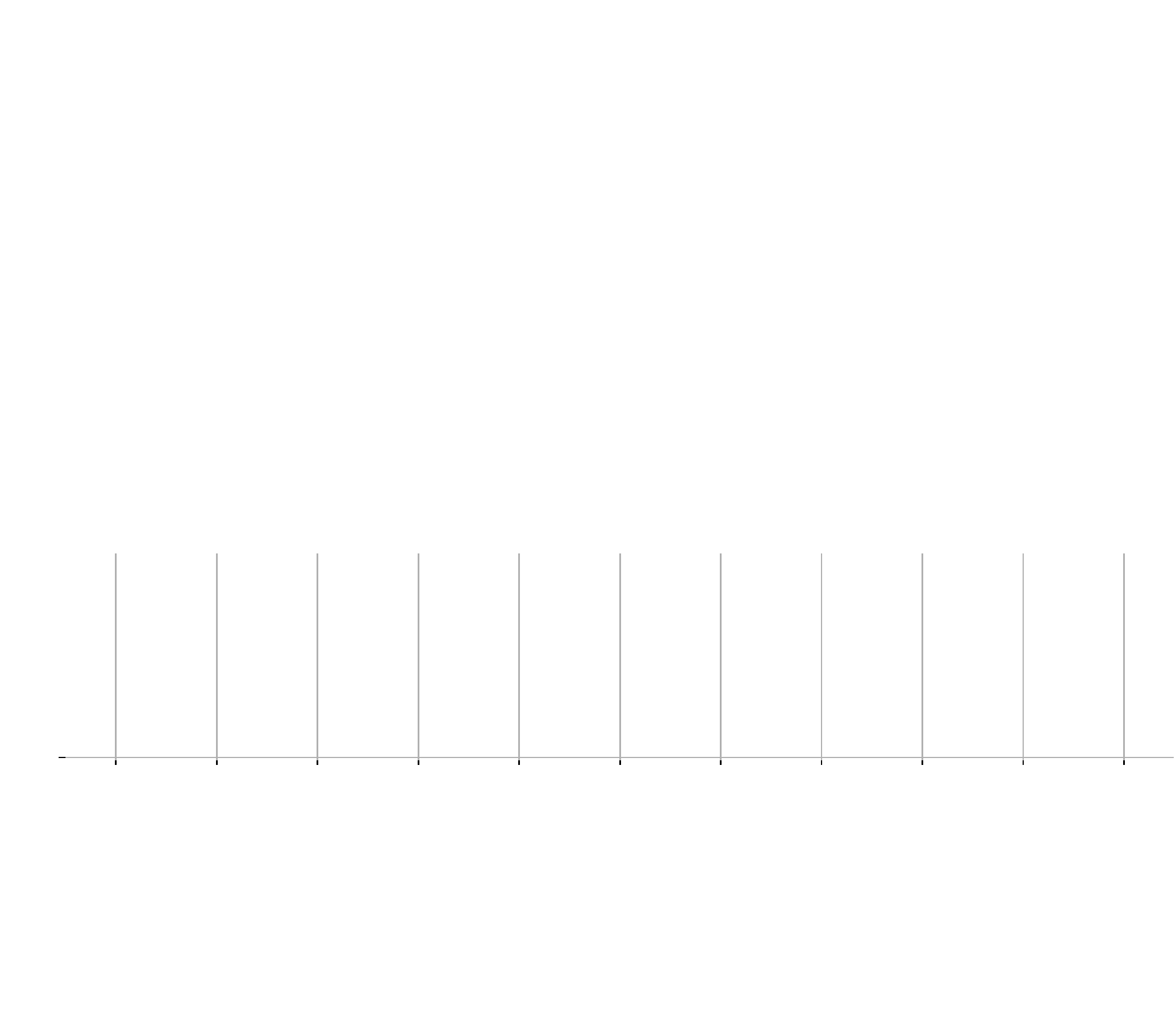
    \caption{Projection-based operational space control for comparison. The secondary tasks are desired joint configurations. Dashed lines: desired values and solid lines: true values. Third pane: joint velocities. Bottom pane: energy stored in the virtual springs by the primary and secondary task impedance controller. 
    \emph{Highly recommended for understanding: the video \texttt{proj.mp4} in the ancillary files.}}\label{fig:projection}
\end{figure}

Consider a trajectory in Cartesian positions $\xx_d(t)$ and joint targets $\qq_d(t)$.
The desired values for $\qq_d$ are shown as dashed lines in the top pane of Fig.~\ref{fig:projection} and the desired Cartesian positions as dashed lines in the second pane.
We apply the control law~\eqref{eq:projection_with_joint_task} to the simulated robot and show the results by solid lines.
In the bottom pane, we show the potential energies stored in the virtual springs by the task impedance controller 
$U_x = \frac{1}{2} \left(\hh(\qq) - \xx_d\right)^\mathsf{T} \KK_x \left(\hh(\qq) - \xx_d\right)$ and the joint impedance controller 
$U_q = \frac{1}{2} \left(\qq - \qq_d\right)^\mathsf{T} \KK_q \left(\qq - \qq_d\right)$.

At every jump in $\qq_d$ at $t \in \{4, 8, 12, 14, 18\}$ we ensure that the new target configuration has the same forward kinematics as the current configuration, i.e., they are on the same self-motion manifold.
Therefore, as we never switch Cartesian target $\xx_d$ and joint target $\qq_d$ simultaneously, we design the joint target $\qq_d$ to be always reachable when switching to it.
This is not true when switching the Cartesian target $\xx_d$ at $t \in \{2, 6, 10, 16\}$: the joint target $\qq_d$ is generally not reachable when jumping the Cartesian target $\xx_d$.

At $t=2$, we command a jump on the $x$-coordinate of the primary task.
The targe is incompatible with the secondary task, and we observe that the secondary task $\qq_d$ is unreachable.
This results in a non-zero potential $U_q$.
However, as the torques from the secondary task are projected out, no torques are acting on the joints.
There is a tensioned virtual spring, which has no effect!
At $t=4$, we choose a reachable joint configuration $\qq_d$ for the new $\xx_d$ and observe that the potential energy $U_q$ converges to zero.
Once we command the $x$-coordinate back to the initial value at $t=6$, the potential energy $U_q$ rises again.
At the end of the time interval $t \in [8, 10]$, all commanded values, both for the primary and secondary task, reach the same values as in the interval $[0, 2]$.
They have been chosen to be compatible, as can be seen from the initial configuration.
However, the potential rises further: the robot does not return to the initial configuration, even though the desired joint configuration is reachable!
The intermediate robot motions have driven it to another configuration, which results in convergence to another local minimum of the potential energy $U_q$.

We repeat the experiment from the new initial configuration at $t=10$.
At $t=12$ and $t=14$, we again command jumps in $\qq_d$ to feasible values.
At $t=12$, the robot tracks the desired joint configuration $\qq_d$ well.
However, at $t=14$, something interesting happens: the jump in $\qq_d$ is entirely in the null space of the primary task.
Nothing happens; and the robot does not move at all since no torques survive the projector.
Still, a highly tensioned virtual spring hides behind the projector.
When interacting with the robot, this may lead to unexpected behavior.
For different configurations, the stability of the local minima of the potential energy $U_q$ through the projector may change, and the robot may snap to the global minimum.
At $t=16$, we command a jump in $\xx_d$ again.
This time the robot moves a lot compared to the comparable jump $t=6$.
At some point, the local minimum of the projected potential is no longer stable, and the robot approaches the global minimum.

After showcasing the problems P1-P6 of a classical approach, let us move to global redundancy resolution coordinates on position level --- as proposed in this paper.

\subsection{Projection-Free Controllers}\label{sec:controllers}
To evaluate the applicability of the new orthogonal foliations in simulation, we state well-known control approaches, now using orthogonal coordinates, and show what the condition (\ref{eq:unified}) implies for the closed-loop dynamics.

The standard rigid multi-body dynamics equation is used to model the manipulators 
\begin{equation}\label{eq:multi-body-dyn}
	\MM(\qq)\qddot + \CC(\qq, \qdot) \qdot + \b{g}(\qq) = \torque .	
\end{equation}

\subsubsection{Impedance Controller}
As in Sec.~\ref{sec:Jacobians-and-Projections} we stack the task coordinates $\xx = \hh(\qq)$ and the (quasi-)orthogonal coordinates $\xxi(\qq)$ and obtain \begin{equation}
	\vphi(\qq) = \begin{bmatrix}
		\hh(\qq)\\\xxi(\qq)	
	\end{bmatrix}
\end{equation}
with the associated Jacobian (\ref{eq:definition_jacobian_Xi_stacked}) $\vphidot = \JJ(\q) \qdot$.
We use an impedance control law \cite{Hogan1985}
\begin{equation}\label{eq:impedance_controller}
	\force_{\varphi}^{\mathrm{imp}} = -\KK_{\varphi} [\b{\varphi}(\qq) - \b{\varphi}_d] - \DD_{\varphi}\JJ\qdot,
\end{equation}
where $\b{\varphi}_d$ denotes the desired position, $\KK_\varphi$ the desired stiffness, and $\DD_{\varphi}$ the damping matrix.
Given a diagonal matrix $\b{Z} = \diag (\zeta_1 \ldots \zeta_m)$ of damping ratios, the damping matrix is computed using the damping design equation \cite{Albu-Schaffer2003} \begin{equation}\label{eq:damping_design}
	\DD_{\varphi} = \MM_{\varphi}^{\nicefrac{1}{2}} \b{Z} \KK_{\varphi}^{\nicefrac{1}{2}} + \KK_{\varphi}^{\nicefrac{1}{2}} \b{Z} \MM_{\varphi}^{\nicefrac{1}{2}},
\end{equation}
where $\AAA^{\nicefrac{1}{2}}$ denotes the matrix-square root of $\AAA$ such that $\AAA = \AAA^{\nicefrac{1}{2}}\AAA^{\nicefrac{1}{2}}$.
The matrix $\MM_{\varphi}$ denotes the transformed mass matrix \begin{equation}\label{eq:mass_transformation}
	\MM_{\varphi} = \left(\JJ\MM^{-1} \JJ\tran\right)^{-1}.
\end{equation}
Suppose $\JJ_x$ and $\JJ_\xi$ are orthogonal with respect to the mass matrix, i.e., we have condition (\ref{eq:unified}) for $\AAA = \MM$ satisfied.
One can verify that the term in the parenthesis of (\ref{eq:mass_transformation}) will have block-diagonal structure (cf. \cite{Khatib1995})
\begin{equation}\label{eq:mass-block-diagonal}
	\Minv_\varphi
	= \JJ \Minv \JJ\tran
	\, \overset{\text{(\ref{eq:unified})}}{=} \, \begin{bmatrix}
		\Minv_x & \bf{0} \\	
		\bf{0} & \Minv_\xi\\	
	\end{bmatrix}.
\end{equation}
Note that we notationally dropped the explicit dependency of $\JJ$, $\MM$, $\DD_{\varphi}$ and $\MM_{\varphi}$ on $\qq$ to remove some clutter.

Using the transformation law for covectors we transform the resulting force from the impedance controller $\force^{\mathrm{imp}}_\varphi$ to joint torques \begin{equation}\label{eq:force_to_torque}
	\torque_\varphi^{\mathrm{imp}} = \JJ\tran \force^{\mathrm{imp}}_\varphi
\end{equation}

Suppose we apply the feedback control law \begin{equation}\label{eq:control-fb-only}
	\torque_1 = \JJ\tran\force_\varphi^{\mathrm{imp}} + \b{g}(\qq)
\end{equation}
on the multi-body dynamics (\ref{eq:multi-body-dyn}) and let us solve for the acceleration $\vphiddot$ \begin{equation}\label{eq:phi-acceleration}
	\vphiddot = \underbrace{\JJ\Minv\JJ\tran}_{\Minv_\varphi}\force^{\mathrm{imp}}_{\varphi} + \underbrace{\dot{\JJ}\qdot}_{\vphiddot_{\mathrm{curv}}} - \underbrace{\JJ\Minv\CC\qdot}_{\vphiddot_{\mathrm{CC}}}.
\end{equation}
The force due to the impedance controller $\force^{\mathrm{imp}}_{\varphi}$ results in an acceleration $\vphiddot$ via the block-diagonal mobility matrix $\Minv_\varphi$ (\ref{eq:mass-block-diagonal}).
Therefore, the impedance controller will not cause immediate accelerations in the task coordinates upon forces in the orthogonal coordinates and vice versa.

\subsubsection{Trajectory Tracking}
The impedance controller will not cause immediate disturbing accelerations between the mutual sets of coordinates.
It may, however, cause delayed couplings via the accumulated velocity $\qdot$.
The joint velocity couples into the acceleration $\vphiddot$ by the last to terms in (\ref{eq:phi-acceleration}): $\vphiddot_{\mathrm{CC}}$ expressing the task acceleration due to Coriolis and centrifugal forces; and $\vphiddot_{\mathrm{curv}}$ depending on the curvature of the self-motion manifolds.

Consequently, to analyze the inertial decoupling of the coordinates we need additional feed-forward terms.
Assume a desired trajectory $(\b{\varphi}_d(t), \dot{\b{\varphi}}_d(t), \ddot{\b{\varphi}}_d(t))$ for the task and the orthogonal coordinates.
We then use the feed-forward controller \begin{equation}\label{eq:feed_forward}
	\force_\varphi^{\mathrm{FF}}	= \MM_\varphi \left[\ddot{\b{\varphi}}_d + \left(\JJ \MM^{-1} \CC - \dot{\JJ}\right) \JJ^{-1} \dot{\b{\varphi}}_d\right] + \DD_\varphi \dot \vvarphi_d.\nonumber
\end{equation}

The resulting force is then transformed to joint torques using the covector transformation (\ref{eq:force_to_torque}).
The final control law is composed of the impedance controller (\ref{eq:impedance_controller}) and the feed-forward terms \begin{equation}\label{eq:total_control}
	\torque_2 = \JJ\tran \left[\force^{\mathrm{imp}}_{\vphi} + \force^{\mathrm{FF}}_{\vphi}\right] + \b{g}(\qq).
\end{equation}
In (\ref{eq:total_control}) all the matrices are evaluated at the actual joint configuration.
The control law is known as \emph{operational space PD+ controller}.

Closed-loop stability of the system can be shown similarly to \cite[Sec.~3.2]{Dietrich2021}, where passivity and global asymptotic stability is shown for a joint space PD+ controller by deriving a storage function in joint coordinates.
When writing the storage function in operational coordinates $\vvarphi$, the same reasoning as in \cite{Dietrich2021} applies for the closed-loop system using (\ref{eq:total_control}).
However, due to singularities in the coordinates, globality is lost, and only local results are feasible.

\subsection{Orthogonal Coordinates: Involutive Case}
First, we show results for the involutive case: one task coordinate and two orthogonal coordinates.
We choose, as before, the $x$-component of the end-effector position as task and will have two orthogonal coordinates $\xi_1$ and $\xi_2$.

We begin showing simulation results using the plane-stack-based locally orthogonal coordinates (Sec.~\ref{sec:plane_stacks}).
We proceed to show simulation results using the neural network-based orthogonal coordinates and present how they overcome the simple approach's limitations and improve decoupling significantly.

\subsubsection{Plane Stack Approximation}
\begin{figure}[h!]
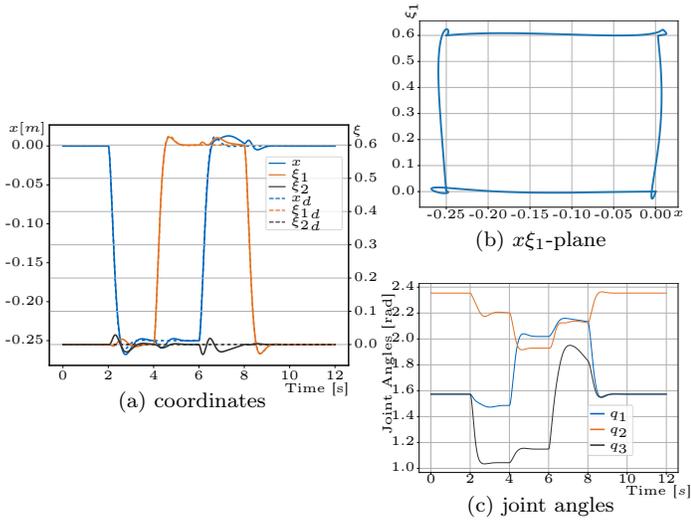

	\centering
	\begin{minipage}{.57\w}
		\subf{coordinates\label{fig:in_plane_trapez_c}}{\textwidth}{\tiny}{in_plane_trapez_c}
	\end{minipage}\hfill
	\begin{minipage}{.45\w}
		\subf{$x\xi_1$-plane\label{fig:in_plane_trapez_phase}}{\textwidth}{\tiny}{in_plane_trapez_phase}\\
		\subf{joint angles\label{fig:in_plane_trapez_q}}{\textwidth}{\tiny}{in_plane_trapez_q}
	\end{minipage}
	\caption{
		Results of dynamic simulation using the plane approximation $\xxi_p(\qq)$ for interleaved steps in the $x$- and $\xi_1$-coordinate. 
	}
	\label{fig:in_plane_trapez}
\end{figure}

Let $\qq_0 \in \mathcal{Q}$ be the chosen base configuration for the locally orthogonal coordinates and consider the vector \begin{equation}\label{eq:grad_x_vector}
	\b{v} = \nabla_{\MM}\hh(\qq_0).
\end{equation}
The vector $\vv$ must be contained in both planes for the two orthogonal coordinates.
We set \begin{equation}
	\b{n}_1 = \frac{\begin{bmatrix}
		v_2 &-v_1 & 0	
	\end{bmatrix}\tran}{\sqrt{v_1^2 + v_2^2}} \qquad
	\b{n}_2 = \frac{\nn_1 \times \vv}{|\nn_1 \times \vv|}
\end{equation}
and use the base planes $E(\q_0, \nn_1)$ and $E(\q_0, \nn_2)$ for the linear, locally orthogonal coordinates (\ref{eq:locally-orthogonal-coordinates}).
These locally orthogonal coordinates and the task coordinates are used in the trajectory tracking controller $\torque_2$ (\ref{eq:total_control}). 
Two interleaved jumps in the $x$- and $\xi_1$-coordinate are commanded.
Fig.~\ref{fig:in_plane_trapez} shows the results of this experiment.

\begin{figure}
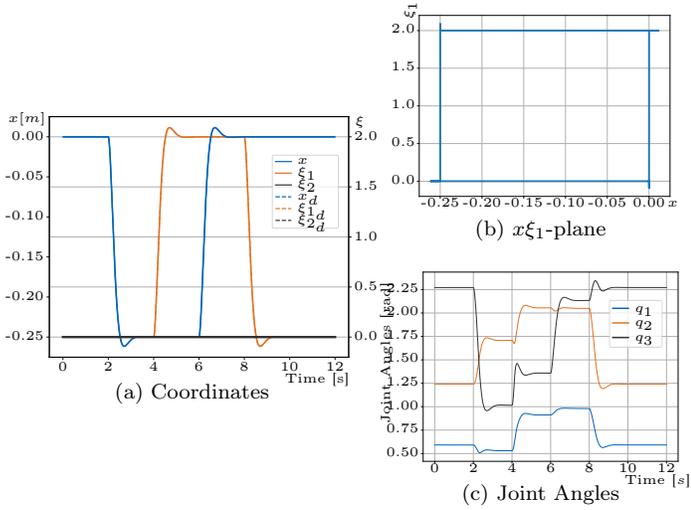

	\centering
	\begin{minipage}{.57\w}
		\subf{Coordinates\label{fig:in_trapez_nn_c}}{\textwidth}{\tiny}{in_nn_trapez_c}
	\end{minipage}\hfill
	\begin{minipage}{.45\w}
		\subf{$x\xi_1$-plane\label{fig:in_trapez_nn_phase}}{\textwidth}{\tiny}{in_nn_trapez_phase}\\
		\subf{Joint Angles\label{fig:in_trapez_nn_q}}{\textwidth}{\tiny}{in_nn_trapez_q}
	\end{minipage}
	\caption{
		Results of dynamic simulation using the neural network model $\xxi_{\mathrm{nn}}(\qq)$ for interleaved steps in the $x$- and $\xi_1$-coordinate.
	}
	\label{fig:in_trapez_nn}
\end{figure}

The initial configuration $\q_0$ is chosen such that the task self-motion manifold shows comparatively low curvature, and the base configuration for the orthogonal coordinates is set to the initial configuration.
Even for relatively large motions in joint space (Fig.~\ref{fig:in_plane_trapez_q}), the dynamics of the task- and orthogonal coordinates look moderately decoupled, which can be observed in the time-evolution of the coordinates and their desired values (Fig.~\ref{fig:in_plane_trapez_c}) as well as in Fig.~\ref{fig:in_plane_trapez_phase} where the trajectory is projected onto the $x\xi_1$-plane.
The decoupling performs better when the configuration is closer to the initial and base configuration $\q_0$.

\begin{figure*}
	\subf{$\xxi_p$, coordinates\label{fig:in_plane_ok_c}}{.33\textwidth}{\tiny}{in_plane_ok_c}
	\subf{$\xxi_p$, coordinates\label{fig:in_plane_fail_c}}{.33\textwidth}{\tiny}{in_plane_fail_c}
	\subf{$\xxi_{\mathrm{nn}}$, coordinates\label{fig:in_nn_c}}{.33\textwidth}{\tiny}{in_nn_c}\\

	\subf{$\xxi_p$, projections\label{fig:in_plane_ok_phase}}{.33\textwidth}{\tiny}{in_plane_ok_phase}
	\subf{$\xxi_p$, projections\label{fig:in_plane_fail_phase}}{.33\textwidth}{\tiny}{in_plane_fail_phase}
	\subf{$\xxi_{\mathrm{nn}}$, projections\label{fig:in_nn_phase}}{.33\textwidth}{\tiny}{in_nn_phase}\\

	\subf{$\xxi_p$, joint space\label{fig:in_plane_ok_q}}{.33\textwidth}{\tiny}{in_plane_ok_q}
	\subf{$\xxi_p$, joint space\label{fig:in_plane_fail_q}}{.33\textwidth}{\tiny}{in_plane_fail_q}
	\subf{$\xxi_{\mathrm{nn}}$, joint space\label{fig:in_nn_q}}{.33\textwidth}{\tiny}{in_nn_q}\\

	\subfloat[$\xxi_p$, trajectory\label{fig:in_plane_ok_manifolds}]{
		\includegraphics[width=.33\textwidth]{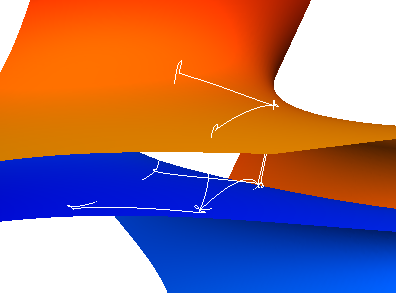}
	}
	\subfloat[$\xxi_p$, trajectory\label{fig:in_plane_fail_manifolds}]{
		\includegraphics[width=.33\textwidth]{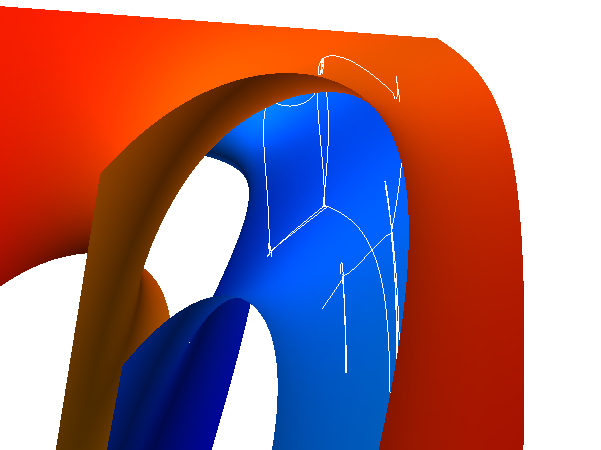}
	}
	\subfloat[$\xxi_{\mathrm{nn}}$, trajectory\label{fig:in_nn_manifolds}]{
		\includegraphics[width=.33\textwidth]{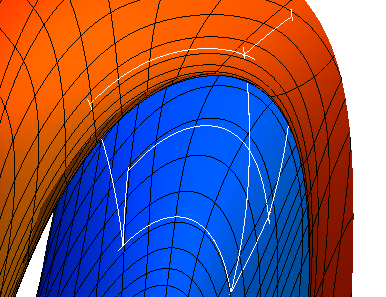}
	}
	\ifx\ieee\undefined \thisfloatpagestyle{empty} \fi
	\caption{
		Simulation results for involutive case using the trajectory tracking controller for commanded jumps in the respective coordinates. 
		While the first and middle columns show results for the simple linear, locally orthogonal coordinates $\xxi_p$, the last columns shows the results using $\xxi_{\mathrm{nn}}$. 
		In the first row we show the time evolution of the desired- and actual values of the coordinates and in the second row we show the trajectory expressed in coordinates projected onto the coordinate planes. 
		Comparison on the magnitude of the motion in joint space to ensure comparability can be read in the third row. 
		Finally, the bottom row shows the trajectory of the manipulator in joint space rendered as the 3D-space curve. 
		Additionally, the self-motion manifolds of the $x$-coordinates for the target values $x=0$ and $x=0.25$ are shown. 
		For the neural network setting we also show the coordinate lines for $\xi_1$ and $\xi_2$ on the self-motion manifolds.
		All joint angles are in radians and Cartesian positions in meters; the orthogonal coordinates $\xi_i$ have no unit.
	}\label{fig:in_all}
\end{figure*}

The first column of Fig.~\ref{fig:in_all} (adgj) shows results for a more extended simulation with multiple commanded jumps in all three coordinates for the same base configuration $\q_0$.
Fig.~\ref{fig:in_plane_ok_manifolds} shows the trajectory of the manipulator in joint space, where additionally, the two relevant task self-motion manifolds are rendered.
The self-motion manifolds show comparatively low curvature in this area.

The linear orthogonal coordinates do not perform well for regions of larger curvature.
The second column in Fig.~\ref{fig:in_all} (behk) shows results for another initial configuration.
Two major problems arise due to the curvature: (1) the decoupling is very poor and (2) the coordinates may contradict each other.
We observe large cross-couplings between the coordinates on which the jump occurs and the other coordinates.
Moreover, we observe that certain targets cannot be reached, which is due to the second problem of mutual contradiction of target values for the coordinates, i.e., task inconsistency.

\subsubsection{Neural Network}\label{sec:nn_in}
Next, we take the neural network $\xxi_{\mathrm{nn}}(\qq)$ approximation trained for $\AAA = \MM$.
We use the same simulation and controller as before but replace the coordinate function $\xxi_p(\qq)$ with the neural network $\xxi_{\mathrm{nn}}(\qq)$.
We first repeat the same interleaved jumps as in Fig.~\ref{fig:in_plane_trapez}.
The results are shown in Fig.~\ref{fig:in_trapez_nn}.
Neither in the time evolution of the coordinates (Fig.~\ref{fig:in_trapez_nn_c}) nor in the trajectory plot (Fig.~\ref{fig:in_trapez_nn_phase}) do we observe any couplings.
The overshoots are due to the damping ratio $\zeta = 0.7$ of the pre-filter for the trajectory generation.

\newcommand{\seeadlerdaunen}{.5\columnwidth}
\begin{figure}
	\hspace{-.25cm}
	\subfloat[Coordinates]{
		\def\svgwidth{\seeadlerdaunen}\tiny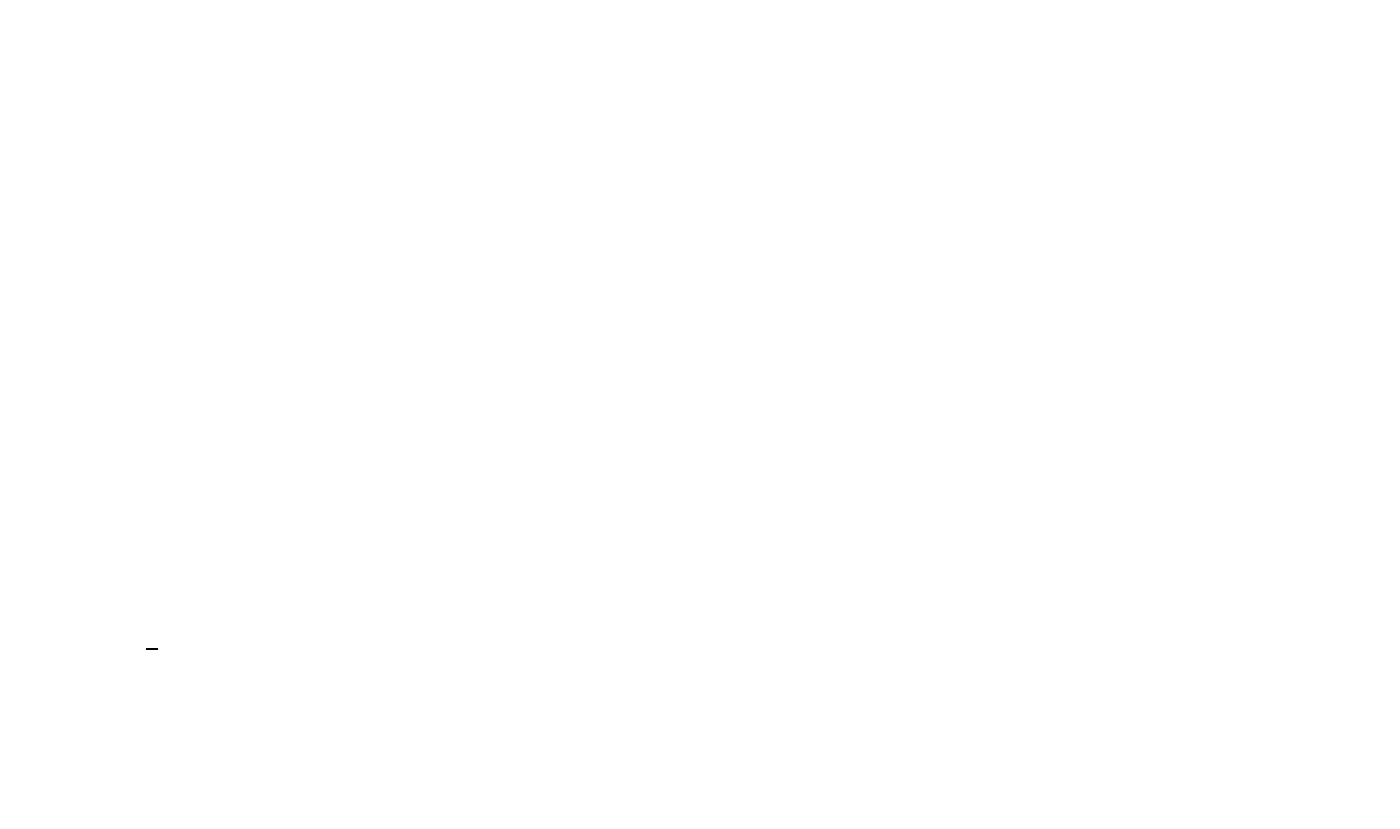
	}
	\subfloat[Disturbing task accelerations\label{fig:task_acceleration_components}]{
		\def\svgwidth{\seeadlerdaunen}\tiny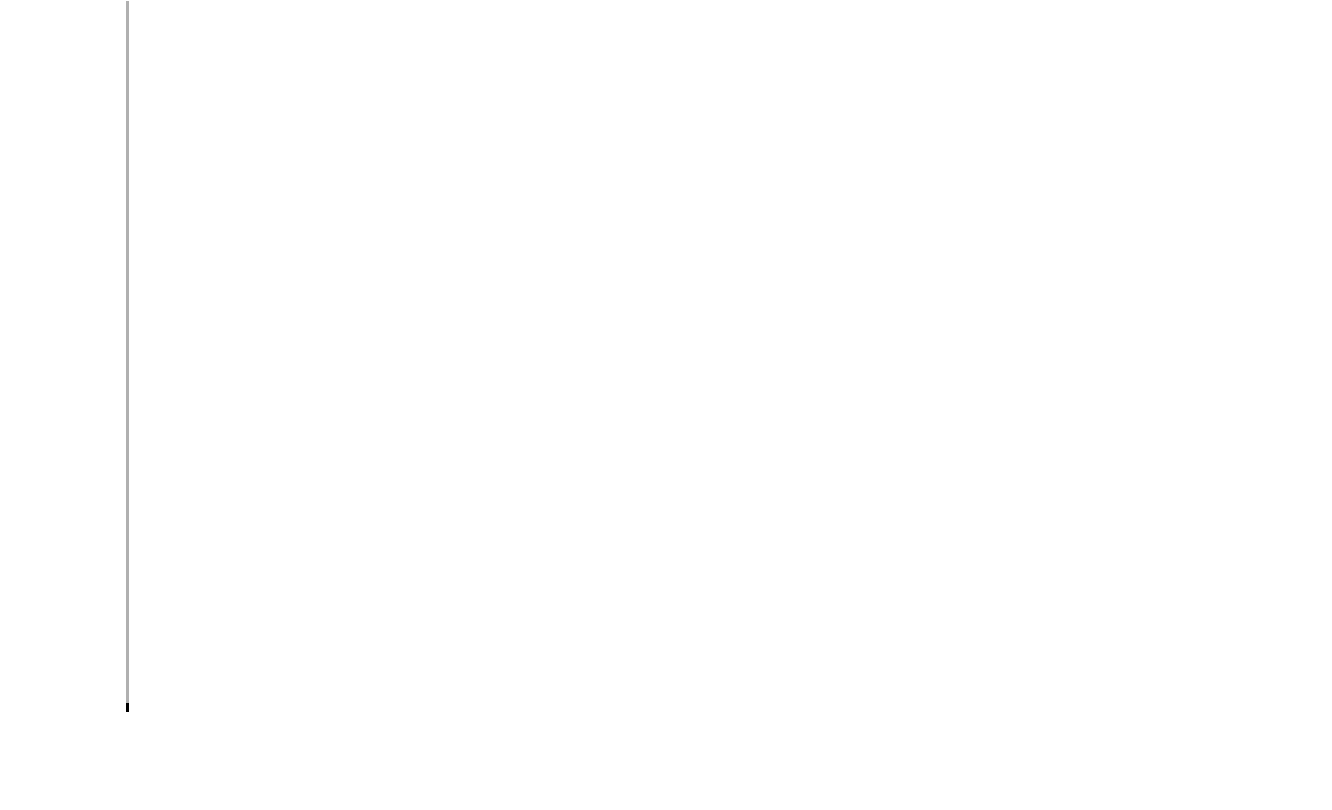
	}\\\vspace{-.2cm}
	\hspace{-.25cm}
	\subfloat[Trajectory in $x\xi_1$-plane]{
		\def\svgwidth{\seeadlerdaunen}\tiny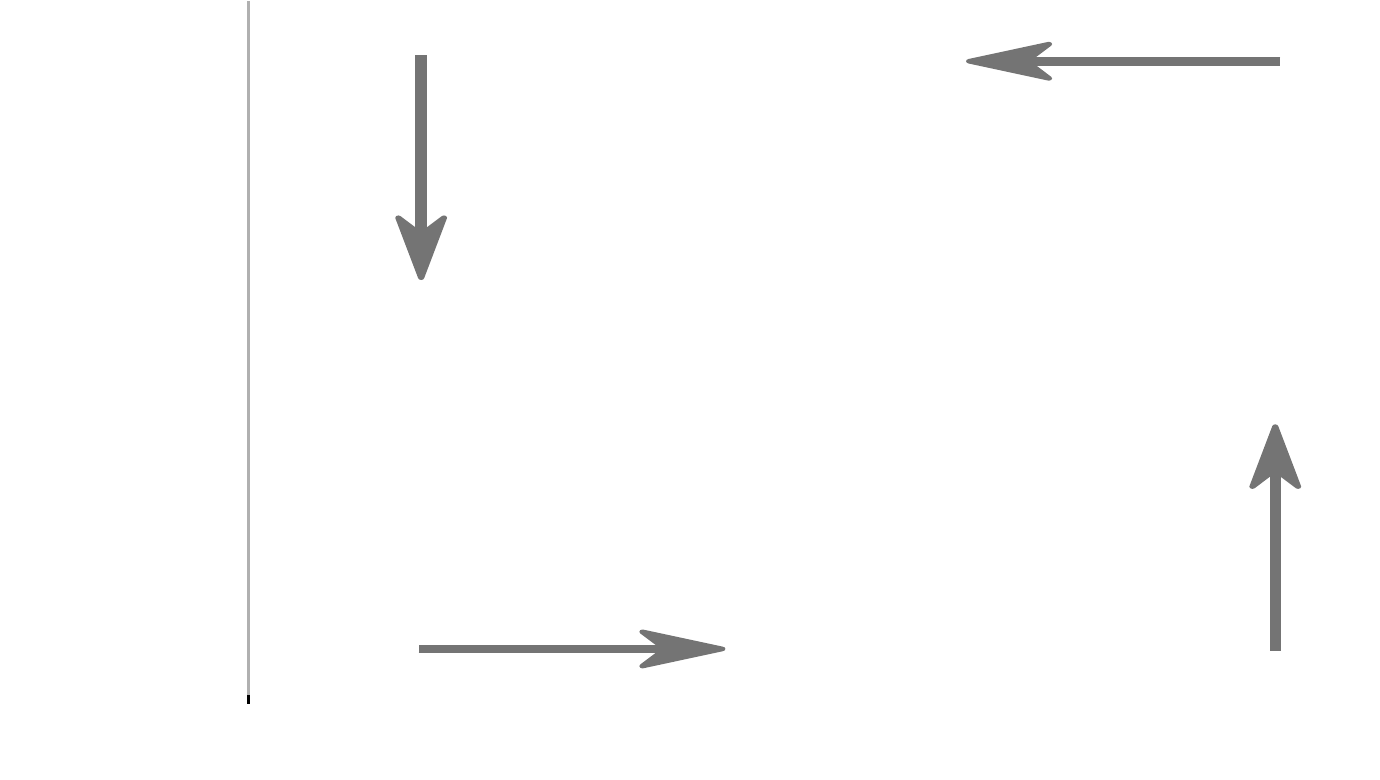
	}
	\subfloat[Joint Angles]{
		\def\svgwidth{\seeadlerdaunen}\tiny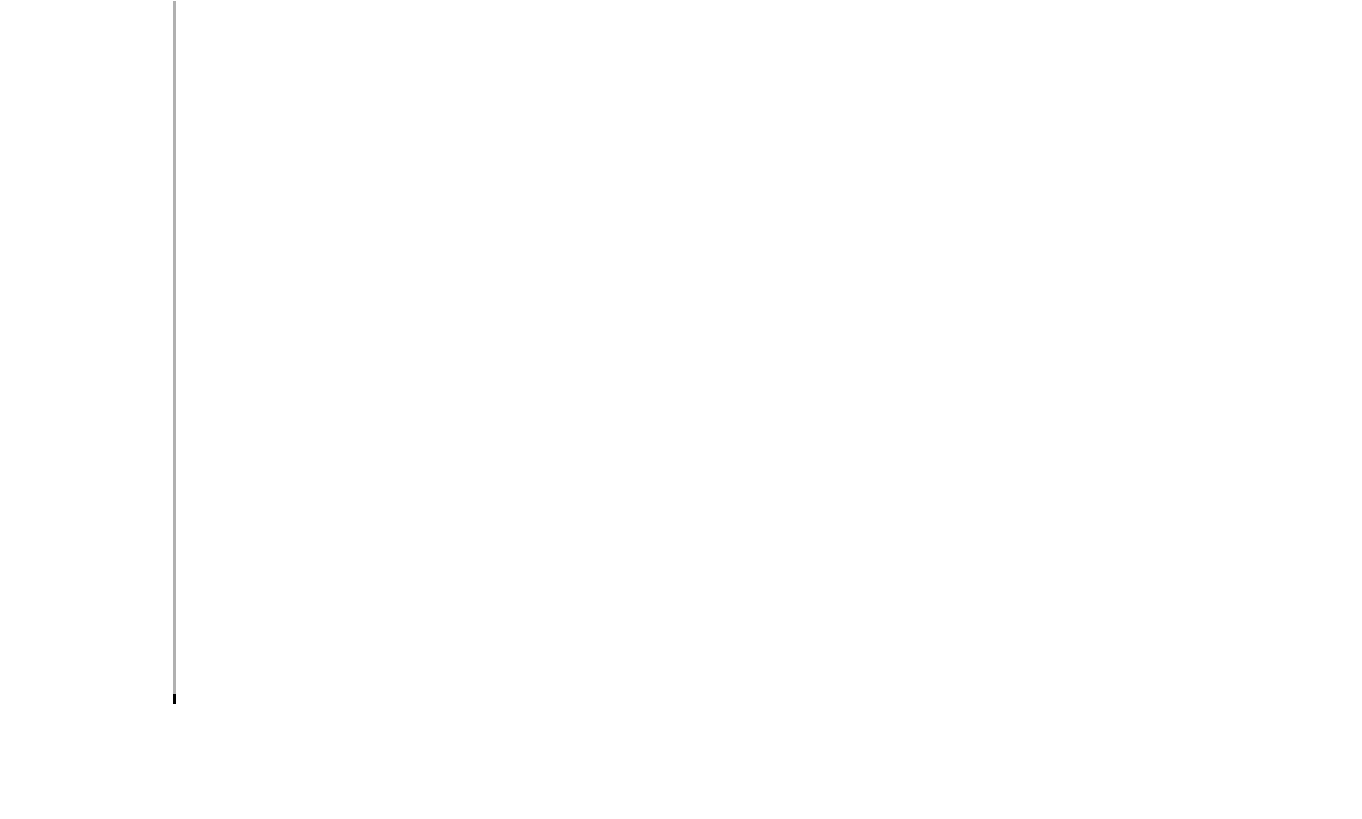
	}
	\caption{
		Simulation results with disabled feed-forward controller for interleaved jumps in the $x$- and $\xi_1$-coordinate. 
		\textbf{(a)}: time evolution of the desired and actual value of the coordinates, 
		\textbf{(b)}: breakdown of the contribution to the disturbing unwanted task accelerations, 
		\textbf{(c)} trajectory of the manipulator in coordinates projected onto the $x\xi_1$-plane and 
		\textbf{(d)}: time evolution of the manipulator in joint space. 
		}
	\label{fig:plots_only_imp}
\end{figure}

The third column in Fig.~\ref{fig:in_all} (cfil) shows the experiment commanding jumps in all the coordinates.
Also, in this experiment, we do not observe mutual coupling between the coordinates.

\subsubsection{Feedback Only}
We also show results for pure feedback control, i.e., we disable the feed-forward controller and only apply the feedback controller $\torque_1$ (\ref{eq:control-fb-only}). 
This way, we analyze the disturbing accelerations due to the curvature of the self-motion manifold and the Coriolis and centrifugal forces and compare them to the accelerations due to errors in the trained models.
We consider the case of regulation: $\dot{x}_{d} = \dot{\xi}_{i_d} = 0$.

For this experiment, we use $\xxi_{\mathrm{nn}}(\qq)$ and command the same jumps as in Fig.~\ref{fig:in_trapez_nn} using the same initial configuration.
Fig.~\ref{fig:plots_only_imp} shows the results.
Using the first row of (\ref{eq:phi-acceleration}) we compute the individual unwanted contributions to the task acceleration $\ddot{x}$.
The term \begin{equation}
	\ddot{x}_{\mathrm{err}} = \JJ_x \MM^{-1} \JJ_\xi\tran \force_\xi^{\mathrm{imp}}
\end{equation}
denotes the contribution due to the training inaccuracies, i.e., if the desired condition (\ref{eq:unified}) is exactly satisfied, this term will vanish.
In contrast to the setting including the feed-forward controller, disturbances and couplings are now clearly observable.
However, when considering the breakdown of the individual contributions to the unwanted task acceleration in Fig.~\ref{fig:task_acceleration_components}, the inaccuracies of the orthogonal coordinates are very low compared to the other terms.
In particular, we have negligible \emph{immediate} task acceleration due to a force $\force_\xi$. Still, we observe delayed responses due to the accumulated speed generated by $\force_\xi$ that couples in via the Coriolis- and centrifugal terms and the curvature of the self-motion manifolds.

\subsubsection{Comparison to the Projection-based Approach}
Fig.~\ref{fig:in_trapez_nn} and Fig.~\ref{fig:in_all}, right column, clearly show the massive benefit of decoupled and consistent coordinates compared to the classical approach in Fig.~\ref{fig:projection}.

\subsection{Quasi-Orthogonal Coordinates: Non-Involutive Case}
\begin{figure*}
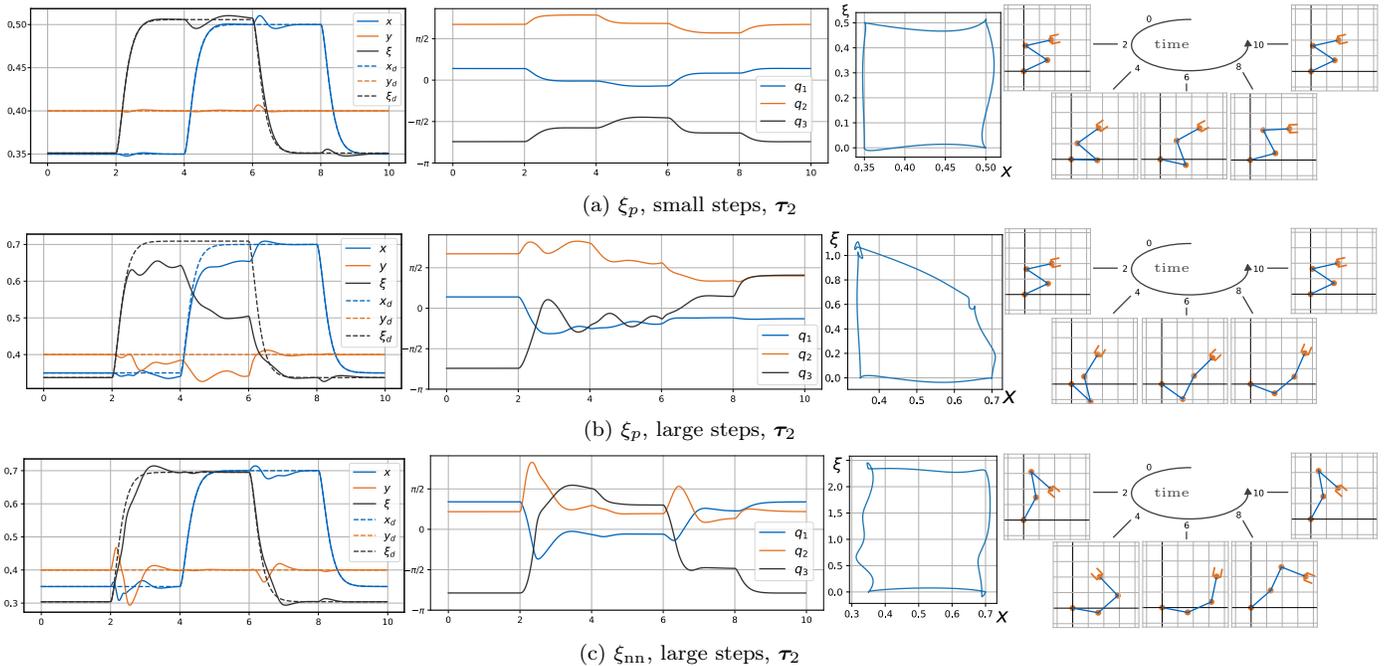

	\centering
	\subf{$\xi_p$, small steps, $\torque_2$\label{fig:ni_plane_small}}{\textwidth}{\tiny}{ni_plane_small}\\\vspace{-.3cm}
	\subf{$\xi_p$, large steps, $\torque_2$\label{fig:ni_plane_large}}{\textwidth}{\tiny}{ni_plane_large}\\\vspace{-.3cm}
	\subf{$\xi_{\mathrm{nn}}$, large steps, $\torque_2$\label{fig:ni_nn_large}}{\textwidth}{\tiny}{ni_nn_large}
	\caption{
		Simulation results for the non-involutive example for jumps in two of the coordinates.
		All joint angles are in radians and Cartesian positions in meters; the orthogonal coordinates $\xi$ have no unit.
		\emph{A video of the simulation with additional information is available as \texttt{orthfol.mp4} in the ancillary files.}
	}
	\label{fig:ni_results}
\end{figure*}
For the non-involutive example, we use the $x$- and $y$-coordinate as task coordinates and have only a single quasi-orthogonal coordinate $\xi$.
The distribution $\Delta$ with the local basis $\{\nabla_{\MM}x, \nabla_{\MM}y\}$ is generally not involutive.

\subsubsection{Plane Stack Approximation}
For the locally orthogonal coordinates we choose \begin{subequations}\begin{align}
	\b{n} &= \frac{\nabla_{\MM}x(\qq_0) \times \nabla_{\MM}y(\qq_0)}{|\nabla_{\MM}x(\qq_0) \times \nabla_{\MM}y(\qq_0)|}\\
	\xi_p(\qq) &= \nn\tran (\q - \q_0).
\end{align}
\end{subequations}

The first two rows of Fig.~\ref{fig:ni_results} show simulation results for jumps in the $x$- and $\xi$-coordinate.
We show the desired and actual values in the first, the evolutions of joint angles in the second, and the projection onto the $x\xi$-plane in the third column. 
In the last column, we show robot configurations upon convergence at the respective times.
For large jumps we have the issue of contradicting targets.
As expected, the overall decoupling is worse than for the involutive example because the problem cannot have a globally optimal solution.

\subsubsection{Neural Network}\label{sec:nn_nonin}
Next, we use $\xi_{\mathrm{nn}}(\qq)$, where the network has been trained for the non-involutive example.
The last row of Fig~\ref{fig:ni_results} shows the results of the simulation with $\xi_{\mathrm{nn}}$.
Compared to the linear approximation with $\xi_p$ the error is more distributed over the space. 
While we have the best decoupling at $\q_0$ for $\xi_p$, there is no priority to any points for the $\xi_{\mathrm{nn}}$.
The third column shows results with $\xi_{\mathrm{nn}}$ for large steps, i.e., we command jumps comparable to those where the linear model failed. Here, the neural network model still performs reasonably well.

Finally, Fig.~\ref{fig:ni_nn_longer} shows simulation results using $\xi_{\mathrm{nn}}$ commanding motion in all the coordinates with (first column) and without (second column) the feed-forward controller.
Especially in the case without the feed-forward term we observe larger couplings between $x$ and $y$ than between $\xi$ and $x/y$.

\subsubsection{Comparison to the Projection-based Approach}
In the non-involutive settings, we cannot have perfect decoupling, which becomes obvious when comparing the results to the involutive case.
The effect of the non-involutivity shows up especially when comparing Figs.~\ref{fig:in_nn_phase}~and~\ref{fig:ni_nn_longer_phase}.
However, a decent performance can be achieved using the neural net.
Compared to the results from the baseline approach in Fig.~\ref{fig:projection}, the residual errors are almost negligible.

\section{Experiment on 7 DoF Robot}\label{sec:hardware}
This section validates the approach experimentally on our 7 DoF \textsc{Kuka} iiwa R800 robot.
We select the position of the end-effector as task coordinates, i.e., we select $\hh: \mathcal{Q} \rightarrow \mathbb{R}^3$; $\hh: \qq \mapsto [x, y, z]\tran$.
Hence, we will have a degree of redundancy of $r = 4$ and four quasi-orthogonal coordinates $\xi_i(\qq)$ for $i \in \{1, 2, 3, 4\}$.
We aim at a kinematic redundancy resolution here and set $\AAA = \II$.

We train the model in a subspace of the configuration space by randomly sampling joint configurations uniformly out of a hyper-cube.
The hyper-cube has an edge length of $\nicefrac{\pi}{2}$ and is centered at $\qq_0 = [0, 0.52, 0, -1.05, 0, 0.79, 0]\tran$.

\begin{figure}
	\centering
	\def\svgwidth{1.1\w}\tiny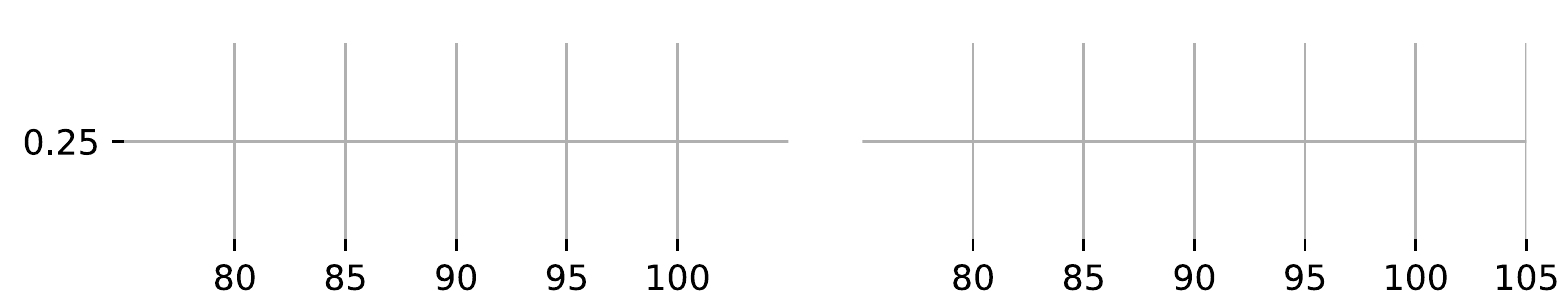
	\ifx\ieee\undefined\vspace{.5cm}\fi
	\caption{Evaluation of the trained network for kinematic orthogonal coordinates for the 7 DoF robot.}
	\label{fig:iiwa_training_hist}
\end{figure}

We train a neural network with $n_1 = 1024$ and $n_2 = 512$ neurons and the first and second layer.
The training was in $200$ epochs with $50$ steps per epoch.
$10^4$ fresh training samples were generated at the beginning of each epoch.
We sample test configurations from the hypercube and evaluate angles in the histogram in Fig.~\ref{fig:iiwa_training_hist}.
The standard deviation of the residual angle between rows of $\JJ_x$ and $\JJ_\xi$ amounts to $\sigma=2.06^\circ$; and between mutual rows of $\JJ_\xi$ we obtain $\sigma=3.44^\circ$.

Finally, we generate a smooth desired trajectory $\vphi_d(t) = (\xx_d, \xxi_d)$ by low-pass filtering jumps in the coordinates as before.
The neural network is trained for kinematic control, i.e., we apply velocity-based control using $\qdot = \JJ^{-1}\dot{\vphi}_d$.

\begin{figure}[h!]
	\centering
	\def\svgwidth{1.1\w}\footnotesize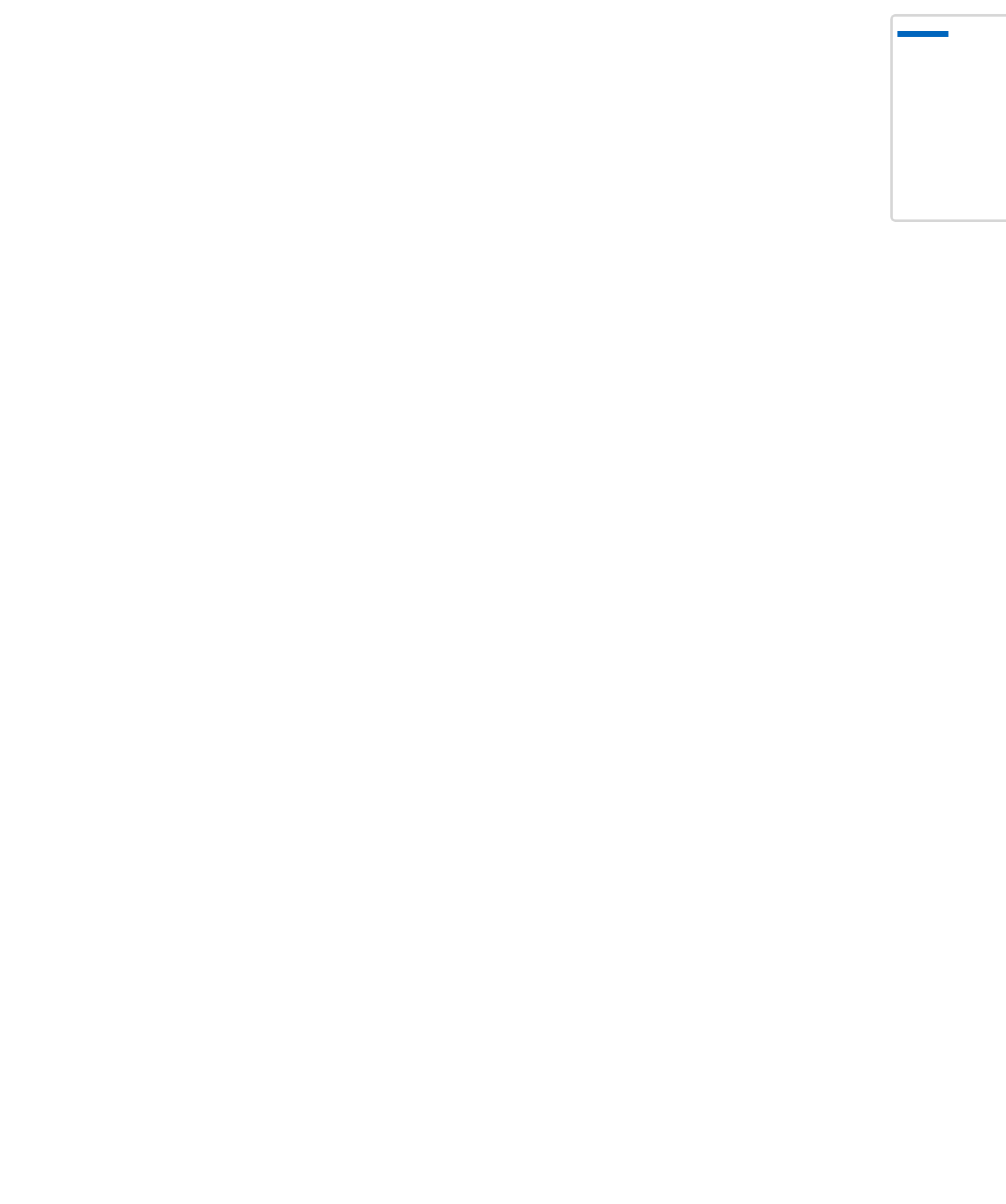
	\caption{Results of kinematic control using orthogonal coordinates on the \textsc{Kuka} LBR iiwa R800 for a desired trajectory in $\xx$ and $\xxi$. 
	Bold lines show desired values, and thin lines in lighter color show measured values. 
	(a) Time history of the task coordinates; 
	(b) Time history of the orthogonal coordinates $\xxi$; and 
	(c) Joint angles over time. This last plot has no desired values.}
	\label{fig:iiwa_exp}
\end{figure}

We show results for a hardware experiment on our 7 DoF robot in Fig.~\ref{fig:iiwa_exp}.
On the top, we show the time history of the primary task $\xx$, and in the middle the $\xxi$ coordinates over time.
Bold lines show desired values, and thin lines in lighter color show measured values.
The bottom pane shows the measured joint positions over time.
Comparing the jump heights in the bottom pane, we observe that the jumps in $\xi_i$ also perform significant motions of the robot.

We interleave jumps in the task and orthogonal coordinates and perform a long and complex journey through the task- and orthogonal space.
Ultimately, we set the final values for $\xx$ and $\xxi$ back to the initial ones.
Our robot has also returned to the same joint configuration!
This can be observed with the help of the dotted lines in the bottom pane.

On the level of the commanded values, there is no observable coupling between the primary task and the secondary task, which is consistent with the very low residuum reported above. 
This demonstrates that the algorithm achieves an almost perfect kinematic decoupling. 
Moreover, note that after returning to the initial Cartesian point, the robot returns to the initial joint configuration. 
In contrast to state of the art, we are free to move at wish in the orthogonal coordinates in between, which can be used for fulfilling secondary tasks. 
On the measured values, one can observe some slight dynamic coupling between coordinates, which is unsurprising, as we are not performing decoupling on dynamic, but only on kinematic level in this experiment. 
Still, remarkably, the dynamic coupling among primary tasks is less than the couplings to the orthogonal coordinates.   
The seventh axis of the robot does not influence the main task.
Interestingly, the second orthogonal coordinate $\xi_2$ learned to exploit this by only controlling this joint.

\section{Discussion and Conclusions}\label{sec:discussion}
We gave a systematic analysis of the geometrical aspects of robot redundancy resolution on configuration (position) level, i.e., in terms of quasi-orthogonal configuration space coordinates. 
Strictly decoupled coordinates can be achieved if the task space is one-dimensional.
For higher dimensions, this is only possible in particular cases of forward kinematics, as shown in \cite[pp. 151--154]{McKay2020} for three dimensions. 
For all other cases, we introduced the concept of quasi-orthogonal coordinates, providing the best possible decoupling in a certain least squares sense.
Beyond the quite exciting geometric insights into the problem, one might ask, what are the practical advantages of this new approach to robotics? 
While it is too early to draw general conclusions, some apparent aspects can already be discussed.

\subsection{Comparison to Projection-based Methods}
Classical projection-based methods destroy passivity and can lead to \emph{hidden springs} that suddenly discharge, causing unexpected behavior and causing shaky and vibrating behavior in certain configurations.
These problems vanish in the case of orthogonal coordinates, where no projection is needed, and convergence analysis works simply based on the sum of all potentials.  
Note that for topological reasons, the potentials may have several (possibly unstable) equilibria, for instance, when controlling on $SO(3)$~\cite{Koditschek1989}.
In the quasi-orthogonal coordinates (non-involutive) case, we can circumvent the need for projectors and avoid the problems of local redundancy resolution.
This comes at the cost of accepting slight couplings.
In this case, an additional projection could be applied to remove the remaining couplings, using the projectors $\PP$ or $\PP\tran$ on the quasi-orthogonal foliations.
The projection will only remove a small portion of the torque $\torque = \JJ_\xi\tran \force_\xi$.

\subsection{Comparison to Global Redundancy Resolution}
Several publications have dealt with a restrictive version of global redundancy resolution (e.g., \cite{Xie2022,Hauser2018}).
These publications solve the redundancy by fixing one joint configuration for every end-effector pose; for closed end-effector paths, the joint configuration will never drift away.
This comes at the cost of virtually removing degrees of freedom, preventing secondary tasks.
Other publications on global redundancy resolution do not fix a one-to-one correspondence but do not provide coordinates (e.g.~\cite{Mussa-Ivaldi1991,DeLuca1992}).

Haug's cyclic differentiable manifold representation~\cite{Haug2023} provides global redundancy resolution and retains the freedom to use redundant degrees of freedom. 
It is based on first identifying linear subspaces in a neighborhood of a base configuration and then re-adding the nonlinearity of the main task by a nonlinear function that depends on both the main task and the additional coordinates (see~\eqref{eq:haug_subspaces}).
In this approach, the two sets of coordinates no longer directly act on the robot via an analytical Jacobian with interpretable directions, which limits the use of the approach for operational space control.

Our approach extends these results: for any fixed desired value $\xxi_d$, we can do the same cyclic tasks as in the global redundancy resolution publications.
However, we can still perform self-motions at any time by commanding time-dependent trajectories $\xxi_d(t)$ and even superimpose these motions to the main task.
The dynamical decoupling will ensure that the main task is not disturbed.
We can perform secondary tasks \emph{and} have global redundancy resolution on position-level.

\subsection{Representation of Orthogonal Foliations}
The objective of finding orthogonal foliations can be written as a system of partial differential equations.
We have presented a proof of concept for the orthogonal foliations by formulating the objective as an optimization problem and using a neural network.
A variety of other solutions to find a $\xxi(\qq)$ optimally satisfying (\ref{eq:unified}) exist.
For instance, discrete methods and finite elements are often used in computer graphics, structure analysis, and materials science to solve PDEs.
We recently discovered that in the field of molecular dynamics, researchers find learned potential fields by Gaussian processes~\cite{Schmitz2022}.
We will investigate if we can apply their methods to our problem in the future.

The approximation by linear local quasi-orthogonal coordinates, which we employed for comparison in the experiments, might provide a practical value. 
If one considered the learning of quasi-orthogonal coordinates too expensive, one could still use the linear approximation in combination with classical projection-based methods instead of defining potentials classically, directly in joint space. 
One can expect in practice that the components which need to be projected out will be substantially reduced. 

The comparison and detailed investigation of different approaches to find a parametrization $\xxi(\qq)$ for quasi-orthogonal foliations based on (\ref{eq:unified}) was not within the scope of this paper and will be addressed in future work.

\subsection{Computational Complexity and Training Times}
In contrast to classical projection-based methods, our approach requires a training phase.
However, we would like to emphasize that the training phase is required once and can happen offline.
During operation, only forward passes of the trained network are needed.
The computation times for a forward pass are deterministic and, therefore, real-time capable.

We measured the time needed for a forward pass, i.e., computing the learned coordinates $\xxi(\qq)$ and the corresponding Jacobian $\JJ_\xi(\qq)$, for the large model for the LBR iiwa robot.
Without any optimization and using a single core on a regular laptop, we achieve a mean forward pass time of $133\mu\text{s}$ with a standard deviation of $28.2\mu\text{s}$ for our na\"{i}ve Eigen-based C++ implementation. 
This could even be further improved using hardware acceleration or multiple cores, and the variance can be brought to zero by a real-time implementation.

The network must be trained for every robot and choice of task space.
The manufacturer of the robot could even do this.
There is no need to retrain the network for different positions in joint- or task space!
Yet, we provide the training times for the experiments in Table~\ref{tab:training_times}.

\begin{table}
    \centering
    \caption{Training times}\label{tab:training_times}
    \begin{tabular}{c|c|c|c|c|c|c}
        Sec. & $n$ & $m$ & $n_1$ & $n_2$ & metric & training minutes \\\hline
        \multirow{2}{*}{\ref{sec:nn_two_dof}} & \multirow{2}{*}{2} & \multirow{2}{*}{1} & \multirow{2}{*}{256} & \multirow{2}{*}{64} & $\II$ & 1.96 \\\cline{6-7}
        & & & & & $\MM$ & 2.05 \\\hline
        \ref{sec:nn_noninv_two_dof} & 3 & 2 & 256 & 64 & $\II$ & 4.33 \\\hline
        \ref{sec:nn_in} & 3 & 1 & 512 & 256 & $\MM$ & 6.69 \\\hline
        \ref{sec:nn_nonin} & 3 & 2 & 512 & 256 & $\MM$ & 4.44 \\\hline
        \ref{sec:hardware} & 7 & 3 & 1024 & 512 & $\II$ & 49.36 \\\hline
    \end{tabular}
\end{table}

\subsection{Interpretability of the Learned Coordinates}
The learned coordinates span all the remaining degrees of freedom.
However, they do so arbitrarily, and the robot motions upon motions in the learned coordinates are not necessarily intuitive and will depend on what the network converges to.
For now, we are happy with any solution that the network finds.
One approach to make them more interpretable is to additionally use interpretable secondary tasks as in the projection framework and incorporate them into the optimization problem.
The cost function will penalize how much the learned coordinates deviate from the secondary tasks.
This was beyond the scope of this paper and will be addressed in future work.

\section*{Acknowledgments}
\addcontentsline{toc}{section}{Acknowledgments}
We want to thank Daniel Matthes from the Technical University of Munich for initial discussions and literature hints, and Benjamin McKay from the University College Cork for his feedback on triply orthogonal webs and foliations and for addressing one of our main questions in his manuscript~\cite[pp. 151--154]{McKay2020}.
Special thanks are also directed to Alexander Dietrich from the DLR for detailed feedback on this paper.

\bibliography{full}

\begin{thebibliography}{10}
\providecommand{\url}[1]{#1}
\csname url@samestyle\endcsname
\providecommand{\newblock}{\relax}
\providecommand{\bibinfo}[2]{#2}
\providecommand{\BIBentrySTDinterwordspacing}{\spaceskip=0pt\relax}
\providecommand{\BIBentryALTinterwordstretchfactor}{4}
\providecommand{\BIBentryALTinterwordspacing}{\spaceskip=\fontdimen2\font plus
\BIBentryALTinterwordstretchfactor\fontdimen3\font minus
  \fontdimen4\font\relax}
\providecommand{\BIBforeignlanguage}[2]{{%
\expandafter\ifx\csname l@#1\endcsname\relax
\typeout{** WARNING: IEEEtran.bst: No hyphenation pattern has been}%
\typeout{** loaded for the language `#1'. Using the pattern for}%
\typeout{** the default language instead.}%
\else
\language=\csname l@#1\endcsname
\fi
#2}}
\providecommand{\BIBdecl}{\relax}
\BIBdecl
\renewcommand{\BIBentryALTinterwordstretchfactor}{4}

\bibitem{Khatib1987}
O.~Khatib, ``A unified approach for motion and force control of robot
  manipulators: The operational space formulation,'' \emph{IEEE Journal on
  Robotics and Automation}, vol.~3, no.~1, pp. 43--53, 1987.

\bibitem{Khatib1995}
------, ``Inertial properties in robotic manipulation: An object-level
  framework,'' \emph{The International Journal of Robotics Research}, vol.~14,
  no.~1, pp. 19--36, 1995.

\bibitem{Nakamura1987}
Y.~Nakamura, H.~Hanafusa, and T.~Yoshikawa, ``Task-priority based redundancy
  control of robot manipulators,'' \emph{The International Journal of Robotics
  Research}, vol.~6, no.~2, pp. 3--15, 1987.

\bibitem{Siciliano1991}
B.~Siciliano and J.-J. Slotine, ``A general framework for managing multiple
  tasks in highly redundant robotic systems,'' in \emph{Proceedings of the IEEE
  International Conference on Advanced Robotics}, 1991, pp. 1211--1216 vol.2.

\bibitem{Baerlocher1998}
P.~Baerlocher and R.~Boulic, ``Task-priority formulations for the kinematic
  control of highly redundant articulated structures,'' in \emph{Proceedings of
  the IEEE/RSJ International Conference on Intelligent Robots and Systems.},
  vol.~1, 1998, pp. 323--329 vol.1.

\bibitem{Liegeois1977}
A.~Liégeois, ``Automatic supervisory control of the configuration and behavior
  of multibody mechanisms,'' \emph{IEEE Transactions on Systems, Man, and
  Cybernetics}, vol.~7, no.~12, pp. 868--871, 1977.

\bibitem{Nakanishi2008}
J.~Nakanishi, R.~Cory, M.~Mistry, J.~Peters, and S.~Schaal, ``Operational space
  control: A theoretical and empirical comparison,'' \emph{The International
  Journal of Robotics Research}, vol.~27, no.~6, pp. 737--757, 2008.

\bibitem{Siciliano1990}
B.~Siciliano, ``Kinematic control of redundant robot manipulators: A
  tutorial,'' \emph{Journal of Intelligent and Robotic Systems}, vol.~3, pp.
  201--212, Sep. 1990.

\bibitem{Dietrich2013}
A.~Dietrich, C.~Ott, and A.~Albu-Sch{\"a}ffer, ``Multi-objective compliance
  control of redundant manipulators: Hierarchy, control, and stability,'' in
  \emph{Proceedings of the IEEE/RSJ International Conference on Intelligent
  Robots and Systems}, Nov. 2013, pp. 3043--3050.

\bibitem{Siciliano2008a}
B.~Siciliano, L.~Sciavicco, L.~Villani, and G.~Oriolo, \emph{Robotics:
  Modelling, Planning and Control}, 1st~ed.\hskip 1em plus 0.5em minus
  0.4em\relax Springer Publishing Company, Incorporated, London, 2009.

\bibitem{Murray1994}
R.~M. Murray, S.~S. Sastry, and L.~Zexiang, \emph{A Mathematical Introduction
  to Robotic Manipulation}, 1st~ed.\hskip 1em plus 0.5em minus 0.4em\relax CRC
  Press, Inc., 1994.

\bibitem{Lynch2017}
K.~M. Lynch and F.~C. Park, \emph{Modern Robotics: Mechanics, Planning, and
  Control}, 1st~ed.\hskip 1em plus 0.5em minus 0.4em\relax Cambridge University
  Press, 2017.

\bibitem{Spong2005}
M.~W. Spong, S.~Hutchinson, and M.~Vidyasagar, \emph{Robot Modeling and
  Control}.\hskip 1em plus 0.5em minus 0.4em\relax Wiley, 2005.

\bibitem{Siciliano2008}
B.~Siciliano and O.~Khatib, Eds., \emph{Springer Handbook of Robotics}.\hskip
  1em plus 0.5em minus 0.4em\relax Springer Berlin Heidelberg, 2008.

\bibitem{Maciejewski1985}
A.~A. Maciejewski and C.~A. Klein, ``Obstacle avoidance for kinematically
  redundant manipulators in dynamically varying environments,'' \emph{The
  International Journal of Robotics Research}, vol.~4, no.~3, pp. 109--117,
  1985.

\bibitem{Dietrich2012}
A.~{Dietrich}, T.~{Wimböck}, A.~{Albu-Schäffer}, and G.~{Hirzinger},
  ``Integration of reactive, torque-based self-collision avoidance into a task
  hierarchy,'' \emph{IEEE Transactions on Robotics}, vol.~28, no.~6, pp.
  1278--1293, 2012.

\bibitem{Hagn2009}
U.~Hagn \emph{et~al.}, ``{DLR MiroSurge}: a versatile system for research in
  endoscopic telesurgery,'' \emph{International Journal of Computer Assisted
  Radiology and Surgery}, vol.~5, no.~2, pp. 183--193, 2010.

\bibitem{Hutzl2014}
J.~{Hutzl}, D.~{Oertel}, and H.~{Wörn}, ``Knowledge-based direction prediction
  to optimize the null-space parameter of a redundant robot in a
  telemanipulation scenario,'' in \emph{IEEE International Symposium on Robotic
  and Sensors Environments (ROSE) Proceedings}, Oct. 2014, pp. 25--30.

\bibitem{Su2019}
H.~{Su}, S.~{Li}, J.~{Manivannan}, L.~{Bascetta}, G.~{Ferrigno}, and E.~{De
  Momi}, ``Manipulability optimization control of a serial redundant robot for
  robot-assisted minimally invasive surgery,'' in \emph{Proceedings of the IEEE
  International Conference on Robotics and Automation}, 2019, pp. 1323--1328.

\bibitem{Walker1990}
I.~D. {Walker}, ``The use of kinematic redundancy in reducing impact and
  contact effects in manipulation,'' in \emph{Proceedings of the IEEE
  International Conference on Robotics and Automation}, vol.~1, May 1990, pp.
  434--439.

\bibitem{Mansfeld2017a}
N.~Mansfeld, B.~Djellab, J.~R. Veuthey, F.~Beck, and S.~Haddadin, ``Improving
  the performance of biomechanically safe velocity control for redundant robots
  through reflected mass minimization,'' in \emph{Proceedings of the IEEE/RSJ
  International Conference on Intelligent Robots and Systems}, 2017, pp.
  5390--5397.

\bibitem{Doty1993}
K.~L. Doty, C.~Melchiorri, and C.~Bonivento, ``A theory of generalized inverses
  applied to robotics,'' \emph{The International Journal of Robotics Research},
  vol.~12, no.~1, pp. 1--19, 1993.

\bibitem{Hermus2022}
J.~Hermus, J.~Lachner, D.~Verdi, and N.~Hogan, ``Exploiting redundancy to
  facilitate physical interaction,'' \emph{IEEE Transactions on Robotics},
  vol.~38, no.~1, pp. 599--615, 2022.

\bibitem{Klein1983}
C.~A. {Klein} and C.-H. {Huang}, ``Review of pseudoinverse control for use with
  kinematically redundant manipulators,'' \emph{IEEE Transactions on Systems,
  Man, and Cybernetics}, vol.~13, no.~2, pp. 245--250, Mar. 1983.

\bibitem{Sentis2005}
L.~Sentis and O.~Khatib, ``Synthesis of whole-body behaviors through
  hierarchical control of behavioral primitives,'' \emph{International Journal
  of Humanoid Robotics}, vol.~02, no.~04, pp. 505--518, 2005.

\bibitem{Wu2023}
X.~Wu, C.~Ott, A.~Albu-Schäffer, and A.~Dietrich, ``Passive decoupled
  multitask controller for redundant robots,'' \emph{IEEE Transactions on
  Control Systems Technology}, vol.~31, no.~1, pp. 1--16, 2023.

\bibitem{Dietrich2017}
A.~Dietrich, X.~Wu, K.~Bussmann, C.~Ott, A.~Albu-Schäffer, and S.~Stramigioli,
  ``Passive hierarchical impedance control via energy tanks,'' \emph{IEEE
  Robotics and Automation Letters}, vol.~2, no.~2, pp. 522--529, 2017.

\bibitem{Folkertsma2017}
G.~A. Folkertsma and S.~Stramigioli, ``\BIBforeignlanguage{English}{Energy in
  robotics},'' \emph{\BIBforeignlanguage{English}{Foundations and
  Trends{\textregistered} in Robotics}}, vol.~6, no.~3, pp. 140--210, Oct.
  2017.

\bibitem{Dietrich2015}
A.~Dietrich, C.~Ott, and A.~Albu-Schäffer, ``An overview of null space
  projections for redundant, torque-controlled robots,'' \emph{The
  International Journal of Robotics Research}, vol.~34, no.~11, pp. 1385--1400,
  2015.

\bibitem{Antonelli2009}
G.~Antonelli, ``Stability analysis for prioritized closed-loop inverse
  kinematic algorithms for redundant robotic systems,'' \emph{IEEE Transactions
  on Robotics}, vol.~25, no.~5, pp. 985--994, 2009.

\bibitem{Ott2015}
C.~Ott, A.~Dietrich, and A.~Albu-Schäffer, ``Prioritized multi-task compliance
  control of redundant manipulators,'' \emph{Automatica}, vol.~53, pp.
  416--423, 2015.

\bibitem{Dietrich2020}
A.~{Dietrich} and C.~{Ott}, ``Hierarchical impedance-based tracking control of
  kinematically redundant robots,'' \emph{IEEE Transactions on Robotics},
  vol.~36, no.~1, pp. 204--221, 2020.

\bibitem{Burdick1989}
J.~W. Burdick, ``On the inverse kinematics of redundant manipulators:
  characterization of the self-motion manifolds,'' in \emph{Proceedings of the
  {IEEE} International Conference on Robotics and Automation}.\hskip 1em plus
  0.5em minus 0.4em\relax {IEEE} Computer Society, 1989, pp. 264--270.

\bibitem{Dyck2022}
M.~Dyck, A.~Sachtler, J.~Klodmann, and A.~Albu-Schäffer, ``Impedance control
  on arbitrary surfaces for ultrasound scanning using discrete differential
  geometry,'' \emph{IEEE Robotics and Automation Letters}, vol.~7, no.~3, pp.
  7738--7746, 2022.

\bibitem{Mussa-Ivaldi1991}
F.~A. Mussa-Ivaldi and N.~Hogan, ``Integrable solutions of kinematic redundancy
  via impedance control,'' \emph{The International Journal of Robotics
  Research}, vol.~10, no.~5, pp. 481--491, 1991.

\bibitem{DeLuca1992}
A.~De~Luca, L.~Lanari, and G.~Oriolo, ``Control of redundant robots on cyclic
  trajectories,'' in \emph{Proceedings of the IEEE International Conference on
  Robotics and Automation}, 1992, pp. 500--506 vol.1.

\bibitem{Hauser2018}
K.~Hauser and S.~Emmons, ``Global redundancy resolution via continuous
  pseudoinversion of the forward kinematic map,'' \emph{IEEE Transactions on
  Automation Science and Engineering}, vol.~15, no.~3, pp. 932--944, 2018.

\bibitem{Xie2022}
Z.~Xie, L.~Jin, X.~Luo, Z.~Sun, and M.~Liu, ``{RNN} for repetitive motion
  generation of redundant robot manipulators: An orthogonal projection-based
  scheme,'' \emph{IEEE Transactions on Neural Networks and Learning Systems},
  vol.~33, no.~2, pp. 615--628, 2022.

\bibitem{Haug2023}
E.~J. Haug, ``{A Cyclic Differentiable Manifold Representation of Redundant
  Manipulator Kinematics},'' \emph{Journal of Mechanisms and Robotics},
  vol.~16, no.~6, p. 061005, 2024.

\bibitem{Haug2022}
E.~J. Haug and A.~Peidro, ``{Redundant Manipulator Kinematics and Dynamics on
  Differentiable Manifolds},'' \emph{Journal of Computational and Nonlinear
  Dynamics}, vol.~17, no.~11, p. 111008, 2022.

\bibitem{Park1999}
J.~{Park}, W.~{Chung}, and Y.~{Youm}, ``On dynamical decoupling of
  kinematically redundant manipulators,'' in \emph{Proceedings of the IEEE/RSJ
  International Conference on Intelligent Robots and Systems}, 1999, pp.
  1495--1500.

\bibitem{Ott2008b}
C.~{Ott}, A.~{Kugi}, and Y.~Nakamura, ``Resolving the problem of
  non-integrability of nullspace velocities for compliance control of redundant
  manipulators by using semi-definite {L}yapunov functions,'' in
  \emph{Proceedings of the IEEE International Conference on Robotics and
  Automation}, 2008, pp. 1999--2004.

\bibitem{Monari2023}
E.~Monari, Y.~Chen, and R.~Vertechy, ``On locally optimal redundancy resolution
  using the basis of the null space,'' in \emph{Proceedings of the IEEE
  International Conference on Robotics and Automation}, 2023, pp. 9665--9671.

\bibitem{Frobenius1877}
G.~Frobenius, ``{Über das Pfaffsche Problem.}'' \emph{Journal für die reine
  und angewandte Mathematik}, vol.~82, pp. 230--315, 1877.

\bibitem{Selig1996}
J.~M. Selig, \emph{Geometric Fundamentals of Robotics}, D.~Gries and F.~B.
  Schneider, Eds.\hskip 1em plus 0.5em minus 0.4em\relax Springer
  Science+Business Media New York, 1996.

\bibitem{Bullo2004}
F.~Bullo and A.~D. Lewis, \emph{Geometric Control of Mechanical Systems}.\hskip
  1em plus 0.5em minus 0.4em\relax Springer Science+Business Media New York,
  2004, vol.~49.

\bibitem{Park1998}
F.~C. Park and J.~W. Kim, ``Manipulability and singularity analysis of multiple
  robot systems: a geometric approach,'' in \emph{Proceedings of the IEEE
  International Conference on Robotics and Automation}, vol.~2, 1998, pp.
  1032--1037.

\bibitem{Bullo1999}
F.~Bullo and R.~M. Murray, ``Tracking for fully actuated mechanical systems: a
  geometric framework,'' \emph{Automatica}, vol.~35, no.~1, pp. 17--34, 1999.

\bibitem{Lachner2020}
J.~Lachner \emph{et~al.}, ``The influence of coordinates in robotic
  manipulability analysis,'' \emph{Mechanism and Machine Theory}, vol. 146, p.
  103722, 2020.

\bibitem{Lee2012}
J.~M. Lee, \emph{Introduction to Smooth Manifolds}, 2nd~ed., S.~Axler and
  K.~Ribet, Eds.\hskip 1em plus 0.5em minus 0.4em\relax Springer New York,
  2012.

\bibitem{Lewiner2003}
T.~Lewiner, H.~Lopes, A.~W. Vieira, and G.~Tavares, ``Efficient implementation
  of marching cubes cases with topological guarantees,'' \emph{Journal of
  Graphics Tools}, vol.~8, no.~2, pp. 1--15, Dec. 2003.

\bibitem{Eisenhart1928}
L.~P. Eisenhart, ``Dynamical trajectories and geodesics,'' \emph{Annals of
  Mathematics}, vol.~30, no. 1/4, pp. 591--606, 1928.

\bibitem{McKay2020}
B.~McKay, ``Introduction to exterior differential systems,'' arXiv:1706.09697
  [math.DG], Sep. 2020.

\bibitem{Bryant2003}
R.~L. Bryant, P.~A. Griffiths, and D.~A. Grossman, \emph{Exterior Differential
  Systems and Euler-Lagrange Partial Differential Equations}.\hskip 1em plus
  0.5em minus 0.4em\relax University of Chicago Press, 2003.

\bibitem{Bobenko2003}
A.~I. Bobenko, D.~Matthes, and Y.~B. Suris, ``{Discrete and smooth orthogonal
  systems: $C^\infty$-approximation},'' \emph{International Mathematics
  Research Notices}, vol. 2003, no.~45, pp. 2415--2459, Jan. 2003.

\bibitem{Hairer2008}
E.~Hairer, S.~P. N{\o}rsett, and G.~Wanner, \emph{Solving Ordinary Differential
  Equations I: Nonstiff Problems}.\hskip 1em plus 0.5em minus 0.4em\relax
  Springer Berlin Heidelberg, 2008.

\bibitem{Floater2005}
M.~S. Floater and K.~Hormann, ``Surface parameterization: a tutorial and
  survey,'' in \emph{Advances in Multiresolution for Geometric
  Modelling}.\hskip 1em plus 0.5em minus 0.4em\relax Springer Berlin
  Heidelberg, 2005, pp. 157--186.

\bibitem{Rado1926}
T.~Radó, ``Aufgabe 41,'' \emph{Jahresbericht der Deutschen
  Mathematiker-Vereinigung}, vol.~35, p.~49, 1926.

\bibitem{Kneser1926}
H.~Kneser, ``Lösung der aufgabe 41,'' \emph{Jahresbericht der Deutschen
  Mathematiker-Vereinigung}, vol.~35, pp. 116--124, 1926.

\bibitem{Choquet1945}
G.~Choquet, ``Sur un type de transformation analytique généralisant la
  repré-sentation conforme et déﬁné au moyen de fonctions harmoniques.''
  \emph{Bulletin des Sciences Mathématiques}, vol.~69, pp. 156--165, 1945.

\bibitem{Axler2001}
W.~R. Sheldon~Axler, Paul~Bourdon, \emph{Harmonic Function Theory}.\hskip 1em
  plus 0.5em minus 0.4em\relax Springer, New York, NY, 2001.

\bibitem{Maneewongvatana1999}
S.~Maneewongvatana and D.~Mount, ``It{\textquoteright}s okay to be skinny, if
  your friends are fat,'' \emph{Center for Geometric Computing 4th Annual
  Workshop on Computational Geometry}, 1999.

\bibitem{Graepel2003}
T.~Graepel, ``Solving noisy linear operator equations by gaussian processes:
  Application to ordinary and partial differential equations,'' in
  \emph{Proceedings of the 20th International Conference on Machine
  Learning}.\hskip 1em plus 0.5em minus 0.4em\relax Washington, DC, USA: AAAI
  Press, 2003, p. 234–241.

\bibitem{Sachtler2022}
A.~Sachtler and A.~Albu-Schäffer, ``Strict modes everywhere - bringing order
  into dynamics of mechanical systems by a potential compatible with the
  geodesic flow,'' \emph{IEEE Robotics and Automation Letters}, vol.~7, no.~2,
  pp. 2337--2344, 2022.

\bibitem{Sachtler2020}
\BIBentryALTinterwordspacing
A.~Sachtler, ``Orthogonal manifold foliations for impedance control of
  redundant kinematic structures,'' Master's thesis, Technical University of
  Munich, 2020. [Online]: \url{https://mediatum.ub.tum.de/1578939}
\BIBentrySTDinterwordspacing

\bibitem{Abadi2015}
M.~Abadi \emph{et~al.}, ``{TensorFlow}: Large-scale machine learning on
  heterogeneous systems,'' in \emph{Proceedings of the USENIX Symposium on
  Operating Systems Desgin and Implementation}, vol.~12, Nov. 2016, pp.
  265--283.

\bibitem{Hogan1985}
N.~Hogan, ``Impedance control: An approach to manipulation: {Part I-Theory;
  Part II-Implementation; Part III-Applications},'' \emph{Journal of Dynamic
  Systems, Measurement and Control}, vol. 107, no.~1, pp. 1--24, 1985.

\bibitem{Albu-Schaffer2003}
A.~{Albu-Schäffer}, C.~{Ott}, U.~{Frese}, and G.~{Hirzinger}, ``Cartesian
  impedance control of redundant robots: recent results with the
  {DLR}-light-weight-arms,'' in \emph{Proceedings of the IEEE International
  Conference on Robotics and Automation}, vol.~3, Sep. 2003, pp. 3704--3709.

\bibitem{Dietrich2021}
A.~Dietrich \emph{et~al.}, ``Practical consequences of inertia shaping for
  interaction and tracking in robot control,'' \emph{Control Engineering
  Practice}, vol. 114, p. 104875, 2021.

\bibitem{Koditschek1989}
D.~Koditschek, ``The application of total energy as a lyapunov function for
  mechanical control systems,'' \emph{Contemporary Mathematics, American
  Mathematical Society, 1989}, vol.~97, 02 1989.

\bibitem{Schmitz2022}
N.~F. Schmitz, K.-R. Müller, and S.~Chmiela, ``Algorithmic differentiation for
  automated modeling of machine learned force fields,'' \emph{Journal of
  Physical Chemistry Letters}, vol.~13, no.~43, p. 10183 – 10189, 2022.

\bibitem{Botsch2010}
M.~Botsch, L.~Kobbelt, M.~Pauly, P.~Alliez, and B.~L{\'e}vy, \emph{Polygon Mesh
  Processing}.\hskip 1em plus 0.5em minus 0.4em\relax AK Peters/CRC Press,
  2010.

\bibitem{Pauly2018}
M.~Pauly, ``Digital 3d geometry processing,'' Course at Ecole Polytechnique
  Fédérale de Lausanne (EPFL), 2018.

\end{thebibliography}
\bibliographystyle{IEEEtran}

\ifx\ieee\undefined
\appendix
\section{Discrete Harmonic Functions on Curved Manifolds}\label{app:harmonic}

The algorithm below is according to \cite{Botsch2010,Pauly2018}.
Let $v_i$ denote the $i$-th vertex of the mesh.
Let $\mathcal{N}_1(v_i)$ denote the one-ring neighborhood of the $i$-th vertex.
Let $\partial\Sc$ denote the (discrete) boundary of the mesh, i.e. all vertices which are on the boundary.

We solve the linear system
\begin{equation}\label{eq:sparse_harmonic_embedding}
	\b{L}\b{u} = \b{b},
\end{equation}
where $\b{L}$ and $\b{b}$ are sparse matrices created by 
\begin{equation}\label{eq:discrete-laplacian}
	\b{L}_{ij} = \begin{cases}
		1, & \text{if } i = j \text{ and } v_i \in \partial\mathcal{S},\\
		\omega_{ij}, & \text{if } v_j \in \mathcal{N}_1(v_i),\\
		-\sum_{v_j \in \mathcal{N}_1(v_i)} \omega_{ij}, & \text{if } i = j \text{ and } v_i \notin \partial\mathcal{S},\\
		0,& \text{otherwise.}
	\end{cases}
\end{equation}
\begin{equation}
	\b{b}_{ij} = \begin{cases}
		x_i, & \text{if } v_i \in \partial\mathcal{S} \text{ and } j = 1,\\
		y_i, & \text{if } v_i \in \partial\mathcal{S} \text{ and } j = 2,\\
		0, & \text{otherwise.}
	\end{cases}
\end{equation}
The $x_i$ and $y_i$ are the predefined targets for the boundary vertices.
Here, we map the boundary $\partial\Sc$ to the unit-circle and set $x_i$ and $y_i$ accordingly.

In (\ref{eq:discrete-laplacian}) weights $\omega_{ij}$ are
\begin{equation}
	\omega_{ij} = \cot\alpha_{ij} + \cot\beta_{ij},
\end{equation}
where $\alpha_{ij}$ and $\beta_{ij}$ are the two angles opposite to the edge $e_{ij}$ from vertex $v_i$ to $v_j$.
The discrete bijective embedding can then be compute by solving (\ref{eq:sparse_harmonic_embedding}) for $\b{u}$.

\fi

\section{Summary of Ancillary Files}
Four ancillary files are available on the \texttt{arXiv}:
\begin{itemize}
	\item \texttt{smm.mp4:} Animations of a redundant robot moving along self-motion manifolds. Helps understanding Fig.~\ref{fig:3dof-2tasks}.
	\item \texttt{proj.mp4:} Controllers and projections visualized in configuration space for the experiment in Fig.~\ref{fig:projection}.
	\item \texttt{orthfol.mp4:} Configuration space visualization and animation for the experiment in Fig.~\ref{fig:ni_nn_large}.
	\item \texttt{self-motion-manifold-3dof.stl:} 3D-printable model of the self-motion manifold shown in Fig.~\ref{fig:one_leaf_with_y}.
\end{itemize}

\onecolumn
\section{Additional Figures}

\begin{figure}[h!]
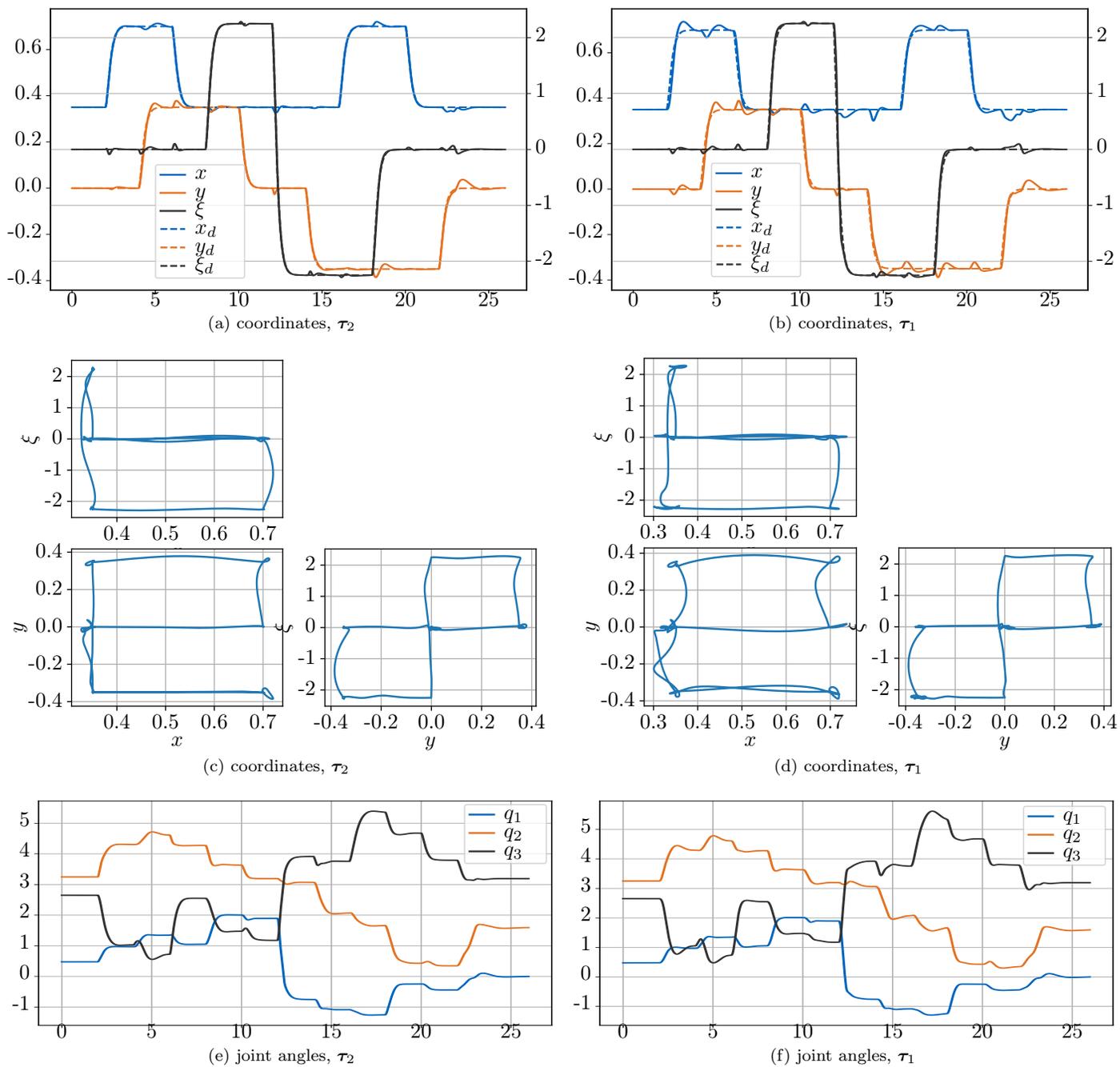

\setlength{\w}{\textwidth}
	\centering
	\subf{coordinates, $\torque_2$\label{fig:ni_nn_longer_c}}{.49\w}{\normalsize}{ni_nn_longer_c}\hfill
	\subf{coordinates, $\torque_1$\label{fig:ni_nn_longer_noff_c}}{.49\w}{\normalsize}{ni_nn_longer_noff_c}\\
	\subf{coordinates, $\torque_2$\label{fig:ni_nn_longer_phase}}{.48\w}{\normalsize}{ni_nn_longer_phase}\hfill
	\subf{coordinates, $\torque_1$\label{fig:ni_nn_longer_noff_phase}}{.48\w}{\normalsize}{ni_nn_longer_noff_phase}\\
	\subf{joint angles, $\torque_2$\label{fig:ni_nn_longer_q}}{.49\w}{\normalsize}{ni_nn_longer_q}\hfill
	\subf{joint angles, $\torque_1$\label{fig:ni_nn_longer_noff_q}}{.49\w}{\normalsize}{ni_nn_longer_noff_q}
	\caption{
		Additional simulation results for the non-involutive example. Left: trajectory tracking, right: impedance control.
		All joint angles are in radians and Cartesian positions in meters; the orthogonal coordinates $\xi_i$ have no unit.
		}
	\label{fig:ni_nn_longer}
\end{figure}
\end{document}